\documentclass[a4paper,fleqn]{cas-dc}
\ExplSyntaxOn
\cs_gset:Npn \__first_footerline:
  { \group_begin: \small \rmfamily \__short_authors: \group_end: }
\ExplSyntaxOff

\renewcommand{\printorcid}{}  % this will make this function do nothing.
\usepackage[authoryear]{natbib}

\usepackage{xcolor}
\usepackage{dashrule}
\usepackage{arydshln}
\usepackage{nicematrix}
\usepackage{pdflscape}
\usepackage{afterpage}
\usepackage{fancyhdr}
\usepackage{caption}
\fancypagestyle{landscapepage}{%
    \fancyhf{} % clear all header and footer fields

}
\usepackage{url}
\makeatletter
\renewcommand{\fnum@table}{{\fontfamily{ptm}\selectfont \tablename~\thetable}}
\renewcommand{\fnum@figure}{{\fontfamily{ptm}\selectfont \figurename~\thefigure}}

\makeatother

\def\tsc#1{\csdef{#1}{\textsc{\lowercase{#1}}\xspace}}
\tsc{WGM}
\tsc{QE}
\begin{document}
\let\WriteBookmarks\relax
\def\floatpagepagefraction{1}
\def\textpagefraction{.001}
% Short title
\shorttitle{}    
% Short author
\shortauthors{}  

% Main title of the paper
\title [mode = title]{Towards scalable organ-level 3D plant segmentation: Bridging the data–algorithm–computing gap}  

% Title footnote mark
% eg: \tnotemark[1]
% \tnotemark[1] 

% Title footnote 1.
% eg: \tnotetext[1]{Title footnote text}
% \tnotetext[1]{} 

% First author
%
% Options: Use if required
% eg: \author[1,3]{Author Name}[type=editor,
%       style=chinese,
%       auid=000,
%       bioid=1,
%       prefix=Sir,
%       orcid=0000-0000-0000-0000,
%       facebook=<facebook id>,
%       twitter=<twitter id>,
%       linkedin=<linkedin id>,
%       gplus=<gplus id>]

\author[1]{Ruiming Du}%[<options>]
\author[2]{Guangxun Zhai}%[]
\author[3]{Tian Qiu}
\author[2]{Yu Jiang}
% Corresponding author indication
\cormark[1]

% Footnote of the first author
% \fnmark[1]

% Email id of the first author
% \ead{}

% URL of the first author
% \ead[url]{}

% Credit authorship
% eg: \credit{Conceptualization of this study, Methodology, Software}
% \credit{}

% Address/affiliation
\affiliation[1]{organization={Department of Biological and Environmental Engineering},
            addressline={Cornell University}, 
            city={Ithaca},
%          citysep={}, % Uncomment if no comma needed between city and postcode
            postcode={14850}, 
            state={NY},
            country={USA}}

% Footnote of the second author
% \fnmark[2]

% Email id of the second author
% \ead{}

% URL of the second author
% \ead[url]{}

% Credit authorship
% \credit{}

% Address/affiliation
\affiliation[2]{organization={Horticulture Section, School of Integrative Plant Science},
            addressline={Cornell AgriTech, Cornell University}, 
            city={Geneva},
%          citysep={}, % Uncomment if no comma needed between city and postcode
            postcode={14456}, 
            state={NY},
            country={USA}}

% Address/affiliation
\affiliation[3]{organization={School of Electrical and Computer Engineering},
            addressline={Cornell University}, 
            city={Ithaca},
%          citysep={}, % Uncomment if no comma needed between city and postcode
            postcode={14850}, 
            state={NY},
            country={USA}}
% Corresponding author text
\cortext[1]{%
  \begin{tabular}[t]{@{}l}
    Corresponding author \\
    {}Email address: yujiang@cornell.edu
  \end{tabular}}
% \ead{yujiang@cornell.edu}
% Footnote text
% \fntext[1]{Email address: yujiang@cornell.edu}

% For a title note without a number/mark
%\nonumnote{}

% Here goes the abstract
\begin{abstract}
The precise characterization of plant morphology provides valuable insights into plant–environment interactions and genetic evolution. A key technology for extracting this information is 3D segmentation, which delineates individual plant organs from complex point clouds. Despite significant progress in general 3D computer vision domains, the adoption of 3D segmentation for plant phenotyping remains limited by three major challenges: i) the scarcity of large-scale annotated datasets, ii) technical difficulties in adapting advanced deep neural networks to plant point clouds, and iii) the lack of standardized benchmarks and evaluation protocols tailored to plant science. This review systematically addresses these barriers by: i) providing an overview of existing 3D plant datasets in the context of general 3D segmentation domains, ii) systematically summarizing deep learning-based methods for point cloud semantic and instance segmentation, iii) introducing Plant Segmentation Studio (PSS), an open-source framework for reproducible benchmarking, and iv) conducting extensive quantitative experiments to evaluate representative networks and sim-to-real learning strategies. Our findings highlight the efficacy of sparse convolutional backbones and transformer-based instance segmentation, while also emphasizing the complementary role of modeling-based and augmentation-based synthetic data generation for sim-to-real learning in reducing annotation demands. In general, this study bridges the gap between algorithmic advances and practical deployment, providing immediate tools for researchers and a roadmap for developing data-efficient and generalizable deep learning solutions in 3D plant phenotyping. Data and code are available at: \url{https://github.com/perrydoremi/PlantSegStudio}.
\end{abstract}

% Use if graphical abstract is present
%\begin{graphicalabstract}
%\includegraphics{}
%\end{graphicalabstract}

% Research highlights
% \begin{highlights}
% \item 
% \item 
% \item 
% \end{highlights}

%\nocite{*}

% Keywords
% Each keyword is seperated by \sep

\maketitle

% Main text
\section{Introduction}\label{sec:Introduction}

Plant morphology addresses the organization of plant organs and provides the taxonomic foundation for understanding plant relationships, environmental adaptations, and genetic evolution \citep{SIMPSON2019469,YANG2020187}. Accurate characterization of morphological traits is essential for breeding programs and plant science research, serving as a cornerstone for advancing agricultural productivity and sustainability \citep{WYATT201651}. Recent advances in imaging technologies and computer vision have enabled sensor-based measurements of morphological traits, generating objective and quantifiable data. These measurements are valuable not only for plant morphology but also for plant genetics, including quantitative trait loci (QTL) analysis, thereby facilitating crop breeding through genetics and genomics approaches. In particular, recent studies have been conducted to extend imaging-based measurements of plant morphological traits from conventional field or whole-plant scales (e.g., plant or plot height) to organ-level resolution (e.g., internode distance or flower and fruit distribution). Access to organ-level traits creates new opportunities to deepen understanding in plant biology and broader life sciences, while also supporting efforts to improve crop adaptability and therefore agriculture resilience to diverse environments \citep{Plantnet,JIN2021202}.

Image segmentation is a key step to extract plant morphological traits at the organ level. Early attempts at imaging-based plant morphology phenotyping have heavily relied on 2D imaging techniques due to the relative ease of image acquisition through RGB cameras and the widespread availability of image segmentation algorithms. These algorithms include both classical computer vision techniques such as thresholding \citep{https://doi.org/10.3732/apps.1400033} and edge detection-based segmentation \citep{WANG20181} and learning-based methods such as convolution neural networks (CNNs) that leverage large annotated image datasets and computing power \citep{doi:10.34133/2020/4152816}. However, 2D imaging-based solutions are highly constrained by the view dependency and the insufficiency of resolving occlusion. Therefore, many recent attempts have focused on 3D imaging techniques.

Among possible representation formats, 3D point clouds have gained substantial popularity in 3D plant phenotyping because they explicitly capture raw 3D coordinates of a scene that can be directly obtained from sensors such as LiDARs while preserving high spatial details. Further, acquiring full-view 3D point clouds dramatically reduces, if not fully avoids, measurement errors caused by view-dependency and occlusion. Similar to 2D imaging-based solutions, 3D point cloud segmentation plays a vital role in the extraction of plant morphological traits. In general, 3D point cloud segmentation can also be categorized as 3D semantic and instance segmentation. 3D semantic segmentation aims to assign a semantic class label to every point in a 3D point cloud, thereby providing a comprehensive mapping of plant organ semantics. The segmentation results enable the analysis of basic organ-level relationships such as hierarchical canopy architecture, leaf-stem ratios, and organ distribution patterns. These traits are essential for studying light interception for photosynthesis, modeling plant resource allocation, and understanding overall organ growth \citep{zhu2020quantification,CONNa,LI2025100002,8468204}. 3D instance segmentation distinguishes between different individual instances of the same class (e.g., identifying and separating each leaf or fruit). The instance segmentation result allows the analysis of intricate spatial arrangements and variation among specific organs. For example, 3D instance segmentation has been used for organ counting, analysis of organ size, shape, and orientation, and organ development and growth \citep{doi:10.34133/plantphenomics.0190,LI2024109435,10.3389/fpls.2022.1012669}.

In early stages, efforts on 3D point cloud segmentation concentrated on semantic segmentation \citep{9028090}. Traditional methods primarily combine hand-crafted features such as point feature histograms (PFH), curvature, and point and surface normal \citep{Rusuinproceedings,Rabbaniarticle} with classical classifiers such as support vector machines (SVM), random forests (RFs), and Bayesian discriminators \citep{Weinmannarticle,6922535,Niemeyerarticle} to classify the semantic meaning of each point \citep {7353481,9028090}. The success of this strategy highly relies on prior knowledge of plant geometry to design local and global features that sufficiently represent the statistical properties of plant point clouds and on exhaustive search of optimal classifiers that adequately use these features \citep{6922535}. In addition to these general 3D point cloud features, plant phenomics researchers actively tailored features for plant morphology, including 3D plant skeletons \citep{BAO201986,doi:10.34133/plantphenomics.0027,10.1007/978-3-030-65414-6_21}, leaf area index (LAI) \citep{s18103576}, and canopy structures \citep{10.1093/aob/mcab087}. Although these methods provided improved results in specific cases, they are resource-demanding for development and cannot be well generalized among plant species or applications. At the same time, 3D point cloud instance segmentation did not receive much attention. A related task is 3D detection in which the proposals (i.e., 3D bounding box) rather than point-wise masks are used for differentiating instances within the same semantic category \citep{NIPS2015_6da37dd3,7780538}. Later on, 3D detection inspired the early research on proposal-based 3D point cloud instance segmentation (Section \ref{sec:Proposal-based methods}).

Recently, deep learning (DL)-based methods have overtaken 3D point cloud semantic segmentation in various domains and have begun focusing on instance segmentation, owing to their powerful ability to learn geometric features directly from data and their superior segmentation performance \citep{qi2017pointnet,wang2018sgpn}. DL-based methods have significantly improved segmentation accuracy and shifted the research focus from feature engineering to deep neural network design across a wide range of computer vision tasks and applications, including autonomous driving, object recognition, photogrammetry, and remote sensing. This shift has led to substantial progress in segmentation-based reasoning of 3D scenes \citep{sun2020scalability,dai2017scannet,Hu_2021_CVPR}. Similar to the success of deep learning in other computer vision tasks and research fields, this improvement has been driven by advancements across the data–algorithm–computing triangle: i) the availability of large-scale annotated datasets that capture statistical patterns, ii) modern deep neural networks capable of learning features from big data, and iii) computing infrastructure (e.g., hardware, streamlined workflows, and evaluation protocols) that supports model training, evaluation, and comparison for continuous improvement.

Organ-level plant morphology phenotyping has advanced through improvements in 3D segmentation as well as broader progress across the data–algorithm–computing triangle \citep{Plantnet,doi:10.34133/2022/9787643,8931235,DU2023380}. Many recent studies in 3D plant phenotyping have employed either custom 3D imaging systems, such as automated multiview stereo systems, or advanced sensors, including LiDAR and laser scanners, to capture high-resolution plant point clouds for extracting morphological traits at the organ level. In parallel, these studies have developed analysis pipelines that leverage DL-based segmentation through supervised or semi-supervised learning schemes, demonstrating satisfactory performance. Nevertheless, the pace of organ-level 3D plant phenotyping has not been sufficient to fully realize its potential, particularly regarding processing scalability and model generalizability. These limitations highlight the need for a systematic review of current challenges and future directions in 3D plant phenotyping, framed through the perspective of the data–algorithm–computing triangle, with particular attention to the unique demands of plant sciences and agriculture.
 
The overarching goal of this review is to provide a systematic synthesis of recent advances in 3D point cloud segmentation for organ level plant phenotyping through the lens of the data–algorithm–computing triangle. Specifically, the review includes four major components:
\begin{enumerate}[i)]
\item An in-depth overview of data collection approaches and publicly available datasets for 3D plant phenotyping at the organ level.
\item A comprehensive analysis of DL-based 3D point cloud segmentation methods.
\item Introduction of Plant Segmentation Studio (PSS), an open source framework designed to streamline benchmark analysis for 3D plant phenotyping.
\item Extensive quantitative experiments on benchmark datasets and representative methods to generate insights for future research.
\end{enumerate}  
% : i) an in depth overview of data collection approaches and publicly available datasets for 3D plant phenotyping at the organ level, ii) a comprehensive analysis of DL-based 3D point cloud segmentation methods, iii) the introduction of Plant Segmentation Studio (PSS), an open source framework designed to streamline benchmark analysis for 3D plant phenotyping, and iv) extensive quantitative experiments on benchmark datasets and representative methods to generate insights for future research. 

We anticipate that this review will not only serve as a key reference for researchers across disciplines interested in DL-based point cloud segmentation for plants but also provide quantitative evidence to inform future research toward scalable 3D segmentation and analysis of organ-level plant phenotyping, thereby advancing the understanding of fundamental life science mechanisms and supporting agricultural sustainability and resilience.

\begin{figure*}%[]
  \centering
   \includegraphics[width=\textwidth]{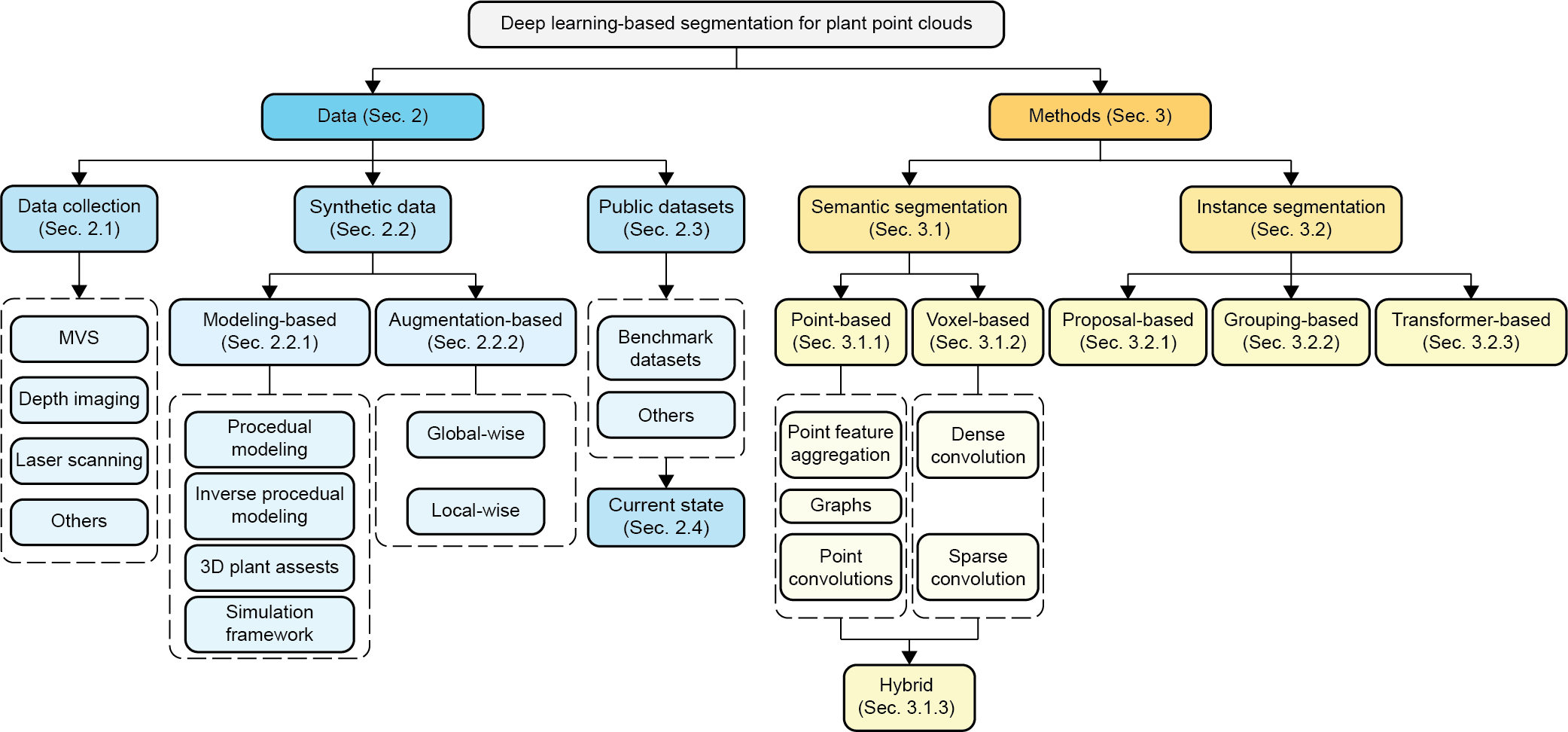}
    \caption{\fontfamily{ptm}\selectfont Review outline of deep learning-based segmentation for plant point clouds.}\label{fig:Fig1}
\end{figure*}

The rest of this manuscript was organized as follows: Section 2 reviews the 3D plant datasets for segmentation-based phenotyping. Section 3 reviews the key deep neural networks for point cloud semantic and instance segmentation. The outline for our review (Section 2 and Section 3) is illustrated in Fig. \ref{fig:Fig1}. Section 4 introduces the development of PSS. Section 5 provides the methods, experiments, and results of selected benchmark datasets, semantic and instance networks, and synthetic training data generation for sim-to-real learning. Section 6 discusses key findings and insights of network design and training strategies for different 3D plant segmentation requirements. Section 7 summarizes and provides perspectives for future directions in 3D segmentation-based plant phenotyping.
% Numbered list
% Use the style of numbering in square brackets.
% If nothing is used, default style will be taken.
%\begin{enumerate}[a)]
%\item 
%\item 
%\item 
%\end{enumerate}  

% Unnumbered list
%\begin{itemize}
%\item 
%\item 
%\item 
%\end{itemize}  

% Description list
%\begin{description}
%\item[]
%\item[] 
%\item[] 
%\end{description}  

% \clearpage %%Remove this from your manuscript

\section{3D plant data collection and synthetic data generation for segmentation}\label{sec:3D plant dataset for phenotyping}

In this section, we reviewed data generation (both data collection and synthetic generation) and publicly available datasets for 3D plant morphology phenotyping, especially at the organ level. While there are many 3D point cloud datasets for plants, the majority of these datasets is for forestry research and forest inventory and collected using terrestrial laser scanning (TLS) \citep{HUI2021219,rs14153842,WANG202086,LIANG201663} or aerial laser scanning (ALS) \citep{BREDE2022113180,XIANG2024114078}. Analyzing these datasets primarily extracts traits at the \textbf{plant-level}, requiring the segmentation of individual trees or the segregation between foliage and woody components. The extracted traits are subsequently used for tree species analysis \citep{HENRICH2024102888,TERRYN2020170,XI20201}, biomass estimation \citep{isprs-archives-XLVIII-1-W2-2023-765-2023,BREDE2022113180,https://doi.org/10.1111/2041-210X.13906}, and foliage property characterization such as foliage density and leaf angle distribution (LAD) \citep{9576621,6249740,https://doi.org/10.1029/2018RG000608,rs12203457}. Within the context of plant phenotyping for agricultural crops, the research interests have been shifted from extracting traits at the plant-level to the \textbf{organ-level} where plant organs (e.g., stem, branch, leaf, flower, and fruit) need to be segmented individually within a single plant. As we aimed to review plant phenotyping for agricultural crops at the organ-level, we only summarized datasets that consist of 3D point clouds with organ-level annotation for agricultural crops in this section.

\subsection{Real world data acquisition}\label{sec:Real-world data}
Organ-level plant segmentation forms the basis for characterizing detailed plant morphological traits and requires high-quality point clouds. Collecting such datasets mainly uses photogrammetry-based solutions such as multi-view stereo and specialized sensors such as depth cameras and laser scanners.

Photogrammetry-based 3D reconstruction is widely used for plant phenotyping because of its affordability and utility, especially multi-view stereo (MVS). MVS reconstructs the 3D structure of a scene using a set of images captured from multiple overlapping viewpoints. Given the camera pose for each image either obtained through hardware configuration or estimated computationally such as using structure from motion (SfM), MVS can compute depth information and generate a detailed 3D reconstruction of the scene from these multiple perspectives.  

SfM-MVS pipeline is highly flexible to deal with various data collection scenarios from controlled environment agriculture (CEA) to in-the-wild environment with consumer-grade RGB cameras or even smartphones \citep{10223838,zhu2024crops3d,SUN2020195,rs12020269,agronomy12081865,rs13112113}. However, the geometric accuracy and resolution of the final 3D data are sensitive to the images quality, the number of viewpoints, and collection environment. Poor texture, insufficient views, or movement of the plant organs will more likely lead to incomplete or inaccurate point clouds. Compared to SfM-MVS, a more robust and repeatable implementation, specifically for CEA environment, is MVS-based high-throughput platforms (HTP) with known camera poses and controlled rotation or movement of plants. Bypassing the SfM and proving precise camera intrinsic and extrinsic parameters, this implementation significantly increases the speed and quality of reconstructed 3D plant data \citep{WU20201848437,Soybean,LI2022106702,10.3389/fpls.2022.897746}. 

Depth cameras (e.g., Time-of-Flight (ToF), structured light, and stereo vision-based)  offer a cost-effective and rapid approach for point cloud collection, especially the development of consumer-grade products (Microsoft Kinect, Intel RealSense, and Orbbec series), which have revolutionized access to 3D imaging by significantly reducing the costs \citep{10.3389/fpls.2023.1097725}. However, they generally suffer from a shorter optimal sensing range and are susceptible to ambient light conditions, and are often integrated into phenotyping platforms \citep{rs11091085,Panicle3D} or robotics \citep{XIE20241624,doi:10.34133/2022/9760269} in laboratories or greenhouses. Such a trade-off makes the plant point cloud stand out in two aspects with different properties: On one hand, the data usually represents potted plants with simple structures for CEA with low resolution and quality compared to laser scanning imaging \citep{CHENE2012122,Plantnet,SONG2025296}. On the other hand, the excellent operational simplicity of depth cameras makes the data capable of describing finer temporal-spatial resolution to enable an easier approach for plant growth monitoring \citep{EschikonPlantStress,visapp20,10.1093/gigascience/giz056}. In addition, the well-established depth calibration methods make the data more feasible to fuse with other modalities (e.g., multi-spectrum, thermal, rgb) to generate multi-modality 3D point clouds \citep{doi:10.34133/plantphenomics.0040,QIU2021106551,WangLiuyangArticle} for studying plant spatial physiology.

Laser scanning encompasses a range of active sensing technologies that obtain 3D information by emitting laser light and analyzing the properties of the reflected signal to determine precise distances to objects. In general, laser scanning takes the leading role in providing high-quality organ-level datasets due to its outstanding capability to capture finer morphological features (e.g., leaf curvature, stem thickness, rice spikes) to generate high-resolution and accurate plant point clouds compared to other modalities. \citep{HAN2021199,SONG2025296,9309060,JIN2021202}. However, given specific laser scanning technologies, the capability for detailing plant organs and the collection scale varies. Currently, there are three main technologies predominant for 3D plant imaging \citep{rs16244720}: Terrestrial laser scanning (TLS) involves stationary systems that scan statistic scenes from fixed positions \citep{10803058,8468204,YUN2025110319}; Highly commercialized handheld laser scanners (HLS), which offer excellent portability and flexibility for small scale 3D imaging \citep{doi:10.34133/plantphenomics.0027,DU2023380}; and mobile laser scanning (MLS) systems, which are usually mounted on various platforms and synergized with georeferencing techniques for large plots or in-the-field 3D imaging \citep{rs9040377,TIOZZOFASIOLO2023104514,doi:10.34133/2022/9760269,10.3389/fpls.2018.00016}. Notable trade-offs are associated with these technologies. Data collection with TLS can be time-consuming due to the need for multiple scan positions, followed by co-registration to mitigate occlusion and cover the entire scene \citep{doi:10.34133/plantphenomics.0179}. While HLS systems reduce occlusion issues compared to static TLS by allowing dynamic scanning around the plant, they require dedicated manual operation for scanning view adjustment, often achieving less throughput compared to TLS and MLS approaches, and are usually used for a single plant in CEA environments \citep{SYAUMaize}. MLS usually achieves less precision compared to TLS or HLS approaches, given the extra registration errors from georeferencing and laser sensors that sacrifice accuracy for throughput in the field \citep{rs15041117,10.3389/fpls.2019.00554}.

Beyond the commonly used depth cameras, MVS, and laser scanning for 3D plant data collection, other specialized techniques, such as X-ray scanning, offer unique capabilities for studying occlusion-free geometry of plant shoot \citep{RoseX,turgut:hal-03144153}, root system \citep{ATKINSON20191,freschet2021starting,tracy2020crop} or achieving ultra-high spatial resolution for tiny organs \citep{10.1371/journal.pone.0075295,wang2017contrast,du2016micron,lowry2024towards}.

In summary, the optimal data collection for specific phenotyping goals hinges on careful assessments of the target plant species, organ characteristics, accuracy, and resolution requirements. No single approach or data format is universally superior, rather, it’s a fitness-for-purpose weighing. Besides, different scenarios will bring different challenges for segmentation. For example, sparse point clouds representing plants with regular or simple structures are typically more amenable to segmentation, with even lightweight network architectures capable of achieving near-optimal performance. In contrast, dense point clouds with complex geometries demand models with stronger local geometric reasoning or computationally efficient algorithms to process high-density data. Likewise, organs with distinct morphologies, such as broad, round leaves versus thin, elongated branches, may require different instance segmentation strategies to achieve robust performance (Discussed in Section \ref{sec:Benchmark results}).

Regardless of the sensing modality or networks employed, the availability of annotated training data remains a critical factor for the success of DL-based segmentation. However, the costly nature of manual annotation continues to limit the broader adoption of advanced DL techniques in plant phenotyping. To address this limitation, the community is pursuing a dual-track strategy: on one hand, enhancing high-throughput, cost-effective imaging systems and expanding high-quality real-world datasets. On the other hand, exploring synthetic data generation to support sim-to-real (sim2real) transfer learning.

\subsection{Synthetic data generation}\label{sec:Synthetic data}

Synthetic data generation can produce large volumes of diverse 3D plant data with perfect, automatically generated ground truth labels, effectively alleviating the annotation burden. 3D plant generation is a broad field including research on functional-structural plant models (FSPMs) for simulating genotype-by-environment (G×E) interactions \citep{vos2010functional}. This review section specifically focuses on synthesis for sim2real plant phenotyping: A strategy that involves training models on synthetic data to solve real-world downstream tasks. In this context, two widely used pipelines are discussed: \textbf{procedural modeling-based (PM-based)}, which simulates plant morphology and sensor interactions from models, and \textbf{augmentation-based}, which leverage existing real-world data to create synthetic variants.

\subsubsection{PM-based}\label{sec:PM-based} 

Early endeavors in 3D synthetic plant generation extensively rely on procedural modeling (PM), with the Lindenmayer system (L-system) and its extension for plant topology simulation being milestones and foundation \citep {LINDENMAYER1968280,10.5555/16564.16608,10.1145/964965.808571}. L-systems employ a set of grammatical rules that can effectively simulate complex branching structures and capture the developmental topology nature inherent in many plants \citep{10.5555/83596}. With the latter efforts expanding the L-systems’ capability to model the interactions between plant parts and their response to the environment \citep{10.1145/166117.166161,mvech1996visual}, L-systems are excellent and have been widely used to generate synthetic plants with high biological plausibility \citep{10.1145/3627101,Lpy-Arabidopsis}. However, driving the procedural models or rules in PM is very demanding and tedious, requiring domain expertise for various manual parameterization \citep{refId0,https://doi.org/10.1111/cgf.13501}. 

To overcome the parameter tuning challenges stemming from most PM pipelines, inverse procedural modeling (IPM) has become the focus \citep{stava2014inverse}. Unlike PM, IPM automates the optimization of procedural models or rules from real-world observations and has fertilized a lot of research on generating 3D synthetic plants for sim2real in plant phenotyping \citep{turgut:hal-03144153,10803058,qiu2025joint3dpointcloud,QIAN2023108124,zhai2024cropcraftinverseproceduralmodeling,SongQingfengArticle,ZAREI2024108922}. Although IPM has demonstrated great capability in generating annotated data, it still requires efforts to choose and calibrate the procedural models. Another alternative is sampling point clouds from existing 3D plant assets. However, most of these plant models are artistically designed for animation purposes, with the priority in virtual realism rather than biological plausibility. Besides, the organ-level annotations are usually not guaranteed, and a post process for labeling is required \citep{Wang_2022_CVPR,10678093}.

For most PM-based approaches, the key limitation hindering their full potential in sim2real application is the discrepancy between the properties of synthetic training data and real-world test data, i.e., the reality gap. It has been demonstrated that introducing the knowledge from source domains can help improve the sim2real performance \citep{WANG2018135}. Under this inspiration, recent advances combine plant modeling with simulation frameworks (e.g., Gazebo, Helios, Helios++) to simulate the plant population and the physics of sensor interactions, e.g., how LiDAR beams propagate, reflect with the plant models in the field \citep{heliosPlusPlus}. This combined pipeline can generate synthetic plant point clouds with imperfections such as occlusion and sensing effects that are more directly comparable and transferable to real sensor data, aiming to bridge the reality gap \citep{qiu2025joint3dpointcloud,rs15092380,TANG2024103903}.

\subsubsection{Augmentation-based}\label{sec:Augmentation-based}

In contrast to PM-based pipelines, augmentation-based pipelines modify existing real-world point clouds to generate synthetic variants. The primary objective is to expand the size and diversity of limited training datasets, thereby enhancing the robustness and generalization capabilities of deep learning models. Specifically, augmentation-based pipelines can be divided into global-wise and local-wise according to the transformation scope \citep{hahner2022quantifyingdataaugmentationlidar}. Global-wise transformations applied to entire point clouds include rotation, which alters plant orientation; scaling, which changes overall plant size; translation, which shifts the plant's spatial position; jittering, which adds small random noise to points; and more recently, different down-sampling strategies tailored for plant point clouds \citep{LI2024172,LiDaweiDownSamplingArticle}. Most of these techniques are cost-effective and do not fundamentally alter the plant's intrinsic structure, which is often used on-the-fly while training the deep neural networks \citep{Pheno4D,AO20221239}. Local-wise methods are becoming an emerging topic for 3D plant phenotyping. They apply local transformations targeting specific plant organs. Such techniques are able to simulate a more natural variability (e.g., different leaf angles, positions, or sizes) or even simplified growth processes through hierarchically transforming certain organs, demonstrating great efficiency in increasing data diversity without the substantial efforts of procedural modeling or sensor simulation \citep{10.3389/fpls.2023.1045545,SYAUMaize}.

However, a persistent challenge within sim2real learning is how to bridge the reality gap. PM-based methods offer high control and the potential to generate biologically plausible data, but can be complex to develop, whereas augmentation-based techniques are often simpler to implement and have higher potential to create unrealistic data, but may be constrained by the diversity of the initial real dataset. This necessitates careful validation of better training strategies with domain adaptation or randomization techniques for sim2real learning \citep{WANG2018135,10.1109/IROS.2017.8202133,James2018SimToRealVS,OpticalFlowinproceedings}.

\subsection{Publicly available organ-level 3D plant datasets}\label{sec:3D Plant Datasets}
An overview of 3D plant datasets that are publicly accessible is summarized in Table \ref{tab:Table1}. Based on data creation method, these datasets are categorized as real-world datasets collected using various 3D imaging techniques or synthetic datasets generated through 3D data synthesis and modeling. Among these datasets, five real world datasets contain annotation for both semantic and instance segmentation at the organ level and are denoted with the asterisk symbol in Table \ref{tab:Table1}. A brief description for each of the five datasets is provided below to help gain a basic understanding of their data acquisition method (and therefore data quality), data size, and intended purpose in 3D plant phenotyping.

\begin{itemize}
\item COS: Cornell orchard single-tree (COS) includes 98 apple tree point clouds from two orchards. Specifically, 48 Gala-W apple trees were grown at Cornell campus orchard (latitude:42.445 N, longitude: 76.462 W) in Ithaca, NY, USA, and 50 apple trees (29 of NY1-SnapDragon, 21 of NY2-RubyFrost) were grown in Cornell Agritech research orchard (latitude: 42.880 N, longitude: 77.006 W) in Geneva, NY, USA. The data collection in both orchards were conducted during the offseason with the maximum visibility of tree trunks and branches through laser sacnning (FARO Focus S350, FARO Technologies Inc., Lake Mary, FL, USA).
\item HR3D: HR3D contains 546 point clouds of three crops (312 tomato, 129 sorghum, and 105 tobacco) through laser scanning (Edge Scan Arm HD, Faro Inc.), covering 20–30 days of development under 3-5 growth conditions (ambient light, shade, high heat, high light, and drought). % In PSS, the three crops were jointly trained and tested.
% and each point cloud was assigned two classes: stem and leaf.
\item SYAU-Maize: SYAU-Maize contains 428 maize point clouds of 5 varieties (Xian Yu 335, LD145, LD502, LD586 and LD 1281). During data collection, maizes were transplanted to pots and point clouds were obtained through laser scanning (FreeScan X3, Tianyuan Inc., Beijing, CN) in an indoor environment.
\item Pheno4D: Pheno4D contains 128 point clouds, in which 77 point clouds of 7 tomato plants over a 20-day growth period, and 49 point clouds of 7 maize plants over a 12-day growth period. Two crops were raised in pots in a greenhouse and were measured daily through laser scanning (Perceptron Inc., Plymouth, MI, USA).
\item SoybeanMVS: SoybeanMVS contains 102 soybean plant point clouds of five varieties (DN251, DN252, DN253, HN48, and HN51) covering the whole soybean growth from the first trifoliolate stage to the full maturity stage. During data collection, 2D images from multiple viewing angles were acquired using a digital camera (Canon EOS 600D SLR, Canon (China) Co. Ltd., Beijing, CN) and subsequently used to reconstruct 3D soybean models through multi-view stereo. A total of 102 point clouds were sampled from the resultant models.
\end{itemize}

\begin{figure}%[]
  \centering
   \includegraphics[width=1.0\columnwidth]{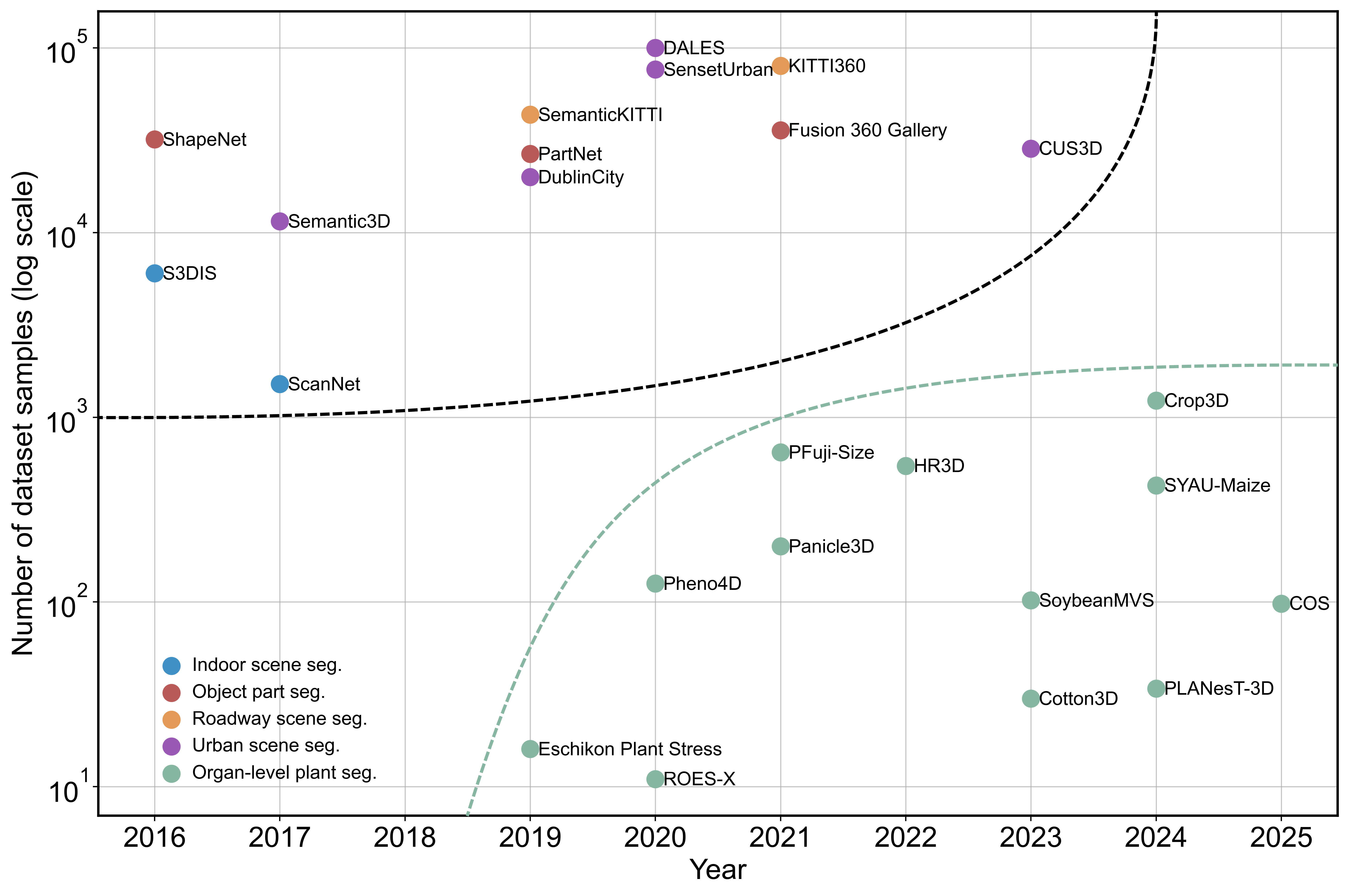}
    \caption{\fontfamily{ptm}\selectfont Data size over publication year of publicly available datasets for both general 3D computer vision (top-left region) and 3D plant phenotyping (bottom-right region). For datasets that only report spatial coverage, we approximated total samples using partition blocks of $1m\times1m$ for indoor scenes and $10m\times10m$ for urban scenes as suggested in \protect\citep{qi2017pointnet,HAN2021199}.}
\label{fig:Fig2}
\end{figure}

While recent efforts have produced more 3D plant datasets, their scale remains considerably smaller than that of widely used benchmark datasets in general 3D computer vision (Figure \ref{fig:Fig2}; compare Table \ref{tab:Table1} with Table \ref{tab:Table2}). On average, only two to three new plant datasets have been released per year, although this number may increase with the continued development and deployment of high throughput phenotyping systems. This trend is promising for advancing algorithmic research in 3D plant analysis, particularly segmentation, which can accelerate plant phenomics and support applications such as crop breeding. Nevertheless, even the largest dataset (Crop3D in 2024 \citep{zhu2024crops3d}) to date only approaches the scale of ScanNet \citep{dai2017scannet}, which was introduced in 2017, and still remains smaller. The seven-year lag in data scale underscores a persistent gap between plant-specific resources and general 3D vision datasets. Furthermore, Crop3D does not provide instance-level organ annotations, which are essential but extremely labor-intensive and costly to generate. To achieve the full potential of 3D computer vision in 3D plant phenotyping, additional datasets of larger scale and richer annotations are urgently needed. At the same time, the availability of these limited yet emerging datasets presents both challenges and unique opportunities for advancing the algorithmic and computational components of the data–algorithm–computing triangle.\\

\begin{table*}
\caption{\fontfamily{ptm}\selectfont Summary of the current available datasets, the datasets marked with * were used in the benchmarks. The number of data samples refer to the samples that are annotated. For simulated datasets, the number of data samples are usually customized by the users.}
\label{tab:Table1}
    \centering
    {\fontfamily{ptm}\selectfont
    \setlength{\tabcolsep}{4pt} % Adjust 
    \begin{tabular}{l p{4cm} c p{1.5cm} >{\centering\arraybackslash}p{1.5cm} p{2.5cm} p{1.5cm} l }
        \toprule
         & Dataset & Year & Plant type & Number of samples & Features & Annotations & Imaging \\
        \midrule
        Real-world & Eschikon Plant Stress \citep{EschikonPlantStress} & 2019 & Sugar beet crop & 16 & X, Y, Z, R, G, B, 25 narrow band reflectance values & \textbackslash & Multi-modalities \\
        & ROES-X \citep{RoseX} & 2020 & Rose & 11 & X, Y, Z & Sem. & X-ray scanning \\
        & *Pheno4D \citep{Pheno4D} & 2020 & Maize; Tomato & 126 & X, Y, Z & Sem., Inst. & Laser scanning \\
        & PFuji-Size \citep{GENEMOLA2021107629} & 2021 & Apple tree; Apple & 6 (Apple tree); 640 (Apple) & X, Y, Z, R, G, B &Sem., Inst. (Apple) & SfM-MVS \\
        & Panicle3D \citep{Panicle3D} & 2021 & Rice & 200 & X, Y, Z & Sem. & Structured light \\
        & *HR3D \citep{CONNa} & 2022 & Tomato; Sorghum; Tobacco & 546 & X, Y, Z & Sem., Inst. & Laser scanning \\
        & Cotton3D \citep{Cotton3D} & 2023 & Cotton & 30 & X, Y, Z, R, G, B & Sem., Inst. & Laser scanning (TLS) \\
        & *SoybeanMVS \citep{Soybean} & 2023 & Soybean & 102 & X, Y, Z, R, G, B & Sem., Inst. & MVS \\
        & *SYAU-Maize \citep{SYAUMaize} & 2024 & Maize & 428 & X, Y, Z & Sem., Inst. & Laser scanning (HLS) \\
        & PLANesT-3D \citep{10223838} & 2024 & Pepper; Rose; Ribes & 34 & X, Y, Z, R, G, B & Sem., Inst. & SfM-MVS \\
        & Crop3D \citep{zhu2024crops3d} & 2024 & Cotton; Maize; Potato; Wheat; Cabbage, etc. & 1230 & X, Y, Z, R, G, B & Sem., Inst. (Plant-level) & Multi-modalities \\
        & *COS \citep{DUASABEinproceedings,doi:10.34133/plantphenomics.0179} & 2025 & Apple tree & 98 & X, Y, Z & Sem., Inst. & Laser scanning (TLS) \\
        \hdashline
        Synthetic & Lpy-Arabidopsis\&Lilac \citep{Lpy-Arabidopsis}& 2020 & Arabidopsis; Lilac & \textbackslash & X, Y, Z, R, G, B & Sem., Inst. & Synthetic \\
        & Simulated roses \citep{turgut:hal-03144153}& 2022 & Rose & \textbackslash & X, Y, Z & Sem., Inst. & Synthetic \\
        & LTree-Panel \citep{qiu2025joint3dpointcloud} & 2025 & Apple tree& \textbackslash & X, Y, Z & Sem., Inst. & Synthetic \\
        \bottomrule
    \end{tabular}
    }
\end{table*}

%\subsection{Comparison with representative domain segmentation datasets}\label{sec:Domain comparison}

\begin{table*}
    \caption{\fontfamily{ptm}\selectfont Representative 3D segmentation datasets across different domains. For certain indoor and urban scene datasets where the total number of samples is not explicitly reported, we provide the total spatial coverage as described in their original publications instead.}
    \label{tab:Table2}
    \centering
    {\fontfamily{ptm}\selectfont
    \setlength{\tabcolsep}{4pt} % Adjust column separation
    \begin{tabular}{l l c l}
        \toprule
         & Dataset & Year & Total samples/Spatial size \\
        \midrule
        Indoor scene seg.& S3DIS \citep{Armeni_2016_CVPR}& 2016 & 6020 $m^{2}$ \\
        & ScanNet \citep{dai2017scannet} & 2017 & 1513 \\
        \hdashline
        \addlinespace[1.2ex]
        Object part seg.& ShapeNet \citep{10.1145/2980179.2980238} & 2016 & 31963 \\
        & PartNet \citep{Mo_2019_CVPR} & 2019 & 26671 \\
        & Fusion 360 Gallery \citep{lambourne2021brepnet} & 2021 & 35858 \\
        \hdashline
        \addlinespace[1.2ex]
        Roadway scene seg. & SemanticKITTI \citep{behley2021ijrr} & 2019 & 43552 \\
        & KITTI360  \citep{Liao2021ARXIV} & 2021 & 80000 \\
        \hdashline
        \addlinespace[1.2ex]
        Urban scene seg. & Semantic3D \citep{isprs-annals-IV-1-W1-91-2017} & 2017 & 30 TLS scans (up to 1.15 $km^{2}$) \\
        & DublinCity \citep{DublinCityinproceedings} & 2019 & 2.0 $km^{2}$ \\
        & SensatUrban \citep{Hu_2021_CVPR} & 2020 & 7.64 $km^{2}$ \\
        & DALES \citep{9150622} & 2020 & 10  $km^{2}$\\
        & CUS3D \citep{rs16061079} & 2023 & 2.85  $km^{2}$\\
        \bottomrule
    \end{tabular}
    }
\end{table*}

% Figure

\section{Deep learning-based methods for 3D point cloud segmentation}\label{sec:Deep learning-based plant point cloud segmentation}

Accurate segmentation of plant point clouds is a crucial prerequisite for 3D plant characterization. In this section, we reviewed representative methods for both point cloud semantic and instance segmentation and their applications for plant phenotyping.

\subsection{Semantic segmentation}\label{sec:semantic_seg}

Semantic segmentation aims to assign a semantic class label to every point in a 3D point cloud. For plant phenotyping at the organ level, this process categorizes points based on their corresponding organ types that are defined by plant biology. In general, semantic segmentation can be divided into point-based, voxel-based, and hybrid methods.

\subsubsection{Point-based methods}\label{sec:point_method}

Point-based methods directly operate on unstructured point clouds. The seminal work of point-based methods is PointNet \citep{qi2017pointnet}. PointNet processes each point independently using a series of shared multi-layer perceptrons (MLPs) to extract features. The most prominent contribution of PointNet is to combine features from individual points as a global feature vector through symmetric functions (e.g., max pooling) that preserve permutation invariance for unordered points in a point cloud. The resultant global feature vector can be used for successive point cloud classification and segmentation. While PointNet has inspired the use of deep learning directly on point clouds, it has limitations in learning local geometric feature and spatial relationship among points, as each point is encoded individually. Successive studies have attempted to addressing these limitations for resolving detailed patterns and processing complex scenes. 

\textbf{Neighboring feature aggregation}: A straightforward extension is to encode features from spatially neighboring points in the metric space, yielding a feature vector that captures both individual point characteristics and the structure of their local region and/or across spatial scales. Point clustering and feature aggregation strategies are the key factors influencing model performance, and extensive studies have investigated diverse options such as ball-query, K nearest neighboring (KNN) search, orientation-specific search, and K-means clustering for point grouping \citep{qi2017PointNetplusplus,jiang2018pointsiftsiftlikenetworkmodule,10.1007/978-3-030-11015-4_29,hu2019randla} and various pooling or weighted combination for feature aggregation \citep{MiningPointCloudinproceedings,8954075,PointConvinproceedings}. Point grouping can also be achieved through transforming raw points into \textbf{graph representations} (e.g., dynamic graphs, superpoint graphs) which capture spatial relationship among points during the construction of the graph. Subsequently, graph theory and processing operations can be used to extract features describing both individual points and their contextual information \citep{8954040,dgcnn,SuperpointGraphsinproceedings,Landrieu2019inproceedings}.

\textbf{Point convolution}: An alternative approach is to extend convolution operations to directly process raw, unordered points. Point clouds can be considered as sparse sampling from a continuous space \citep{8578372}, presenting challenges to design convolution operations with satisfactory computation and memory use efficiency \citep{PointConvinproceedings,thomas2019KPConv,DPCinproceedings} or permutation invariance \citep{xu2021paconv}. A viable option is to reparameterize convolution operations, so these operations can be analytically traceable on unordered points. A notable work is deep parametric continuous convolution in which a MLP is used as the function approximator to estimate continuous convolution values at an arbitrary point sampled from a continuous space, namely, a point in the point cloud \citep{8578372,HORNIK1991251}. Another option is to use positional information to adaptively associate points with kernel weights to reduce the computational complexity. For instance, position adaptive convolution (PAConv) predefines a set of weight matrices and uses an MLP to calculate coefficients of weight matrices at a point based on the relative distances between that point and neighboring ones \citep{xu2021paconv}. This process dynamically creates convolution kernels adapted to point position without computationally expensive one-to-one mapping.

In summary, the major advantage of point-based methods lies in their ability to operate directly on raw point cloud data, thereby preserving fine-grained shape information in the final result. Owing to this advantage, many studies have investigated point-based neural networks for plant point cloud segmentation \citep{Plantnet,RoseX,turgut:hal-03144153,LI2025100002,doi:10.34133/2022/9787643,SYAUMaize,10243159}. Point-based methods still encounter computational challenges, particularly regarding GPU memory limitations, when processing high-resolution and large-scale dense point clouds. Although downsampling can effectively reduce input data volume and, consequently, computational costs, it introduces the critical trade-off between reducing computational complexity and preserving essential information. Excessive downsampling risks the elimination of many plant organs, which are crucial for accurate organ-level plant phenotyping. Therefore, specialized downsampling techniques tailored to this domain are necessary \citep{LI2024172,LiDaweiDownSamplingArticle}. To circumvent this trade-off, sliding window-based inference techniques have been used. In this approach, the point cloud of a scene or object is partitioned into smaller patches, enabling the network to process subsets of points without directly addressing the balance between resolution and computational load. However, this strategy inherently restricts the network's total receptive field and thus its ability to capture global point relationships.

\subsubsection{Voxel-based methods}\label{sec:Voxel-based}

Voxelization converts raw, unordered points into gridded voxels, enabling the use of discrete convolution operations. 3D convolutional neural network (3D CNNs) are a direct choice and provide hierarchical feature extraction through a stack of convolution layers that preserve both local features in shallow layers and global features in deeper layers. This feature of 3D CNNs has proven to be crucial for segmentation tasks \citep {qi2017PointNetplusplus,7298965}. Early attempts have explored dense 3D CNN structure that applies 3D convolution operations across the entire dense voxel grids \citep{Dense3DCNNinproceedings,SEGCloudinproceedings}. The performance of dense 3D CNNs highly relies on the resolution of the voxelization process. The memory usage grows at least cubically for storing dense 3D voxels and calculating 3D convolution \citep{7353481}, requiring a trade-off between computation cost and processing resolution. This trade-off is non-trivial because plant point clouds must have sufficient spatial resolution to keep information of certain types of plant organ (e.g., stems or fruits) for organ-level plant phenotyping \citep{8931235,Plantnet}. In some cases, such a trade-off may not even be available, so dense 3D CNNs have been less used in successive studies. 

Since many voxels are empty (i.e., no point in a voxel) in high-resolution voxel grids, sparse convolution operation has been proposed to exploit this characteristic and perform computation only on non-empty (also known as active) voxels and their neighbors, which drastically reduces computational complexity and memory usage yet avoids the need of spatial resolution loss \citep{graham2015sparse,3DSemanticSegmentationWithSubmanifoldSparseConvNet}. An important development is the submanifold sparse convolution (SSC) operation that shares the list of active points between the input and output to regulate the spread of point activation after convolution. SSC then only computes active input points to avoid unnecessary resource uses compared to dense convolution. Integrating SSC and strided sparse convolution with the U-Net like architecture \citep{10.1007/978-3-319-46723-8_49} forms a network that is capable of handling high-resolution point clouds with leading performance. More recently, sparse convolution operations have been widely adopted to construct neural networks for 3D point cloud segmentation because of the development of specialized libraries such as SpConv \citep{Spconv2022}, Minkowski Engine \citep{choy20194d}, and TorchSparse series \citep{tang2022torchsparse,tangandyang2023torchsparse}. These libraries provide highly optimized implementations of sparse convolution operations, making the training and deployment of deep 3D sparse CNNs much more feasible.

Sparse convolution operations are very suitable for organ-level plant phenotyping because these operations can offer both computation efficiency for large volumetric data points and capability of extracting features across multiple scales through a stack of operations \citep{doi:10.34133/plantphenomics.0080,qiu2025joint3dpointcloud,10.1117/12.3061446,GE202573,10802820}. Determining the proper step size of the voxelization process remains challenging. If the resolution is too coarse, details of small plant organs (e.g., thin branches, stems) can be smoothed or even diminished. Conversely, if the resolution is too fine, the number of active voxels can still become very large, increasing computation costs even with sparse convolutions. Since plant organ size varies at different plant growth stages and among different plant species, it is hard to suggest common values and worthy of further investigation. 

\subsubsection{Hybrid methods}\label{sec:Hybrid methods}

Hybrid methods include methods that combine point clouds with other data representations (e.g., RGB-D images, voxels) \citep{ASurveyarticle}. Specifically, in this review, we refer hybrid methods to those that leverage the advantages of point-based and voxel-based representations (i.e., combine point and voxel representations in data processing). 

The essence of these hybrid methods is to leverage the fine-grained details preserved from light-weight point-based feature extractors (e.g., MLPs) while benefiting from the high capability for neighboring feature learning and the structured nature of voxel-based convolutions. These two streams of information are subsequently fused to generate a comprehensive feature set for segmentation. Representative network includes point-voxel CNN (PVCNN) \citep{liu2019pvcnn} and its successor sparse point-voxel convolution (SPVConv) that integrates sparse convolution to extend its segmentation performance and capability of handling large 3D scenes \citep{tang2020searching}. These hybrid approaches are gaining popularity for plant point cloud segmentation \citep{agriculture15020175}.

\subsection{Instance segmentation}\label{sec:review_inst_seg}

Instance segmentation extends semantic segmentation and further identifies individual instances of the same class. Based on the order of semantic and instance identification, instance segmentation can be categorized into three strategies: proposal-based, grouping-based, and transformer-based.

\subsubsection{Proposal-based methods}\label{sec:Proposal-based methods}

Proposal-based 3D instance segmentation methods, inspired by their 2D counterparts \citep{he2017mask,10.1007/978-3-030-58523-5_38,NEURIPS2020_cd3afef9}, use a top-down strategy that generates potential object proposals (i.e., instance bounding boxes) and then refines these proposals to predict final instance masks~\citep{10.5555/3454287.3454892,8953913,NeuralBFinproceedings,Top-DownBeatsBottom-Up}. While this strategy is intuitive and has demonstrated great success in 2D instance segmentation, it did not achieve expected performance for processing 3D point clouds \citep{Chen_HAIS_2021_ICCV,jiang2020pointgroup,vu2022softgroup,10.1109/TPAMI.2023.3326189,Top-DownBeatsBottom-Up}. A major bottleneck is the precision of proposal generation which influences feature extraction capacity \citep{NeuralBFinproceedings}. In 2D images, proposals can have slight offset from objects of interest but still contain sufficient information for successive feature extraction. In 3D point clouds, points are collected only from object surface and usually cannot represent the inner structure of an object. Therefore, subtle positional shift of proposals (i.e., 3D bounding box) may result in losing a considerable amount object points, presenting significant challenges to feature extraction and subsequent segmentation. This challenge is originated from the nature of data collection of point clouds, limiting the development and application of proposal-based methods. Little or no information has been reported to use proposal-based methods for plant point cloud instance segmentation. As a result, we did not provide further review on this strategy.

\subsubsection{Grouping-based methods}\label{sec:Cluster-based methods}

Grouping-based methods usually build upon a bottom-up strategy that extracts features and predict semantic labels at the point level and then groups points into individual instances. There are two key factors influencing the performance of grouping-based methods: point-wide semantic prediction accuracy and grouping strategy. Advances in semantic segmentation are reviewed in Section \ref{sec:semantic_seg}, so we focus on the improvement of grouping strategy in this section.  

While simple clustering algorithms can be used to group semantic points, recent studies have investigated semantic latent embeddings \citep{LearningandMemorizinginbook,10.1007/978-3-030-58577-8_16,9577908} or graph representations \citep {liang2021instance,Engelmann20CVPR,9157103} to enhance feature richness and representativeness for improved grouping performance. Among recent efforts, PointGroup plays a critical role in advancing the grouping-based 3D instance segmentation \citep{jiang2020pointgroup}. The key of PointGroup is the dual-set point features compromising common point features extracted from MLPs and offset vectors of individual points relative to instance centroids. The common point features are useful to predict semantic labels but may mis-assign instance identification for each point, whereas the offset vectors are good at grouping points into individual instances. Collective utilization of the dual-set features leverages complementary strengths of each feature set. This strategy is particularly promising for handling complex plant point clouds, as it excels at disentangling tightly packed organ instances within individual plants \citep{DU2023380,10.1117/12.3061446,10872974}. Building upon PointGroup, SoftGroup \citep{vu2022softgroup,10.1109/TPAMI.2023.3326189} was proposed to address point grouping errors caused by hard semantic prediction which is a common challenge in the bottom-up strategy of grouping-based methods. SoftGroup enables each point to be associated with multiple semantic scores rather than a single class label, transforming semantic labeling from hard classification to soft probability estimation. This change mitigates the error propagation from the semantic identification step to the grouping step in instance segmentation. 

Grouping-based methods have two general limitations. One is that the point grouping step relies on the semantic identification step, lacking direct instance-level supervision during training and therefore demanding a post-processing refinement \citep{2211.15766}. Another is that grouping-based methods usually have many hyper-parameters that significantly influence model performance yet need careful tuning in different datasets, hindering the generalizability of grouping-based methods across diverse applications \citep{Top-DownBeatsBottom-Up}. 

\subsubsection{Transformer-based methods}\label{sec:Transformer-based methods}

In the past two years, transformer-based methods have been extensively studied and quickly become the leading analysis strategy for 3D instance segmentation. Transformer-based methods take the advantage of attention mechanism to unify feature learning for joint optimization of semantic prediction and instance identification without separating and then combine results from the semantic and instance tasks. Mask3D \citep{Schult23ICRA} is one of the pioneering transformer-based models and integrates 3D sparse convolution U-Net \citep{choy20194d} as feature encoder with a transformer decoder consisting of masked and sampled attentions \citep{cheng2021mask2former}. The sparse U-Net extracts voxel feature maps at multiple spatial scales that are fed into the transformer decoder where each query represents one instance in the scene. In the decoder, each query is cross-attended with the multiscale voxel feature maps and self-attended with other queries to enforce one query per instance. 

Around the same time, superpoint transformer (SPFormer) was proposed to combine 3D sparse U-Net and transformer decoder as a unified end-to-end model for 3D instance segmentation \citep{2211.15766}. SPFormer is different from Mask3D in the model architecture designing. SPFormer explicitly integrates bottom-up and top-down strategies into the unified model architecture. In the encoding stage, in addition to feature maps from 3D sparse U-Net, SPFormer groups raw points into superpoints and applies superpoint pooling to merge raw point features and superpoint features. This process is considered to provide the bottom-up grouping information. In the decoding stage, the transformer decoder cross-attends the superpoint features to generate instance proposals, which is the top-down instance proposal information. Both Mask3D and SPFormer demonstrated significant improvements for 3D instance segmentation, but SPFormer could be a better choice for analyzing large-scale point clouds because of computational efficiency by superpoint \citep{SuperpointGraphsinproceedings}.   

As of writing this review, OneFormer3D \citep{kolodiazhnyi2024oneformer3d} achieves the leading performance for 3D instance segmentation. While maintaining fast inference speed, OneFormer3D further streamlines the integration of feature maps from sparse convolution network and query decoding in transformer decoders by parallel training of semantic and instance queries and a cost-effective query selection. As a result, Oneformer3D solves semantic, instance, and panoptic segmentation jointly. 
 
In summary, transformer-based decoders enable the direct integration of feature learning from input point clouds or voxels into semantic prediction and instance masking. This process minimizes or even avoids subjective design choices and hyperparameter tuning of hand-crafted top-down (proposal-based methods) or bottom-up (grouping-based methods) strategies for instance differentiation. Further, such a streamlined integration would allow joint optimization of multiple 3D segmentation tasks, which usually provide a performance boost in practice. Given these advantages, transformer-based methods have been investigated in 3D plant phenotyping in very recent years \citep{qiu2025joint3dpointcloud,DUASABEinproceedings}.

\section{Plant segmentation studio}\label{PSS}

In addition to the qualitative review in Section \ref{sec:3D plant dataset for phenotyping} and Section \ref{sec:Deep learning-based plant point cloud segmentation}, a quantitative review would be valuable to objectively identify common challenges of 3D point cloud segmentation for organ-level plant phenotyping. The first barrier for conducting a quantitative review for 3D plant phenotyping is the lack of a unified framework that can standardize data management and streamline processing pipeline for comparative studies. In general computer vision, such frameworks like MMDetection, MMDetection3D \citep{mmdet3d2020}, and Detectron2 have enabled the research community to quickly identify issues or recognize significant improvements for substantial advancement. Inspired by this, we developed Plant Segmentation Studio (PSS) as the first attempt to address this considerable gap and used PSS to perform computational experiments for a quantitative review on 3D plant segmentation for organ-level plant phenotyping. 

PSS builds upon MMDetection3D that is a modular framework for a wide range of 3D computer vision tasks including 3D semantic segmentation and detection. Compared with MMDetection3D, PSS is tailored to 3D plant phenotyping by customizing dataset preparation, including representative modules and algorithms for 3D segmentation, and streamlining the configuration of analysis pipelines. 

\subsection{Dataset preparation}\label{Sec: Simplifying data preparation pipeline}

MMDetection3D typically uses a comprehensive dataset-specific pipeline optimized for widely adopted benchmarks, where built-in metadata, such as the mapping between semantic meaning to annotations, and training-validation splits, are hardcoded for efficient data handling and results review. While beneficial for standard benchmark datasets, this approach presents less flexibility for customized applications in two ways:
\begin{enumerate}[i)]
\item Requiring repetitive manual configuration for newly added datasets.
\item Inflexibility when adjusting dataset splits.
\end{enumerate}  

PSS collects and integrates several popular plant point cloud datasets, providing detailed meta information, and implements a simplified data preparation pipeline to convert these datasets into a standardized format. It features a unified, intuitive JSON-based configuration system. Users only need to:
\begin{enumerate}[i)]
\item Structure their point cloud data in a straightforward intermediate format (e.g., txt, csv, ply point clouds).
\item Specify essential metadata within a single JSON file, as shown in Fig. \ref{fig:Fig3}(a) to provide a global semantic-annotations mapping for both semantic and instance segmentation.
\end{enumerate}  

This approach enables an intuitive access to user-customized data and significantly lowers the workload associated with data conversion and standardization.

\subsection{Representative 3D segmentation algorithms}\label{Sec: Improving algorithm compatibility}

The default workflow of MMDetection3D emphasizes 3D object detection, which does not directly support instance segmentation, a critical task for segmentation-based 3D plant phenotyping. To overcome this algorithmic gap, PSS incorporates two representative instance segmentation networks specifically optimized to be compatible with the simplified dataset structure. Specifically, PSS incorporates two advanced segmentation networks: OneFormer3D \citep{kolodiazhnyi2024oneformer3d}, which uses learnable queries to simultaneously perform semantic and instance segmentation; and SoftGroup \citep{vu2022softgroup}, which employs grouping algorithms combined with subsequent refinement steps for instance segmentation. Additionally, a baseline voxel-based method, SCUNet (Section \ref{Sec: Benchmark networks}), is also implemented for benchmarking purposes. 

Given that hyperparameter tuning across various plant species datasets can be tedious, PSS also provides automated tools for inferring essential network hyperparameters directly from user-supplied datasets. Furthermore, all newly integrated segmentation networks can seamlessly interact with existing semantic segmentation backbones provided by MMDetection3D, enabling researchers to efficiently create customized algorithms.

\subsection{Streamlined inference pipelines}

MMDetection3D’s extensive flexibility in network support and customizable design results in complex computational workflows due to intricate module inheritance relationships. Domain scientists in plant phenotyping and agriculture often lack the computer vision expertise required to effectively run these networks and review the results. 

To enhance the usability, except for the efforts in Section \ref{Sec: Simplifying data preparation pipeline}. PSS provides a unified inference pipeline supporting all included segmentation networks. This standardized inference interface ensures comprehensive and consistent processing of segmentation outputs across different network architectures. Specifically, for instance segmentation, PSS applies the appropriate inference logic based on whether the model is transformer-based or grouping-based and provides detailed inference outputs (e.g., shifted coordinates for grouping-based methods. By integrating simplified processes from initial data preparation through final inference within an intuitive workflow, PSS significantly lowers technical barriers. As a result, researchers can efficiently achieve reproducible results, perform systematic algorithm comparisons, and accelerate the development and validation of novel segmentation methods suitable for diverse 3D plant phenotyping applications.

Overall, PSS aims to address existing gaps in data handling, algorithm integration, and computational workflow usability, reducing required expertise and facilitating broader adoption of deep learning-based segmentation approaches among domain scientists. While this study alone does not yet fully establish an integrated data-algorithm-computing nexus, we envision it as a foundational contribution—a step toward enabling future innovations and fostering continued advancement in the field of plant phenotyping.

\begin{figure*}
\begin{center}
    \includegraphics[width=0.9\textwidth]{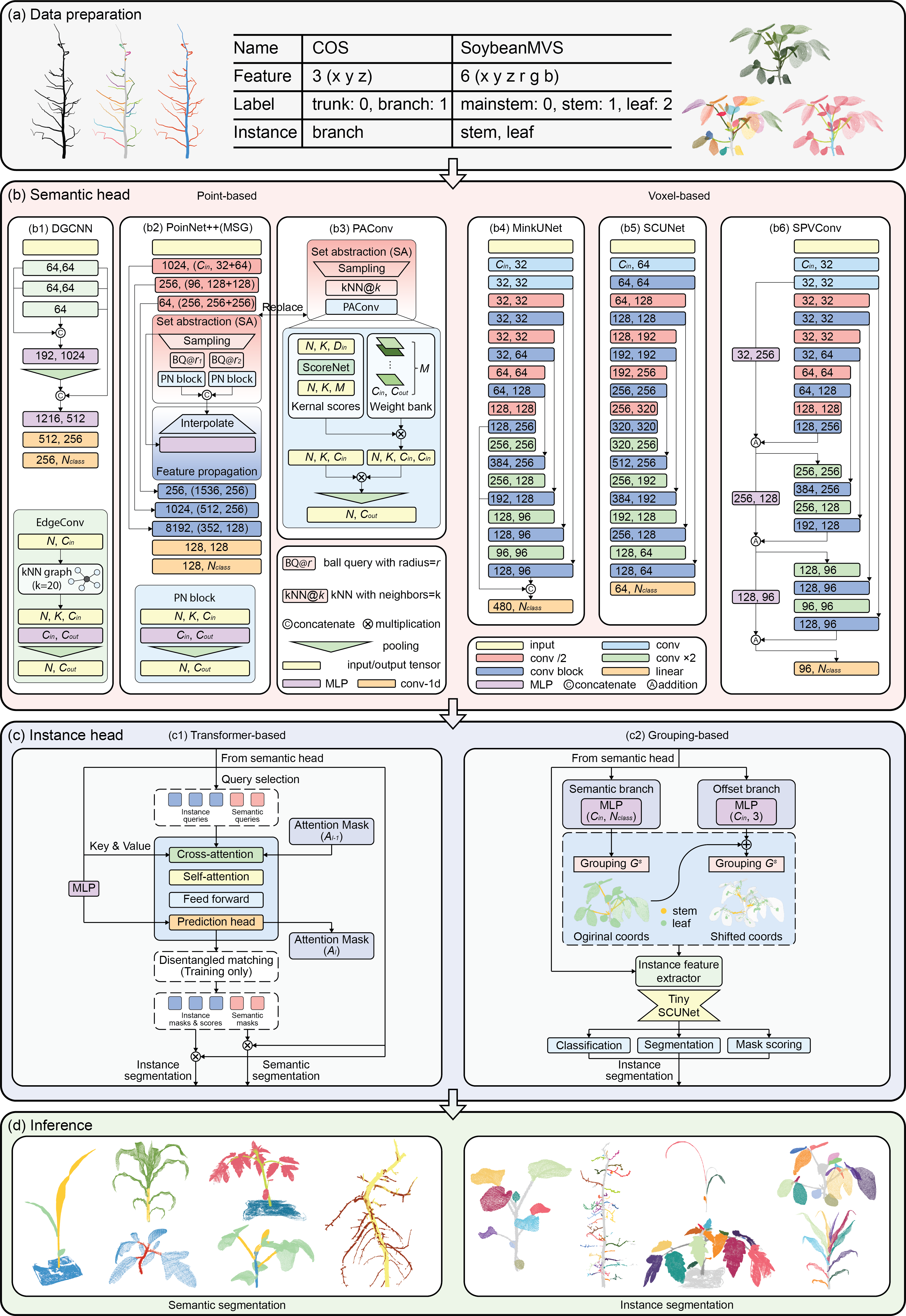}
    \caption{\fontfamily{ptm}\selectfont PSS outline, from data preparation (a), semantic networks (b), instance networks (c) to inference (d).}
    \label{fig:Fig3}
\end{center}
\end{figure*}

\section{Benchmarks}\label{sec:Benchmarks}
\subsection{Methods}\label{Dateset split}
\subsubsection{Benchmark dataset statistics}\label{Benchmark datasets}
The split and annotations for 5 benchmark datasets used through all experiments are listed in Table \ref{tab:Table3}. 

\begin{table*}
    \caption{\fontfamily{ptm}\selectfont Summary of benchmark datasets. HR3D and Pheno4D, which contains multiple species (Table \ref{tab:Table1}), are jointly trained and tested in all experiments.}
    \label{tab:Table3}
    \centering
    {\fontfamily{ptm}\selectfont
    \setlength{\tabcolsep}{4pt} % Adjust 
    \begin{tabular}{ll c c l}
        \toprule
        Dataset & Source & Train& Test & PSS annotations \\
        \midrule
        COS & \citep{DUASABEinproceedings,doi:10.34133/plantphenomics.0179} & 72 & 26 & 0:trunk, 1:branch\\
        HR3D & \citep{CONNa,CONNb} &377& 169 & 0:stem, 1: leaf \\
        SYAU-Maize & \citep{SYAUMaize} & 349 & 79  & 0:stem, 1: leaf\\
        Pheno4D & \citep{Pheno4D} & 90 & 36  & 0:ground, 1:stem, 2:leaf \\
        SoybeanMVS & \citep{Soybean} &65 & 37  & 0:mainstem, 1:stem, 2:leaf\\
        \bottomrule
    \end{tabular}
    }
\end{table*}

\begin{figure*}
\begin{center}
    \includegraphics[width=0.9\textwidth]{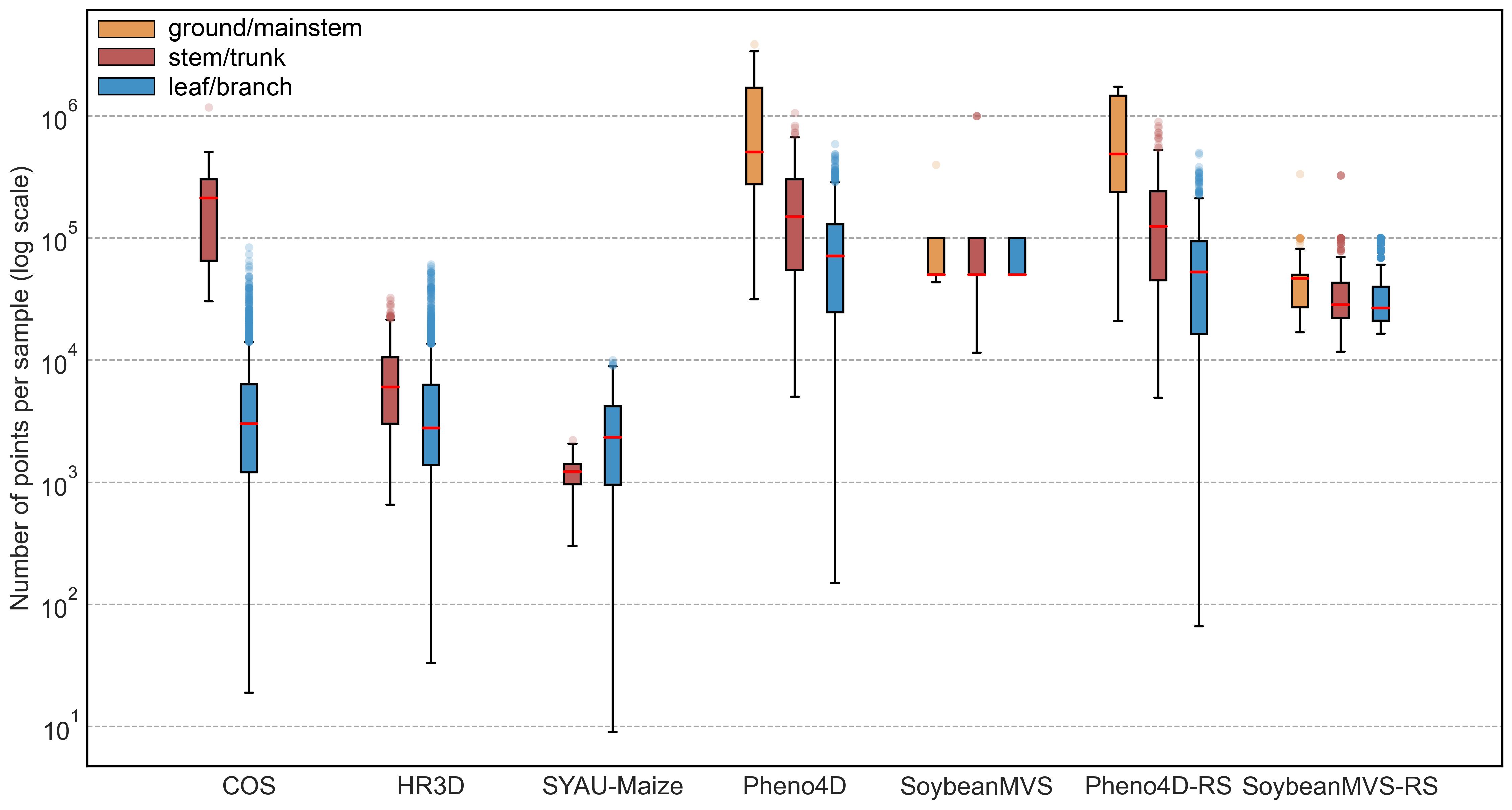}
    \caption{\fontfamily{ptm}\selectfont Distribution of number of points per sample of different semantic classes across seven datasets. Due to the substantial size of the Pheno4D and SoybeanMVS datasets, we created random-sampled subsets (denoted as Pheno4D-RS and SoybeanMVS-RS), reserving the original organ instances statistics (Fig. \ref{fig:Fig5}),  to facilitate instance-level evaluations while reserve the maintaining computational efficiency.}
    \label{fig:Fig4}
\end{center}
\end{figure*}

Processing plant point cloud datasets for DL-based segmentation poses several challenges due to the diverse points and organs distributions, and the sensitivity of morphological representation to point cloud resolution differences across datasets. As shown in Fig. \ref{fig:Fig4}, the number of points per semantic class varies considerably between datasets. Pheno4D and SoybeanMVS offer the highest point densities per sample, often ranging from $10^5$ to $10^6$, providing rich detail across multiple organ classes. In contrast, HR3D and SYAU-Maize contain far fewer points, which may limit their ability to capture fine-grained, organ-level features. Additionally, COS exhibits a highly imbalanced trunk-to-branch class distribution, potentially biasing feature learning during network training.

\begin{figure*}
\begin{center}
    \includegraphics[width=0.9\textwidth]{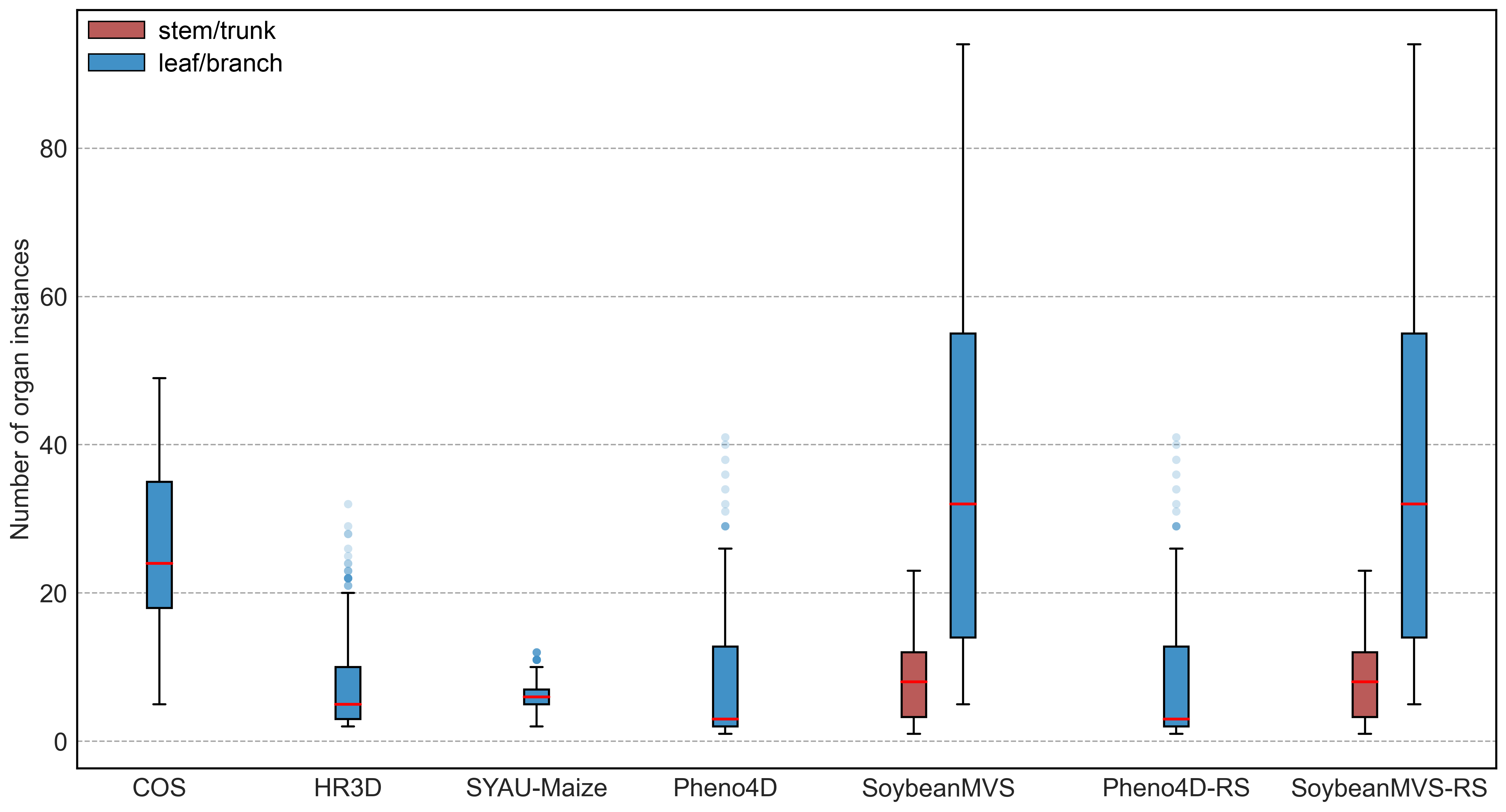}
    \caption{\fontfamily{ptm}\selectfont Distribution of number of organ instances across the seven datasets. The number of certain organs of 0 or 1 in a dataset are ignored for illustration. (e.g., trunk for COS dataset).}
    \label{fig:Fig5}
\end{center}
\end{figure*}

Fig. \ref{fig:Fig5} illustrates the variation in annotated organ instance counts across datasets. HR3D, SYAU-Maize, and Pheno4D include relatively few annotated organ instances (typically fewer than 20 per sample), whereas SoybeanMVS provides a broader range, supporting richer instance-level variability.

\begin{figure*}
\begin{center}
    \includegraphics[width=\textwidth]{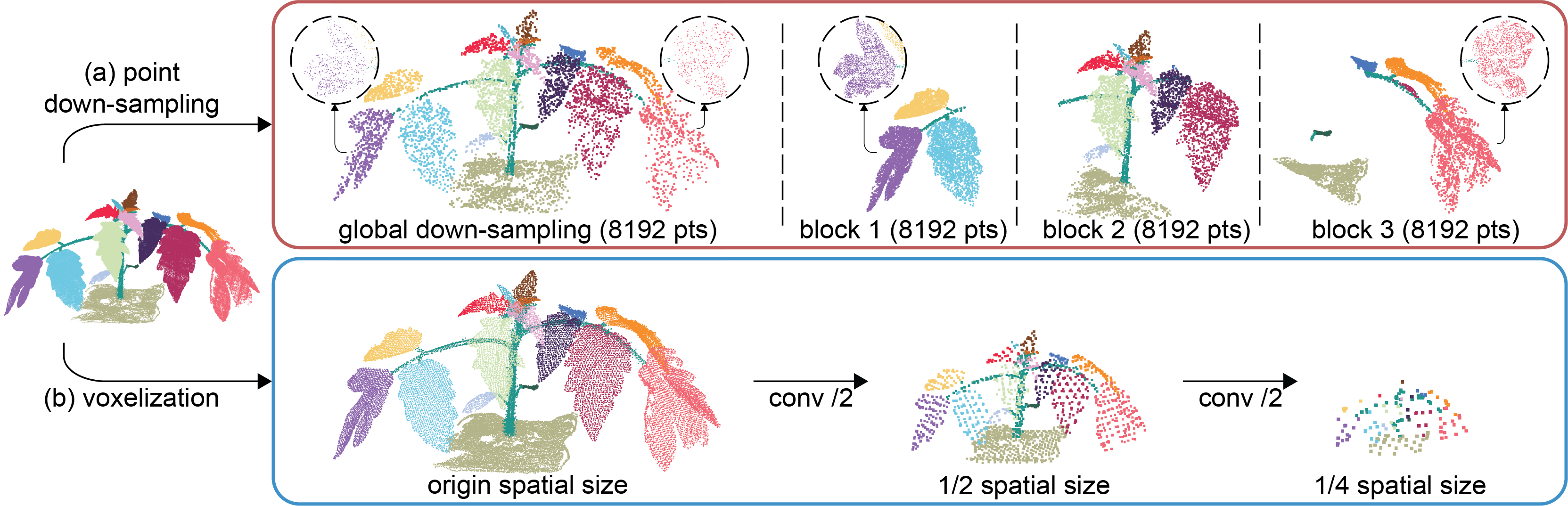}
    \caption{\fontfamily{ptm}\selectfont (a) For point-based methods, point downsampling is necessary to reduce computational cost (downsampling to 8192 points in illustration). However, global downsampling can lead to severe loss of morphological detail. To alleviate this, block-wise downsampling within smaller spatial regions is employed to preserve more information. (b) For voxel-based methods, voxelization transforms point clouds into structured 3D grids. As the network deepens, the resolution of intermediate feature maps decreases significantly due to strided convolutions (stride=2 in illustration). In such cases, skip connections across multiple resolutions are critical for retaining fine-grained morphological information throughout voxel-based network.}
    \label{fig:Fig6}
\end{center}
\end{figure*}

A common challenge across all plant point cloud datasets is the trade-off between input resolution and morphological completeness. Plant morphology comprises thin, elongated, and intricately connected structures that are highly sensitive to data resolution changes. However, deep learning models, whether point-based or voxel-based, require resolution reduction to manage memory and computational constraints. Techniques such as point downsampling or strided convolution for hierarchical feature learning can significantly degrade geometric fidelity, especially in high-resolution datasets, leading to the loss of critical fine-grained information necessary for accurate segmentation (Fig. \ref{fig:Fig6}).

To address these limitations, resolution-aware strategies are essential. In point-based approaches, block-wise downsampling within constrained regions helps preserve structural details, and sliding-window inference during prediction further improves coverage. For voxel-based methods, integrating skip connections between multiple voxel resolutions is crucial for capturing both coarse and fine features.

\subsubsection{Benchmark models}\label{Sec: Benchmark networks}

For semantic segmentation tasks, encoder–decoder architectures have proven highly effective at capturing large receptive fields through sequential downsampling and upsampling operations \citep{xu2021paconv}. Deep neural networks built on this structure typically achieve a balance between computational efficiency and segmentation accuracy. Most grouping-based or transformer–based instance segmentation networks can be regarded as extensions of semantic segmentation models, incorporating an additional instance head on top of the semantic head.

In this study, we benchmark 6 semantic segmentation networks and 6 instance segmentation networks (Fig. \ref{fig:Fig3}). The semantic segmentation baselines include PointNet++, DGCNN, PAConv, SCUNet, MinkUNet, and SPVConv. For instance segmentation, OneFormer3D and SoftGroup are established baselines, while 4 additional models are new implementations that combine different semantic and instance heads (Table \ref{tab:Table4}). The details of these networks are as follows:
% Benchmark configurations and hyperparameter settings from both the original repositories and Plant Segmentation Studio (PSS) are detailed in Table \ref{tab:Table4}.

\textbf{PointNet++}
enhances the original PointNet architecture by introducing a hierarchical set abstraction (SA) structure to capture local geometric features. It consists of multiple SA layers as the encoder and feature propagation (FP) layers as the decoder. Each SA layer comprises three components: a sampling layer, a grouping layer, and a PointNet block. Point resolution is progressively reduced through farthest point sampling, while local features are aggregated via a ball query of neighboring points. The decoder then restores the original resolution through interpolation across FP layers, allowing the network to efficiently capture both local and global structure.

\textbf{DGCNN} 
introduces EdgeConv, an operation that dynamically constructs local neighborhood graphs. EdgeConv explicitly models relationships between a point and its k-nearest neighbors (kNN) and aggregates features through max-pooling to produce robust global descriptors. Unlike static graph approaches, DGCNN recomputes graph structures in the feature space at each layer, enabling the network to capture increasingly rich semantic relationships and geometric detail across all points.

\textbf{PAConv}
introduces a position-adaptive convolution for point clouds by dynamically assembling convolutional kernels. Instead of fixed kernels, PAConv generates kernels from a learnable Weight Bank based on the relative positions of points, with weights predicted by ScoreNet. This flexible kernel assembly improves segmentation performance on irregular point clouds and can be integrated into various backbones. In our benchmarks, PAConv replaces all MLP layers in a PointNet++ backbone, and we refer to this configuration simply as “PAConv.”

\textbf{Sparse convolution U-Net (SCUNet)}
extends the 3D U-Net architecture with sparse convolution operations to address the sparsity of 3D point clouds. Unlike dense voxelization methods subject to high memory and computational costs, SCUNet uses submanifold and sparse convolutions to process only occupied voxels efficiently. The encoder gradually aggregates multi-scale features, while the decoder restores spatial resolution through transposed sparse convolutions and skip connections. This design improves computational efficiency and accuracy. Currently, there are three widely-used backends: SpConv \citep{Spconv2022}, Minkowski Engine \citep{choy20194d}, and TorchSparse \citep{tang2022torchsparse,tangandyang2023torchsparse++,tangandyang2023torchsparse} support sparse convolutions. In benchmarks, SCUNet is implemented using SpConv following OneFormer3D’s configurations.

\textbf{MinkUnet}
builds upon SCUNet by incorporating skip concatenation between encoder and decoder layers to fuse fine-grained spatial details with high-level semantic features. Our implementation uses TorchSparse, leveraging its optimizations for sparse tensor operations. Detailed backend comparisons are available in \citep{mmdet3d2020}.

\textbf{Sparse point-voxel convolution (SPVConv)}
is a hybrid framework combining point-based and voxel-based representations. A dual-branch design enables point-based modules to capture fine-grained details while voxel-based modules, structured as a sparse convolutional U-Net, provide a broad receptive field and contextual information. Feature fusion occurs at multiple stages, enabling SPVConv to leverage the strengths of both representations. We use TorchSparse for implementation in benchmarks.

\textbf{OneFormer3D}
integrates voxel features with learnable semantic and instance queries processed by six sequential transformer decoders. Each decoder applies cross-attention between queries and voxel features, alongside self-attention among queries. Attention masks are progressively refined to spatially constrain cross-attention of queries to more specific voxel
regions for distinguishing different semantics and instances.

\textbf{SoftGroup}
employs a semantic branch to predict point-level semantic scores and an offset branch to estimate vectors shifting points toward their respective instance centers. Grouping is performed on these dual coordinates using a ball query, guided by semantic scores. In the refinement stage, a lightweight U-Net predicts instance class labels, masks, and confidence scores to produce the final instance segmentation results.

\begin{table*}[ht]
    \caption{\fontfamily{ptm}\selectfont Summary of semantic and instance segmentation networks in the benchmark. For simplicity, we name the 4 new implemented networks according to their semantic and instance heads combinations. E.g., Mink-SG (MinkUnet with SoftGroup grouping), SPVConv-SG (SPVConv with SoftGroup grouping), MinkFormer (MinkUnet with OneFormer3D query decoder), and SPVFormer (SPVConv with OneFormer3D query decoder). Notably, for SPVConv, its point-branch use the points after quantization, therefore, we consider it both hybrid and voxel-based methods.}
    \label{tab:Table4}
    \centering
    {\fontfamily{ptm}\selectfont
    \setlength{\tabcolsep}{4pt} % Adjust column separation
    \begin{tabular}{l l p{3cm} p{6cm} p{3cm}}
        \toprule
         &  & Representation & Encoder & Decoder \\
        \midrule
        Semantic & PointNet++ & Point-based & Set abstractions (w/ MLP) & Feature propagation \\
         & DGCNN & Point/Graph-based & Edge convolution & MLP \\
         & PAConv & Point-based & Set abstractions (w/ position adaptive convolution) & Feature propagation \\
         & SCUNet & Voxel-based & \multicolumn{2}{{p{9cm}}}{SCUNet (w/ SpConv backend)} \\
         & MinkUnet & Voxel-based & \multicolumn{2}{{p{9cm}}}{SCUNet (w/ Torchsparse backend, skip concatenation)} \\
         & SPVConv & Hybrid/Voxel-based & \multicolumn{2}{{p{9cm}}}{SCUNet (w/ Torchsparse backend, Point-wise MLP, skip addition)}\\ 

        \hdashline
         &  &  & Semantic head & Instance head \\
        \midrule
        Instance & SoftGroup & Voxel-based & SCUNet \citep{vu2022softgroup} & Grouping \\
         & Mink-SG & Voxel-based & MinkUnet & Grouping \\
         & SPVConv-SG & Hybrid & SPVConv & Grouping \\
         & OneFormer3D & Voxel-based & SCUNet \citep{kolodiazhnyi2024oneformer3d} & Transformer decoder \\
         & MinkFormer & Voxel-based & MinkUnet & Transformer decoder \\
         & SPVFormer & Hybrid/Voxel-based & SPVConv & Transformer decoder \\
        \bottomrule
    \end{tabular}
    }
\end{table*}

\subsubsection{Synthetic training data generation}\label{Synthetic training data}

Based on the discussion of synthetic 3D plant data generation in Section \ref{sec:Synthetic data} and their potential for sim2real learning, this section focuses on two representative implementations of PM-based and augmentation-based pipelines applied to COS datasets:

\textbf{L-TreeGen}
comprises two key components: (i) a tree generation (TG) module based on inverse procedural modeling to create biologically plausible tree meshes, and (ii) a virtual laser scanning (VLS) module to replicate the sensing characteristics of laser scanners used in the field. The TG module operates in two steps. First, trunk and primary branch morphologies are interpolated from key statistics (e.g., trunk height, skeleton structure, branch radius) derived from real tree point clouds. Next, higher-order branches are hierarchically scaled and attached to primary branches to create realistic, complex branch structures. Biological constraints, including tapering effects (progressive thinning of trunks and branches) and collision detection between organs, are incorporated to ensure morphological plausibility. The VLS module, implemented using Helios++ \citep{heliosPlusPlus}, enables configurable sensor parameters to simulate acquisition imperfections such as noise, occlusion, and variations in resolution, angle, and scanner positioning. L-TreeGen not only produces biologically realistic tree models but also replicates sensor artifacts, generating imperfect training data designed to improve neural network generalization to real-world conditions.

\textbf{Deformation}
uses physics-based elastic deformation to deform plant structures through simulated external forces to generate new plant point clouds. This process has two main steps. First, the original plant point cloud is voxelized, and physical properties (e.g., Young’s modulus, Poisson’s ratio) are assigned to voxels representing different organs. Second, external forces of controlled magnitude, direction, and application positions (i.e., voxel vertices) are applied. Throughout the whole process, elastic deformation mechanics ensure accurate force transmission and preserve organ topology, minimizing artifacts. By systematically changing applied forces, deformation enhances dataset variability while introducing fewer distortions compared to brute-force augmentation techniques such as scaling, rotation, flipping, or jittering. Following the strategy in \citep{SYAUMaize}, we conducted deformation experiments using a voxel size of 0.001 and external forces ranging from -5 to 5 N along the X, Y, and Z axes.

The evaluation of L-TreeGen and deformation provides insight into the practical benefits and limitations of each approach. Both demonstrate effective strategies for reducing annotation burdens and improving model generalization. PM-based methods emphasize biological realism but involve higher complexity, whereas augmentation-based methods focus on feature diversity and scalability, often at the cost of reduced biological fidelity.

\subsection{Experiments}\label{sec:Experiments}
The experiments are divided into 3 parts. First, we benchmark the performance of 6 semantic segmentation networks. Next, we select 2 representative semantic heads and 2 instance heads to construct 4 new instance networks, which are evaluated alongside the original SoftGroup and OneFormer3D. Finally, we use the best-performing instance network to investigate the effectiveness of synthetic data for sim2real learning. All experiments were conducted in PSS.

\subsubsection{Data preparation for semantic and instance segmentation}\label{sec:Segmentation datasets preparation}
All 5 benchmark datasets (Table \ref{tab:Table3}) are used to evaluate the performance of semantic and instance segmentation networks. During training, each dataset is augmented on-the-fly to ensure a sufficient training set size. Specifically, random sampling is applied for point-based semantic segmentation, while flipping and scaling are used for voxel-based semantic segmentation. These augmentation methods are chosen to enable each network to reach its maximum performance potential.

\subsubsection{Data preparation for sim2real learning}\label{sec:Synthetic training data preparation}
Using the established L-TreeGen and deformation pipelines, we conducted 3 experiments on the COS dataset to assess the potential of synthetic data for sim2real learning and to determine which synthetic data generation strategy is most suitable for different segmentation tasks. The training and testing splits are identical to those in the semantic and instance segmentation benchmarks (72 training and 26 testing samples). To ensure balanced representation, training samples were equally selected from two orchards (36 from each), as the trees exhibit different morphological characteristics in each orchard. This balance was also maintained when selecting base trees for synthetic data generation.

For clarity, we define the real-world trees used to generate synthetic data as \textbf{base trees}, with the number of base trees in a subset denoted as $K_{b}$. Unless noted, each subset of base trees was used to generate 150 synthetic trees for training. We categorize the experiment setups as follows:
\begin{itemize}
    \item \textbf{0-shot}: Training only on synthetic data, followed by direct testing on real-world data.
    \item \textbf{$K_{b}$-shot}: Pre-training with synthetic data, fine-tuning with $K_{b}$ base trees, and testing on real-world data.
    \item \textbf{Vanilla (Baseline)}: Training exclusively on all real-world data without synthetic data.
\end{itemize}
The three experiments are designed as follows:

\begin{itemize}
    \item \textbf{Robustness of L-TreeGen and deformation in 0-shot segmentation}: When the number of base trees is small, their composition can strongly influence the geometric feature distribution of the generated synthetic trees, potentially introducing learning bias. To investigate this effect across different synthetic data generation methods, we randomly sampled six base trees ($K_{b}=6$, where 3 from each orchard) without replacement from the training set of 72 trees to form 12 non-overlapping subsets (12 folds). We then evaluated the 0-shot performance for each subset.
    \item \textbf{Segmentation performance with different base tree sizes for synthetic data generation}: The 12 subsets from the above experiment were ranked by instance segmentation performance (using AP metric). We aggregated base trees from the top-performing folds to create larger subsets with $K_{b}=12$, $18$, and $24$, representing upper-bound subsets. Similarly, subsets were formed from the lowest-ranked folds to establish lower-bound subsets. We evaluated both 0-shot and $K_{b}$-shot performance for all subsets.
    \item \textbf{Resolution comparison of L-TreeGen}: More realistic synthetic data, incorporating sensor imperfections, may improve network generalization on real-world data. In this experiment, both TG and VLS modules of L-TreeGen were used to generate synthetic trees at three sensor angular resolutions: 0.3, 0.06, and 0.03 degrees (denoted as VLS03, VLS006, and VLS003). Lower values indicate higher point cloud resolution. Using the upper-bound subset with $K_{b}=24$, we compared 0-shot and $K_{b}$-shot performance to assess the effect of resolution and establish the best achievable performance of synthetic data for sim2real learning relative to the vanilla setup.
\end{itemize}

\begin{figure*}
\begin{center}
    \includegraphics[width=\textwidth]{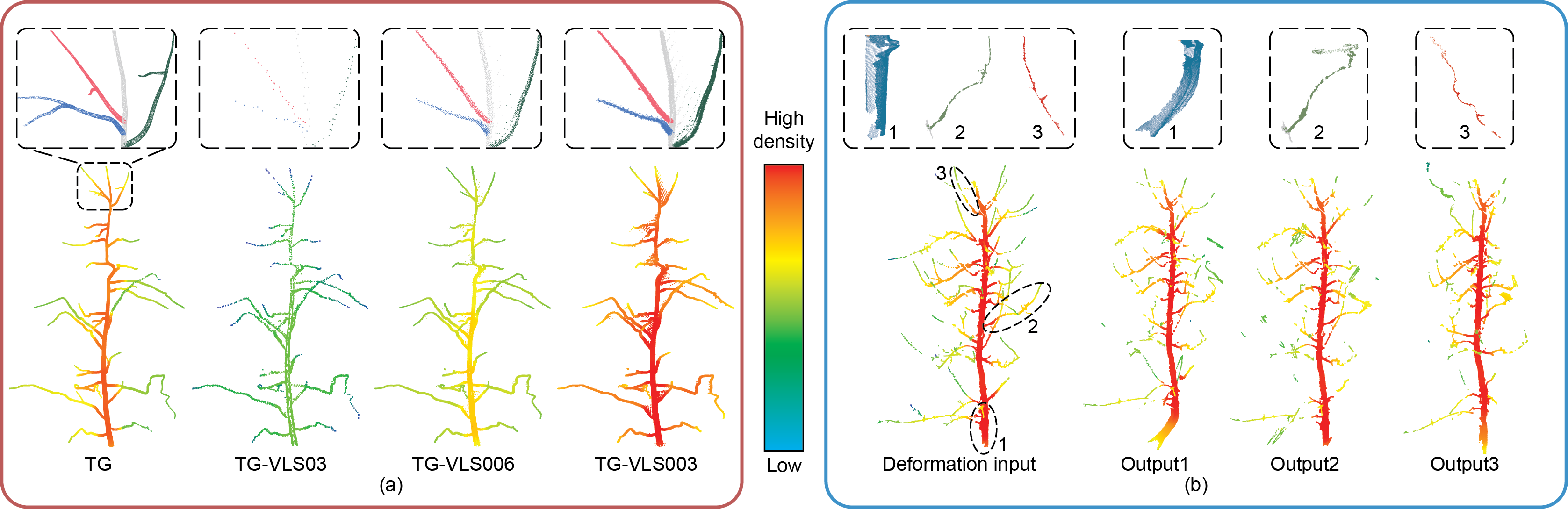}
    \caption{\fontfamily{ptm}\selectfont Apple tree point clouds generated from L-TreeGen (a), and deformation (b). For L-TreeGen, the points density and geometry details highly rely on the sensor resolution used in VLS module. For deformation, local-wise transformation are applied to different organs, and the points density are close to the original input point clouds.}
    \label{fig:Fig7}
\end{center}
\end{figure*}

Detailed data preparation for these experiments is shown in Table \ref{tab:Table5}, and examples of synthetic apple tree point clouds generated using L-TreeGen and deformation are illustrated in Fig. \ref{fig:Fig7}.

\begin{table*}
    \caption{\fontfamily{ptm}\selectfont Data preparation for L-TreeGen and deformation generation. TG means we only use TG module in L-TreeGen pipeline for generation; TG+VLS03 means we use both TG and VLS modules and set the angular resolution degree to 0.3 to generate more realistic synthetic trees. $K_{b}$ represent the number of base trees selected from training sets from both orchards, with the ratio in the last column indicating the proportion of the total testing set (26 real trees identical in all benchmarks) represented by $K_{b}$.}
    \label{tab:Table5}
    \centering
    {\fontfamily{ptm}\selectfont
    \renewcommand{\arraystretch}{1.2} % Increase row height
    \setlength{\tabcolsep}{4pt} % Adjust spacing
    \begin{NiceTabular}{l c|c c c c}
        \toprule
        L-TreeGen & Deformation & Subsets & \multicolumn{2}{c}{$K_{b}$} & Ratio ($K_{b}$/Testing set size) \\
        \cmidrule(lr){4-5}
        & & & Orchard1 & Orchard2 & \\
        \midrule
        TG1 & D1 & 12 (Folds) & 3 & 3 & 0.23 \\
        TG2 & D2 & 2 (L, U bound) & 6 & 6 & 0.46 \\
        TG3 & D3 & 2 (L, U bound) & 9 & 9 & 0.69 \\
        TG4 & D4 & 2 (L, U bound) & 12 & 12 & 0.92 \\
        TG4+VLS03 & & 1 (U bound) & 12 & 12 & 0.92 \\
        TG4+VLS006 & & 1 (U bound) & 12 & 12 & 0.92 \\
        TG4+VLS003 & & 1 (U bound) & 12 & 12 & 0.92 \\
        \bottomrule
    \end{NiceTabular}
}
\end{table*}

\subsubsection{Evaluation metrics}\label{Evaluation metrics}

Semantic and instance segmentation performance is evaluated separately following standard procedures from \citep{dai2017scannet}, \citep{kolodiazhnyi2024oneformer3d}, and \citep{Schult23ICRA}. Since semantic segmentation is generally less challenging and serves primarily as a preprocessing step for instance segmentation, a broader set of metrics is reported for semantic segmentation to evaluate its potential integration with instance segmentation heads. These include Intersection over Union (IoU), precision, recall, F1-score, inference speed (FPS), and model parameters.

For instance segmentation, we focus on accuracy, reporting average precision (AP) at IoU thresholds of 25\% (AP25) and 50\% (AP50), as well as mean AP across thresholds from 50\% to 95\% with a step size of 5\%. Semantic performance of instance networks is also recorded. AP25 and AP50 measure network performance under relatively moderate criteria, while mean AP reflects overall capability under increasingly strict thresholds.

Additionally, for the robustness experiment on 0-shot segmentation with L-TreeGen and deformation, we report both the mean and standard deviation (SD) of mIoU, AP25, AP50, and mean AP across all 12 folds.

\subsection{Benchmark results}\label{sec:Benchmark results} 
\subsubsection{Semantic segmentation}\label{sec:Semantic segmentation} 

The semantic segmentation benchmark across 5 datasets for various networks is summarized in Table \ref{tab:Table6}. 

\begin{figure}
\begin{center}
	\includegraphics[width=\columnwidth]{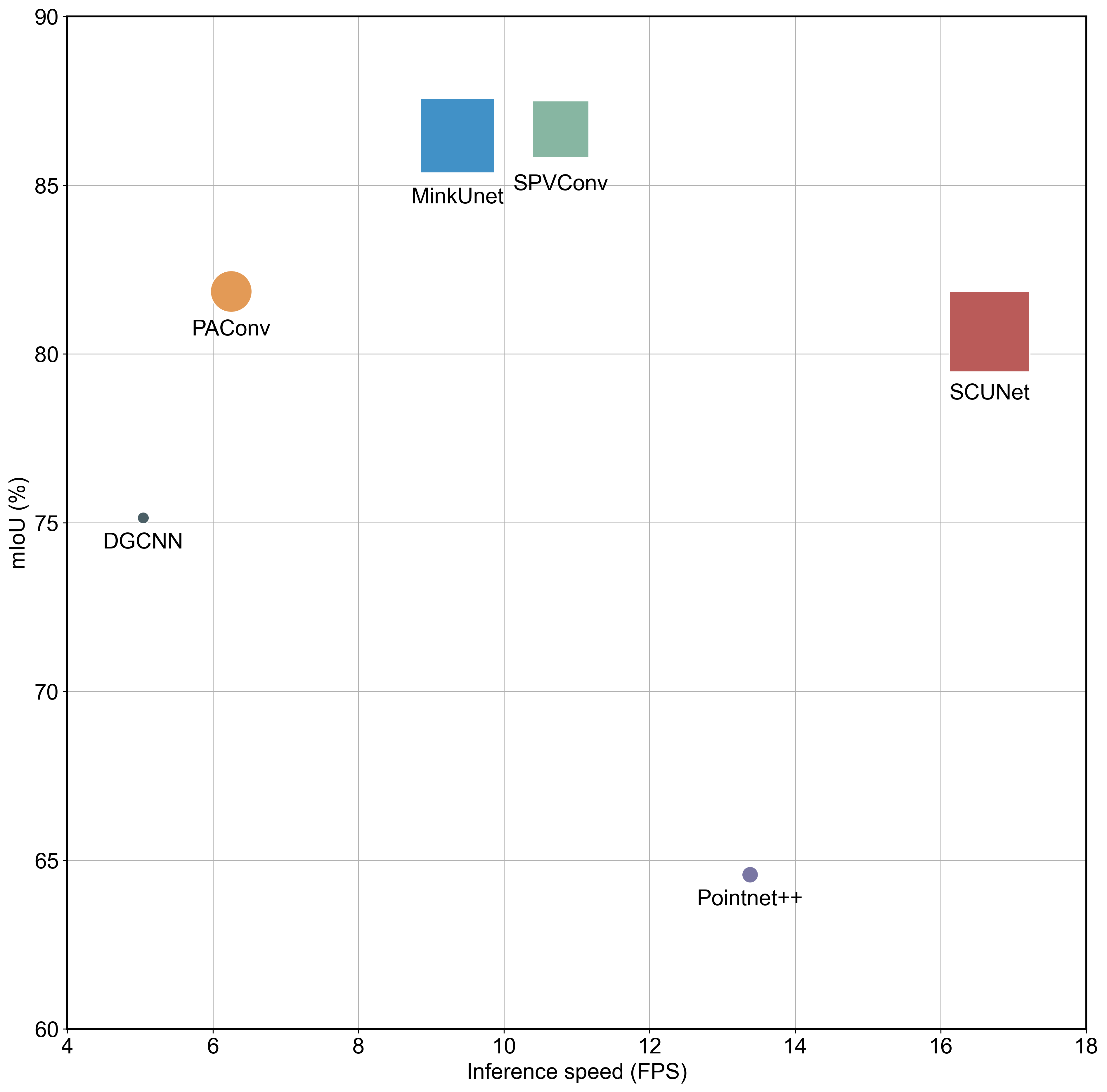}
	\caption{\fontfamily{ptm}\selectfont Semantic segmentation benchmarking on six networks across all datasets. Circles: Point-based methods, squares: Voxel-based methods. The radius reflects the model size, larger radius indicates that the model has more parameters (FPS: Frame/Samples per seconds).}
\label{fig:Fig8}
\end{center}
\end{figure}

\begin{itemize} 

\item \textbf{Voxel-based networks generally outperformed point-based networks}: Among all six semantic networks, SPVConv achieved the highest mean mIoU of 86.66\% across all datasets. MinkUNet, SCUNet, and PAConv also performed well, with mIoUs of 86.47\%, 80.66\%, and 81.85\%, respectively. In contrast, point-based networks with fewer parameters, such as DGCNN (0.96M) and PointNet++ (1.88M), delivered lower segmentation accuracies. These results indicate that voxel-based methods with sparse convolution, although requiring more parameters, consistently outperform lighter point-based approaches. For example, SPVConv, the lightest and best-performing voxel/hybrid-based network, achieved a 5.88\% mIoU improvement with 84.89\% more parameters compared to PAConv, the best-performing point-based method. 

\item \textbf{Efficient sparse convolution backends and skip connections are crucial for SCUNet-like networks}: In voxel-based and hybrid methods (e.g., SCUNet, MinkUNet, and SPVConv), the sparse convolution backend plays a key role. In our benchmark, MinkUNet and SPVConv, both using the more efficient Torchsparse backend, achieved better performance with fewer parameters than SCUNet, which uses the SpConv backend. 

Additionally, skip connections that propagate information across the voxel-based U-Net, whether via addition or concatenation, proved highly beneficial, and sometimes contributed more to performance than simply increasing hidden layer channels. For instance, MinkUNet concatenates voxel features of varying spatial resolutions before the final linear layer (Fig. \ref{fig:Fig3} (b4)), while SPVConv continuously refines voxel-branch features with point-branch features through addition (Fig. \ref{fig:Fig3} (b6)). Although SCUNet, MinkUNet, and SPVConv share similar voxel-convolution U-Net structures, MinkUNet and SPVConv demonstrated superior ability to capture fine-grained features and accurately segment challenging minor classes (e.g., the mainstem in the SoybeanMVS dataset). In contrast, SCUNet, despite using larger base and convolution channel sizes ([64, 128, 192, 256, 320], Fig. \ref{fig:Fig3} (b5)), failed to detect this class. 

\item \textbf{Point convolution significantly improves basic point-based networks}: For point-based methods, PAConv notably improved performance by integrating position-adaptive convolution operations into the PointNet++ backbone, achieving a 17.28 percentage point increase over PointNet++. PAConv also delivered competitive results on the HR3D, Pheno4D, and Pheno4D-RS datasets, with only slight decreases of 0.54, 1.45, and 0.71 percentage points, respectively, compared to the top-performing methods. This demonstrates that point convolution, which captures local geometry more effectively, can substantially enhance lightweight networks. 

\item \textbf{Comprehensive evaluation}: In terms of inference speed, SCUNet achieved the highest performance at 16.67 frames per second (FPS). On average, voxel-based methods (SCUNet, MinkUNet, and SPVConv) achieved 12.27 FPS, while point-based methods (PointNet++, DGCNN, and PAConv) averaged 8.22 FPS. Even MinkUNet, the slowest voxel-based method, achieved 9.36 FPS, which is adequate for most downstream applications. A more intuitive comparison of mIoU, inference speed, and network parameters is provided in Fig. \ref{fig:Fig8}. These findings suggest voxel-based methods are the optimal choice for integration with instance segmentation, given their advanced feature learning capabilities, faster processing speeds, and acceptable model sizes. 
\end{itemize}

\afterpage{%
    \clearpage
    \thispagestyle{landscapepage}
\begin{landscape}
\begin{table}
  \caption{\fontfamily{ptm}\selectfont Semantic segmentation results. The best results are in boldface.}
  \label{tab:Table6}
  \centering
  {\fontfamily{ptm}\selectfont
  \makebox[\textwidth][c]{%
    \begin{tabular}{l l *{19}{c}}
      \toprule
      Network &  & \multicolumn{2}{c}{COS} & \multicolumn{2}{c}{HR3D} & \multicolumn{2}{c}{SYAU-Maize} & \multicolumn{3}{c}{Pheno4D} & \multicolumn{3}{c}{SoyBeanMVS} & \multicolumn{3}{c}{Pheno4D-RS} & \multicolumn{3}{c}{SoyBeanMVS-RS} & Mean \\
      \cmidrule(lr){3-4} \cmidrule(lr){5-6} \cmidrule(lr){7-8} \cmidrule(lr){9-11} \cmidrule(lr){12-14} \cmidrule(lr){15-17} \cmidrule(lr){18-20}
      \text{(Params)}       &        & trunk & branch & stem & leaf & stem & leaf & ground & stem & leaf & mainstem & stem & leaf & ground & stem & leaf & mainstem & stem & leaf &  \\
      \midrule
      % --- PointNet++ group ---
      \multirow[t]{6}{*}{PointNet++} 
           & Prec & \multicolumn{1}{c}{93.27} & \multicolumn{1}{c}{84.93} & \multicolumn{1}{c}{74.62} & \multicolumn{1}{c}{90.69} & \multicolumn{1}{c}{68.15} & \multicolumn{1}{c}{97.83} & \multicolumn{1}{c}{96.13} & \multicolumn{1}{c}{81.43} & \multicolumn{1}{c}{90.22} & \multicolumn{1}{c}{--} & \multicolumn{1}{c}{48.97} & \multicolumn{1}{c}{86.57} & \multicolumn{1}{c}{96.87} & \multicolumn{1}{c}{80.33} & \multicolumn{1}{c}{88.11} & \multicolumn{1}{c}{--} & \multicolumn{1}{c}{49.09} & \multicolumn{1}{c}{83.95} & \multicolumn{1}{c}{} \\ \multicolumn{1}{l}{(1.88M)}    
           & Rec & \multicolumn{1}{c}{92.21} & \multicolumn{1}{c}{86.83} & \multicolumn{1}{c}{62.27} & \multicolumn{1}{c}{94.55} & \multicolumn{1}{c}{71.08} & \multicolumn{1}{c}{97.52} & \multicolumn{1}{c}{99.53} & \multicolumn{1}{c}{60.96} & \multicolumn{1}{c}{92.82} & \multicolumn{1}{c}{--} & \multicolumn{1}{c}{49.64} & \multicolumn{1}{c}{88.98} & \multicolumn{1}{c}{99.54} & \multicolumn{1}{c}{62.12} & \multicolumn{1}{c}{92.33} & \multicolumn{1}{c}{--} & \multicolumn{1}{c}{42.24} & \multicolumn{1}{c}{91.05} & \multicolumn{1}{c}{} \\
           & F1 & \multicolumn{1}{c}{92.74} & \multicolumn{1}{c}{85.87} & \multicolumn{1}{c}{67.89} & \multicolumn{1}{c}{92.58} & \multicolumn{1}{c}{69.58} & \multicolumn{1}{c}{97.67} & \multicolumn{1}{c}{97.80} & \multicolumn{1}{c}{69.72} & \multicolumn{1}{c}{91.50} & \multicolumn{1}{c}{--} & \multicolumn{1}{c}{49.30} & \multicolumn{1}{c}{87.76} & \multicolumn{1}{c}{98.19} & \multicolumn{1}{c}{70.06} & \multicolumn{1}{c}{90.17} & \multicolumn{1}{c}{--} & \multicolumn{1}{c}{45.41} & \multicolumn{1}{c}{87.36} & \multicolumn{1}{c}{} \\
           & IoU & \multicolumn{1}{c}{86.46} & \multicolumn{1}{c}{75.24} & \multicolumn{1}{c}{51.38} & \multicolumn{1}{c}{86.19} & \multicolumn{1}{c}{53.36} & \multicolumn{1}{c}{95.45} & \multicolumn{1}{c}{95.69} & \multicolumn{1}{c}{53.52} & \multicolumn{1}{c}{84.34} & \multicolumn{1}{c}{--} & \multicolumn{1}{c}{32.72} & \multicolumn{1}{c}{78.19} & \multicolumn{1}{c}{96.44} & \multicolumn{1}{c}{53.92} & \multicolumn{1}{c}{82.10} & \multicolumn{1}{c}{--} & \multicolumn{1}{c}{29.38} & \multicolumn{1}{c}{77.56} & \multicolumn{1}{c}{} \\
           & FPS & \multicolumn{2}{c}{8.46} & \multicolumn{2}{c}{37.53} & \multicolumn{2}{c}{40.78} & \multicolumn{3}{c}{1.26} & \multicolumn{3}{c}{1.23} & \multicolumn{3}{c}{1.81} & \multicolumn{3}{c}{2.6} & \multicolumn{1}{c}{13.38} \\
           & mIoU & \multicolumn{2}{c}{80.85} & \multicolumn{2}{c}{68.79} & \multicolumn{2}{c}{74.40} & \multicolumn{3}{c}{77.85} & \multicolumn{3}{c}{36.97} & \multicolumn{3}{c}{77.49} & \multicolumn{3}{c}{35.64} & \multicolumn{1}{c}{64.57} \\
      \midrule
      \addlinespace
      % --- DGCNN group ---
      \multirow[t]{6}{*}{DGCNN} 
           & Prec & \multicolumn{1}{c}{91.00} & \multicolumn{1}{c}{91.44} & \multicolumn{1}{c}{85.97} & \multicolumn{1}{c}{95.08} & \multicolumn{1}{c}{93.10} & \multicolumn{1}{c}{99.13} & \multicolumn{1}{c}{99.80} & \multicolumn{1}{c}{86.70} & \multicolumn{1}{c}{94.32} & \multicolumn{1}{c}{54.42} & \multicolumn{1}{c}{66.20} & \multicolumn{1}{c}{88.89} & \multicolumn{1}{c}{99.76} & \multicolumn{1}{c}{90.73} & \multicolumn{1}{c}{93.06} & \multicolumn{1}{c}{65.30} & \multicolumn{1}{c}{69.51} & \multicolumn{1}{c}{85.20} & \multicolumn{1}{c}{} \\ \multicolumn{1}{l}{(0.96M)}        
           & Rec & \multicolumn{1}{c}{96.16} & \multicolumn{1}{c}{81.19} & \multicolumn{1}{c}{80.57} & \multicolumn{1}{c}{96.62} & \multicolumn{1}{c}{88.35} & \multicolumn{1}{c}{99.51} & \multicolumn{1}{c}{99.77} & \multicolumn{1}{c}{79.93} & \multicolumn{1}{c}{96.56} & \multicolumn{1}{c}{30.27} & \multicolumn{1}{c}{56.51} & \multicolumn{1}{c}{93.40} & \multicolumn{1}{c}{99.89} & \multicolumn{1}{c}{79.91} & \multicolumn{1}{c}{97.06} & \multicolumn{1}{c}{17.92} & \multicolumn{1}{c}{49.04} & \multicolumn{1}{c}{94.85} & \multicolumn{1}{c}{} \\
           & F1 & \multicolumn{1}{c}{93.51} & \multicolumn{1}{c}{86.01} & \multicolumn{1}{c}{83.19} & \multicolumn{1}{c}{95.85} & \multicolumn{1}{c}{90.66} & \multicolumn{1}{c}{99.32} & \multicolumn{1}{c}{99.78} & \multicolumn{1}{c}{83.17} & \multicolumn{1}{c}{95.42} & \multicolumn{1}{c}{38.90} & \multicolumn{1}{c}{60.97} & \multicolumn{1}{c}{91.09} & \multicolumn{1}{c}{99.82} & \multicolumn{1}{c}{84.98} & \multicolumn{1}{c}{95.02} & \multicolumn{1}{c}{28.12} & \multicolumn{1}{c}{57.50} & \multicolumn{1}{c}{89.76} & \multicolumn{1}{c}{} \\
           & IoU & \multicolumn{1}{c}{87.81} & \multicolumn{1}{c}{75.46} & \multicolumn{1}{c}{71.21} & \multicolumn{1}{c}{92.02} & \multicolumn{1}{c}{82.92} & \multicolumn{1}{c}{98.65} & \multicolumn{1}{c}{99.57} & \multicolumn{1}{c}{71.19} & \multicolumn{1}{c}{91.25} & \multicolumn{1}{c}{24.15} & \multicolumn{1}{c}{43.85} & \multicolumn{1}{c}{83.64} & \multicolumn{1}{c}{99.65} & \multicolumn{1}{c}{73.88} & \multicolumn{1}{c}{90.51} & \multicolumn{1}{c}{16.36} & \multicolumn{1}{c}{40.35} & \multicolumn{1}{c}{81.43} & \multicolumn{1}{c}{} \\
           & FPS & \multicolumn{2}{c}{1.91} & \multicolumn{2}{c}{11.61} & \multicolumn{2}{c}{20.33} & \multicolumn{3}{c}{0.26} & \multicolumn{3}{c}{0.26} & \multicolumn{3}{c}{0.4} & \multicolumn{3}{c}{0.53} & \multicolumn{1}{c}{5.04} \\
           & mIoU & \multicolumn{2}{c}{81.63} & \multicolumn{2}{c}{81.62} & \multicolumn{2}{c}{90.78} & \multicolumn{3}{c}{87.34} & \multicolumn{3}{c}{50.55} & \multicolumn{3}{c}{88.01} & \multicolumn{3}{c}{46.05} & \multicolumn{1}{c}{75.14} \\
      \midrule
      \addlinespace
      % --- PAConv group ---
      \multirow[t]{6}{*}{PAConv} 
           & Prec & \multicolumn{1}{c}{95.17} & \multicolumn{1}{c}{91.85} & \multicolumn{1}{c}{93.65} & \multicolumn{1}{c}{98.20} & \multicolumn{1}{c}{94.81} & \multicolumn{1}{c}{99.35} & \multicolumn{1}{c}{99.93} & \multicolumn{1}{c}{95.74} & \multicolumn{1}{c}{97.09} & \multicolumn{1}{c}{73.99} & \multicolumn{1}{c}{75.14} & \multicolumn{1}{c}{92.32} & \multicolumn{1}{c}{99.87} & \multicolumn{1}{c}{96.27} & \multicolumn{1}{c}{97.44} & \multicolumn{1}{c}{81.66} & \multicolumn{1}{c}{78.43} & \multicolumn{1}{c}{91.11} & \multicolumn{1}{c}{} \\ \multicolumn{1}{l}{(11.78M)}        
           & Rec & \multicolumn{1}{c}{95.95} & \multicolumn{1}{c}{90.37} & \multicolumn{1}{c}{92.98} & \multicolumn{1}{c}{98.38} & \multicolumn{1}{c}{91.32} & \multicolumn{1}{c}{99.63} & \multicolumn{1}{c}{99.93} & \multicolumn{1}{c}{89.92} & \multicolumn{1}{c}{98.86} & \multicolumn{1}{c}{24.15} & \multicolumn{1}{c}{72.34} & \multicolumn{1}{c}{95.14} & \multicolumn{1}{c}{99.94} & \multicolumn{1}{c}{92.70} & \multicolumn{1}{c}{98.72} & \multicolumn{1}{c}{17.05} & \multicolumn{1}{c}{71.33} & \multicolumn{1}{c}{96.72} & \multicolumn{1}{c}{} \\
           & F1 & \multicolumn{1}{c}{95.56} & \multicolumn{1}{c}{91.11} & \multicolumn{1}{c}{93.31} & \multicolumn{1}{c}{98.29} & \multicolumn{1}{c}{93.03} & \multicolumn{1}{c}{99.49} & \multicolumn{1}{c}{99.93} & \multicolumn{1}{c}{92.74} & \multicolumn{1}{c}{97.97} & \multicolumn{1}{c}{36.42} & \multicolumn{1}{c}{73.71} & \multicolumn{1}{c}{93.71} & \multicolumn{1}{c}{99.90} & \multicolumn{1}{c}{94.45} & \multicolumn{1}{c}{98.08} & \multicolumn{1}{c}{28.21} & \multicolumn{1}{c}{74.71} & \multicolumn{1}{c}{93.83} & \multicolumn{1}{c}{} \\
           & IoU & \multicolumn{1}{c}{91.49} & \multicolumn{1}{c}{83.66} & \multicolumn{1}{c}{87.46} & \multicolumn{1}{c}{96.63} & \multicolumn{1}{c}{86.97} & \multicolumn{1}{c}{98.98} & \multicolumn{1}{c}{99.86} & \multicolumn{1}{c}{86.46} & \multicolumn{1}{c}{96.02} & \multicolumn{1}{c}{22.26} & \multicolumn{1}{c}{58.37} & \multicolumn{1}{c}{88.16} & \multicolumn{1}{c}{99.81} & \multicolumn{1}{c}{89.49} & \multicolumn{1}{c}{96.22} & \multicolumn{1}{c}{16.42} & \multicolumn{1}{c}{59.63} & \multicolumn{1}{c}{88.38} & \multicolumn{1}{c}{} \\
           & FPS & \multicolumn{2}{c}{3.03} & \multicolumn{2}{c}{15.48} & \multicolumn{2}{c}{22.95} & \multicolumn{3}{c}{0.42} & \multicolumn{3}{c}{0.41} & \multicolumn{3}{c}{0.64} & \multicolumn{3}{c}{0.85} & \multicolumn{1}{c}{6.25} \\
           & mIoU & \multicolumn{2}{c}{87.58} & \multicolumn{2}{c}{92.05} & \multicolumn{2}{c}{92.98} & \multicolumn{3}{c}{94.11} & \multicolumn{3}{c}{56.27} & \multicolumn{3}{c}{\underline{95.18}} & \multicolumn{3}{c}{54.81} & \multicolumn{1}{c}{81.85} \\
      \midrule     
      \addlinespace
    % --- SCUNet group ---
    \multirow[t]{6}{*}{SCUNet} 
          & Prec & \multicolumn{1}{c}{97.91} & \multicolumn{1}{c}{89.69} & \multicolumn{1}{c}{93.81} & \multicolumn{1}{c}{98.08} & \multicolumn{1}{c}{96.32} & \multicolumn{1}{c}{99.59} & \multicolumn{1}{c}{99.91} & \multicolumn{1}{c}{92.61} & \multicolumn{1}{c}{97.79} & \multicolumn{1}{c}{--} & \multicolumn{1}{c}{72.68} & \multicolumn{1}{c}{93.70} & \multicolumn{1}{c}{99.78} & \multicolumn{1}{c}{95.15} & \multicolumn{1}{c}{96.98} & \multicolumn{1}{c}{--} & \multicolumn{1}{c}{74.70} & \multicolumn{1}{c}{91.72} & \multicolumn{1}{c}{} \\ \multicolumn{1}{l}{(43.81M)}      
          & Rec & \multicolumn{1}{c}{94.42} & \multicolumn{1}{c}{96.01} & \multicolumn{1}{c}{92.51} & \multicolumn{1}{c}{98.43} & \multicolumn{1}{c}{94.52} & \multicolumn{1}{c}{99.73} & \multicolumn{1}{c}{99.97} & \multicolumn{1}{c}{92.40} & \multicolumn{1}{c}{97.79} & \multicolumn{1}{c}{--} & \multicolumn{1}{c}{80.13} & \multicolumn{1}{c}{94.23} & \multicolumn{1}{c}{99.90} & \multicolumn{1}{c}{91.77} & \multicolumn{1}{c}{98.12} & \multicolumn{1}{c}{--} & \multicolumn{1}{c}{79.00} & \multicolumn{1}{c}{94.62} & \multicolumn{1}{c}{} \\
          & F1 & \multicolumn{1}{c}{96.13} & \multicolumn{1}{c}{92.74} & \multicolumn{1}{c}{93.15} & \multicolumn{1}{c}{98.25} & \multicolumn{1}{c}{95.42} & \multicolumn{1}{c}{99.66} & \multicolumn{1}{c}{99.94} & \multicolumn{1}{c}{92.51} & \multicolumn{1}{c}{97.79} & \multicolumn{1}{c}{--} & \multicolumn{1}{c}{76.22} & \multicolumn{1}{c}{93.96} & \multicolumn{1}{c}{99.84} & \multicolumn{1}{c}{93.43} & \multicolumn{1}{c}{97.55} & \multicolumn{1}{c}{--} & \multicolumn{1}{c}{76.79} & \multicolumn{1}{c}{93.15} & \multicolumn{1}{c}{} \\
          & IoU & \multicolumn{1}{c}{92.55} & \multicolumn{1}{c}{86.47} & \multicolumn{1}{c}{87.18} & \multicolumn{1}{c}{96.57} & \multicolumn{1}{c}{91.23} & \multicolumn{1}{c}{99.32} & \multicolumn{1}{c}{99.88} & \multicolumn{1}{c}{86.06} & \multicolumn{1}{c}{95.68} & \multicolumn{1}{c}{--} & \multicolumn{1}{c}{61.58} & \multicolumn{1}{c}{88.62} & \multicolumn{1}{c}{99.68} & \multicolumn{1}{c}{87.67} & \multicolumn{1}{c}{95.21} & \multicolumn{1}{c}{--} & \multicolumn{1}{c}{62.32} & \multicolumn{1}{c}{87.18} & \multicolumn{1}{c}{} \\
          & FPS & \multicolumn{2}{c}{28.51} & \multicolumn{2}{c}{47.72} & \multicolumn{2}{c}{25.85} & \multicolumn{3}{c}{4.87} & \multicolumn{3}{c}{2.39} & \multicolumn{3}{c}{4.01} & \multicolumn{3}{c}{3.37} & \multicolumn{1}{c}{16.67} \\
          & mIoU & \multicolumn{2}{c}{89.51} & \multicolumn{2}{c}{91.88} & \multicolumn{2}{c}{\underline{95.28}} & \multicolumn{3}{c}{93.87} & \multicolumn{3}{c}{50.06} & \multicolumn{3}{c}{94.19} & \multicolumn{3}{c}{49.83} & \multicolumn{1}{c}{80.66} \\
    \midrule     
    \addlinespace
% --- MinkUnet group ---
    \multirow[t]{6}{*}{MinkUNet} 
          & Prec & \multicolumn{1}{c}{96.09} & \multicolumn{1}{c}{94.48} & \multicolumn{1}{c}{93.57} & \multicolumn{1}{c}{98.36} & \multicolumn{1}{c}{97.25} & \multicolumn{1}{c}{99.46} & \multicolumn{1}{c}{99.96} & \multicolumn{1}{c}{95.80} & \multicolumn{1}{c}{98.08} & \multicolumn{1}{c}{56.94} & \multicolumn{1}{c}{81.61} & \multicolumn{1}{c}{94.97} & \multicolumn{1}{c}{99.96} & \multicolumn{1}{c}{97.30} & \multicolumn{1}{c}{97.58} & \multicolumn{1}{c}{59.58} & \multicolumn{1}{c}{83.07} & \multicolumn{1}{c}{94.81} & \multicolumn{1}{c}{} \\ \multicolumn{1}{l}{(37.87M)}      
          & Rec & \multicolumn{1}{c}{97.28} & \multicolumn{1}{c}{92.16} & \multicolumn{1}{c}{93.62} & \multicolumn{1}{c}{98.35} & \multicolumn{1}{c}{92.73} & \multicolumn{1}{c}{99.80} & \multicolumn{1}{c}{99.96} & \multicolumn{1}{c}{93.39} & \multicolumn{1}{c}{98.82} & \multicolumn{1}{c}{72.49} & \multicolumn{1}{c}{76.29} & \multicolumn{1}{c}{95.71} & \multicolumn{1}{c}{99.90} & \multicolumn{1}{c}{93.41} & \multicolumn{1}{c}{99.12} & \multicolumn{1}{c}{70.35} & \multicolumn{1}{c}{77.68} & \multicolumn{1}{c}{95.66} & \multicolumn{1}{c}{} \\
          & F1 & \multicolumn{1}{c}{96.68} & \multicolumn{1}{c}{93.31} & \multicolumn{1}{c}{93.59} & \multicolumn{1}{c}{98.35} & \multicolumn{1}{c}{94.93} & \multicolumn{1}{c}{99.63} & \multicolumn{1}{c}{99.96} & \multicolumn{1}{c}{94.58} & \multicolumn{1}{c}{98.45} & \multicolumn{1}{c}{63.78} & \multicolumn{1}{c}{78.86} & \multicolumn{1}{c}{95.34} & \multicolumn{1}{c}{99.93} & \multicolumn{1}{c}{95.32} & \multicolumn{1}{c}{98.35} & \multicolumn{1}{c}{64.52} & \multicolumn{1}{c}{80.28} & \multicolumn{1}{c}{95.23} & \multicolumn{1}{c}{} \\
          & IoU & \multicolumn{1}{c}{93.57} & \multicolumn{1}{c}{87.45} & \multicolumn{1}{c}{87.96} & \multicolumn{1}{c}{96.76} & \multicolumn{1}{c}{90.36} & \multicolumn{1}{c}{99.26} & \multicolumn{1}{c}{99.91} & \multicolumn{1}{c}{89.72} & \multicolumn{1}{c}{96.95} & \multicolumn{1}{c}{46.82} & \multicolumn{1}{c}{65.10} & \multicolumn{1}{c}{91.09} & \multicolumn{1}{c}{99.86} & \multicolumn{1}{c}{91.05} & \multicolumn{1}{c}{96.75} & \multicolumn{1}{c}{47.62} & \multicolumn{1}{c}{67.06} & \multicolumn{1}{c}{90.90} & \multicolumn{1}{c}{} \\
          & FPS & \multicolumn{2}{c}{15.06} & \multicolumn{2}{c}{19.62} & \multicolumn{2}{c}{17.74} & \multicolumn{3}{c}{4.33} & \multicolumn{3}{c}{2.21} & \multicolumn{3}{c}{3.51} & \multicolumn{3}{c}{3.03} & \multicolumn{1}{c}{9.36} \\
          & mIoU & \multicolumn{2}{c}{\textbf{90.51}} & \multicolumn{2}{c}{\underline{92.36}} & \multicolumn{2}{c}{94.81} & \multicolumn{3}{c}{\underline{95.52}} & \multicolumn{3}{c}{\textbf{67.67}} & \multicolumn{3}{c}{\textbf{95.89}} & \multicolumn{3}{c}{\underline{68.53}} & \multicolumn{1}{c}{\underline{86.47}} \\
    \midrule    
    \addlinespace
% --- SPVConv group ---           
    \multirow[t]{6}{*}{SPVConv} 
        & Prec & \multicolumn{1}{c}{96.98} & \multicolumn{1}{c}{92.15} & \multicolumn{1}{c}{94.51} & \multicolumn{1}{c}{98.23} & \multicolumn{1}{c}{97.23} & \multicolumn{1}{c}{99.56} & \multicolumn{1}{c}{99.93} & \multicolumn{1}{c}{94.94} & \multicolumn{1}{c}{98.38} & \multicolumn{1}{c}{61.83} & \multicolumn{1}{c}{87.11} & \multicolumn{1}{c}{93.65} & \multicolumn{1}{c}{99.97} & \multicolumn{1}{c}{95.71} & \multicolumn{1}{c}{97.04} & \multicolumn{1}{c}{63.69} & \multicolumn{1}{c}{84.97} & \multicolumn{1}{c}{95.20} & \multicolumn{1}{c}{} \\ \multicolumn{1}{l}{(21.78M)}      
        & Rec & \multicolumn{1}{c}{95.95} & \multicolumn{1}{c}{94.09} & \multicolumn{1}{c}{93.07} & \multicolumn{1}{c}{98.61} & \multicolumn{1}{c}{94.06} & \multicolumn{1}{c}{99.80} & \multicolumn{1}{c}{99.97} & \multicolumn{1}{c}{94.34} & \multicolumn{1}{c}{98.51} & \multicolumn{1}{c}{63.28} & \multicolumn{1}{c}{71.92} & \multicolumn{1}{c}{97.68} & \multicolumn{1}{c}{99.83} & \multicolumn{1}{c}{92.19} & \multicolumn{1}{c}{98.58} & \multicolumn{1}{c}{71.70} & \multicolumn{1}{c}{79.59} & \multicolumn{1}{c}{96.26} & \multicolumn{1}{c}{} \\
        & F1 & \multicolumn{1}{c}{96.46} & \multicolumn{1}{c}{93.11} & \multicolumn{1}{c}{93.78} & \multicolumn{1}{c}{98.42} & \multicolumn{1}{c}{95.62} & \multicolumn{1}{c}{99.68} & \multicolumn{1}{c}{99.95} & \multicolumn{1}{c}{94.64} & \multicolumn{1}{c}{98.45} & \multicolumn{1}{c}{62.55} & \multicolumn{1}{c}{78.79} & \multicolumn{1}{c}{95.62} & \multicolumn{1}{c}{99.90} & \multicolumn{1}{c}{93.92} & \multicolumn{1}{c}{97.80} & \multicolumn{1}{c}{67.46} & \multicolumn{1}{c}{82.19} & \multicolumn{1}{c}{95.73} & \multicolumn{1}{c}{} \\
        & IoU  & \multicolumn{1}{c}{93.16} & \multicolumn{1}{c}{87.10} & \multicolumn{1}{c}{88.30} & \multicolumn{1}{c}{96.88} & \multicolumn{1}{c}{91.60} & \multicolumn{1}{c}{99.36} & \multicolumn{1}{c}{99.90} & \multicolumn{1}{c}{89.82} & \multicolumn{1}{c}{96.94} & \multicolumn{1}{c}{45.51} & \multicolumn{1}{c}{65.00} & \multicolumn{1}{c}{91.61} & \multicolumn{1}{c}{99.80} & \multicolumn{1}{c}{88.54} & \multicolumn{1}{c}{95.70} & \multicolumn{1}{c}{50.90} & \multicolumn{1}{c}{69.77} & \multicolumn{1}{c}{91.81} & \multicolumn{1}{c}{} \\
        & FPS & \multicolumn{2}{c}{17.18} & \multicolumn{2}{c}{24.16} & \multicolumn{2}{c}{20.94} & \multicolumn{3}{c}{4.31} & \multicolumn{3}{c}{2.2} & \multicolumn{3}{c}{3.58} & \multicolumn{3}{c}{3.11} & \multicolumn{1}{c}{10.78} \\
        & mIoU & \multicolumn{2}{c}{\underline{90.13}} & \multicolumn{2}{c}{\textbf{92.59}} & \multicolumn{2}{c}{\textbf{95.48}} & \multicolumn{3}{c}{\textbf{95.56}} & \multicolumn{3}{c}{\underline{67.37}} & \multicolumn{3}{c}{94.68} & \multicolumn{3}{c}{\textbf{70.83}} & \multicolumn{1}{c}{\textbf{86.66}} \\
      % \addlinespace[ex]
      \bottomrule
    \end{tabular}
  }% end of makebox
  }
\end{table}
\end{landscape}
    \clearpage
}

\begin{figure*}
\begin{center}
\includegraphics[width=\textwidth]{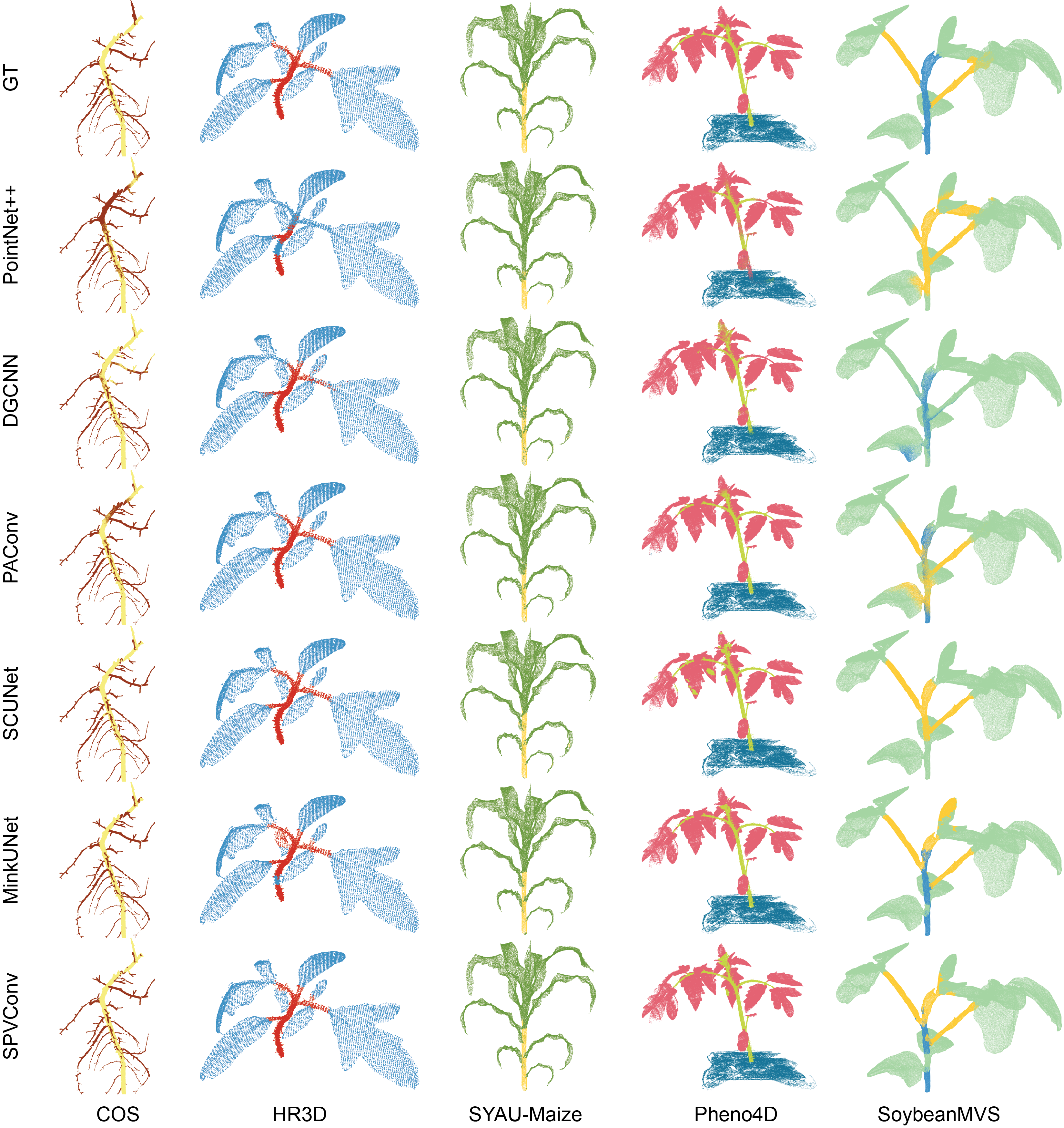}
    \caption{\fontfamily{ptm}\selectfont Qualitative results of semantic segmentation.}
    \label{fig:Fig9}
\end{center}
\end{figure*}

\subsubsection{Instance segmentation}\label{sec:Instance segmentation}

Based on the semantic results in Section \ref{sec:Semantic segmentation}, we selected the two top-performing semantic networks (MinkUNet and SPVConv) for integration with instance segmentation heads. The benchmark results of instance segmentation across 5 datasets are summarized in Table \ref{tab:Table7}. 

\begin{itemize}
    \item \textbf{Enhancing semantic heads improves overall performance}: Among all instance segmentation networks, SPVFormer achieved the best overall results, with AP25, AP50, and AP scores of 90.04\%, 79.11\%, and 64.48\%, respectively, demonstrating its ability to handle both easy and challenging segmentation requirements. MinkFormer also delivered a competitive performance, ranking second in AP50 and AP with scores of 77.35\% and 63.25\%. Both models outperformed the current state-of-the-art OneFormer3D by 1.76 and 0.53 percentage points on AP, respectively. These results demonstrate the advantage of combining advanced semantic heads with robust instance heads to further boost performance.

    The benefits are even more evident for grouping-based methods. Integrating MinkUNet and SPVConv with the SG instance head led to Mink-SG and SPVConv-SG surpassing the original SG by 17.16 and 12.38 percentage points on AP. Mink-SG also slightly outperformed all other models on the COS and HR3D datasets. Additionally, Mink-SG and SPVConv-SG achieved the highest AP on the SoybeanMVS-RS stem (38.14\%) and SoybeanMVS-RS leaf (40.88\%) classes, outperforming the best transformer-based method’s results (achieved by SPVFormer) by 7.55 and 4.22 percentage points, respectively. 

    \item \textbf{Transformer-based instance networks lead overall performance}: On average, transformer-based networks achieved an 11.62 percentage point improvement in AP compared to grouping-based methods. However, the strong feature representations provided by MinkUNet and SPVConv show that grouping-based approaches can still be optimized to perform competitively, especially on challenging datasets like SoybeanMVS-RS.

    \item \textbf{Instance networks improve semantic segmentation performance}: Table \ref{tab:Table7} also shows the changes in semantic mIoU when moving from semantic-only training to joint semantic-and-instance training for corresponding networks (e.g., MinkUNet vs. Mink-SG and MinkFormer). For SG and OneFormer3D, which both use a standard structured sparse convolution U-Net backbone, their semantic mIoU was compared with SCUNet. In most cases, mIoU improved with joint training, indicating that optimizing a network for the more complex task of instance segmentation enhances its ability to handle simpler tasks like semantic segmentation.
\end{itemize}

\begin{table*}[ht]
\definecolor{darkgreen}{RGB}{0,200,0}
\definecolor{darkred}{RGB}{200,0,0}
  \caption{\fontfamily{ptm}\selectfont Instance segmentation results. The best results are in boldface, and the second best results are underlined. The green and red numbers indicate the mIoU changes from semantic-only networks (Results in Table \ref{tab:Table6}) to instance networks with the corresponding semantic head (e.g., Mink-SG and MinkFormer compared to MinkUNet).}
  \label{tab:Table7}
  {\fontfamily{ptm}\selectfont
  \centering
  \setlength{\tabcolsep}{8pt}  % Increased column separation for more even distribution
  \begin{tabular}{p{2cm} p{2cm} *{8}{c}}  % Using *{8}{c} for uniform numeric columns
    \toprule
    Network &  & COS  & HR3D & SYAU-Maize & \multicolumn{2}{c}{Pheno4D-RS} & \multicolumn{2}{c}{SoyBeanMVS-RS} & Mean \\
    \cmidrule(lr){3-3} \cmidrule(lr){4-4} \cmidrule(lr){5-5} \cmidrule(lr){6-7} \cmidrule(lr){8-9}
    & & branch & leaf & leaf & stem & leaf & stem & leaf & \\
    \midrule
    \multirow[t]{5}{*}{SoftGroup}
      & Ap25 & 77.25  & 89.54 & 92.47 & 52.06 & 87.69 & 59.05 & 49.52 & 72.51 \\
      & Ap50 & 47.50  & 85.82 & 87.47 & 39.20 & 80.60 & 43.13 & 27.62 & 58.76 \\
      & Ap   & 23.52 & 74.33 & 68.91 & 14.99 & 67.84 & 25.50 & 19.03 & 42.02 \\
      & mIoU & 91.48  & 90.13   & 89.87   & \multicolumn{2}{c}{89.69} & \multicolumn{2}{c}{54.46} & 83.13 \\
      &Compared to SCUNet& \textcolor{darkgreen}{(+1.97)} & \textcolor{darkred}{(-1.75)} & \textcolor{darkred}{(-5.41)} & \multicolumn{2}{c}{\textcolor{darkred}{(-4.50)}} & \multicolumn{2}{c}{\textcolor{darkgreen}{(+4.63)}} & \textcolor{darkred}{(-1.01)} \\
      \midrule
    \addlinespace
    
    \multirow[t]{5}{*}{Mink-SG}
      & Ap25 & \underline{87.60}  & 91.71 & 97.07 & \underline{96.56} & \textbf{90.48} & 72.89 & 70.13 & 86.63 \\
      & Ap50 & \textbf{79.62} & 89.74 & 95.45 & 81.98 & 82.90 & \textbf{55.89} & 49.75 & 76.48 \\
      & Ap   & \textbf{56.42} & \textbf{81.72} & 86.97 & 44.29 & 67.57 & \textbf{38.14} & \underline{39.12} & 59.18 \\
      & mIoU & 92.20 & 92.81 & 94.37 & \multicolumn{2}{c}{95.28} & \multicolumn{2}{c}{73.89} & 89.71 \\
      &Compared to MinkUNet& \textcolor{darkgreen}{(+1.69)} & \textcolor{darkgreen}{(+0.45)} & \textcolor{darkred}{(-0.44)} & \multicolumn{2}{c}{\textcolor{darkred}{(-0.61	)}} & \multicolumn{2}{c}{\textcolor{darkgreen}{(+5.36)}} & \textcolor{darkgreen}{(+1.29)} \\
      \midrule
    \addlinespace
    
    \multirow[t]{5}{*}{SPVConv-SG}
      & Ap25 & 86.74 & 91.13 & 91.91 & 91.01 & 88.97 & 67.42 & 72.81 & 84.28 \\
      & Ap50 & 74.85 & 88.85 & 83.98 & 69.87 & \textbf{84.75} & 50.23 & \textbf{53.30} & 72.26 \\
      & AP   & 52.02 & 79.83 & 74.23 & 30.44 & 67.85 & \underline{35.58} & \textbf{40.88} & 54.40 \\
      & mIoU & 91.91 & 92.74 & 94.33 & \multicolumn{2}{c}{95.76} & \multicolumn{2}{c}{72.80} & 89.51 \\
      &Compared to SPVConv & \textcolor{darkgreen}{(+1.78)} & \textcolor{darkgreen}{(+0.15)} & \textcolor{darkred}{(-1.15)} & \multicolumn{2}{c}{\textcolor{darkgreen}{(+1.08)}} & \multicolumn{2}{c}{\textcolor{darkgreen}{(+1.97)}} & \textcolor{darkgreen}{(+0.77)} \\   
      \midrule
    \addlinespace
    
    \multirow[t]{5}{*}{OneFormer3D}
      & Ap25 & 86.21 & \textbf{95.41} & \underline{98.97} & \textbf{100} & 89.21 & 75.32 & \textbf{82.44} & \underline{89.65} \\
      & Ap50 & 74.96 & \underline{91.64} & \underline{97.65} & \underline{97.33} & 82.02 & 43.91 & 51.01 & 76.93 \\
      & AP   & 54.53 & \underline{81.09} & \underline{92.74} & 82.27 & 70.13 & 24.03 & 34.27 & 62.72 \\
      & mIoU & 92.19 & 94.12 & 96.12 & \multicolumn{2}{c}{96.43} & \multicolumn{2}{c}{77.51} & 91.27 \\
      &Compared to SCUNet& \textcolor{darkgreen}{(+2.68)} & \textcolor{darkgreen}{(+2.24)} & \textcolor{darkgreen}{(+0.84)} & \multicolumn{2}{c}{\textcolor{darkgreen}{(+2.24)}} & \multicolumn{2}{c}{\textcolor{darkgreen}{(+27.68)}} & \textcolor{darkgreen}{(+7.14)} \\ 
      \midrule
    \addlinespace
    
    \multirow[t]{5}{*}{MinkFormer}
      & Ap25 & 86.03 & 95.12 & \textbf{99.01} & \textbf{100} & 87.88 & \underline{75.35} & 73.83 & 88.17 \\
      & Ap50 & 74.12 & \textbf{91.78} & \textbf{97.85} & \textbf{100} & 84.11 & 43.15 & 50.47 & \underline{77.35} \\
      & AP   & 50.62 & \underline{81.35} & \textbf{94.06} & \textbf{85.68} & \textbf{71.17} & 25.12 & 34.74 & \underline{63.25} \\
      & mIoU & 91.71 & 93.99 & 95.70 & \multicolumn{2}{c}{97.57} & \multicolumn{2}{c}{76.71} & 91.14 \\
      &Compared to MinkUNet& \textcolor{darkgreen}{(+1.2)} & \textcolor{darkgreen}{(+1.63)} & \textcolor{darkgreen}{(+0.89)} & \multicolumn{2}{c}{\textcolor{darkgreen}{(+1.68)}} & \multicolumn{2}{c}{\textcolor{darkgreen}{(+8.18)}} & \textcolor{darkgreen}{(+2.72)} \\ 
      \midrule
    \addlinespace
    
    \multirow[t]{5}{*}{SPVFormer}
      & Ap25 & \textbf{88.02} & \underline{95.39} & 98.94 & \textbf{100} & \underline{89.91} & \textbf{79.52} & \underline{78.48} & \textbf{90.04} \\
      & Ap50 & \underline{76.63} & 91.36 & 97.56 & \textbf{100} & \underline{84.27} & \underline{51.03} & \underline{52.94} & \textbf{79.11} \\
      & AP   & \underline{55.49} & 80.71 & 92.41 & \underline{85.12} & \underline{70.41} & 30.59 & 36.66 & \textbf{64.48} \\
      & mIoU & 91.76 & 94.05 & 96.08 & \multicolumn{2}{c}{97.75} & \multicolumn{2}{c}{76.46} & 91.22 \\
      &Compared to SPVConv& \textcolor{darkgreen}{(+1.63)} & \textcolor{darkgreen}{(+1.46)} & \textcolor{darkgreen}{(+0.6)} & \multicolumn{2}{c}{\textcolor{darkgreen}{(+3.07)}} & \multicolumn{2}{c}{\textcolor{darkgreen}{(+5.63)}} & \textcolor{darkgreen}{(+2.48)} \\
    \bottomrule
  \end{tabular}
  }
\end{table*}

\begin{figure*}
\begin{center}
\includegraphics[width=\textwidth]{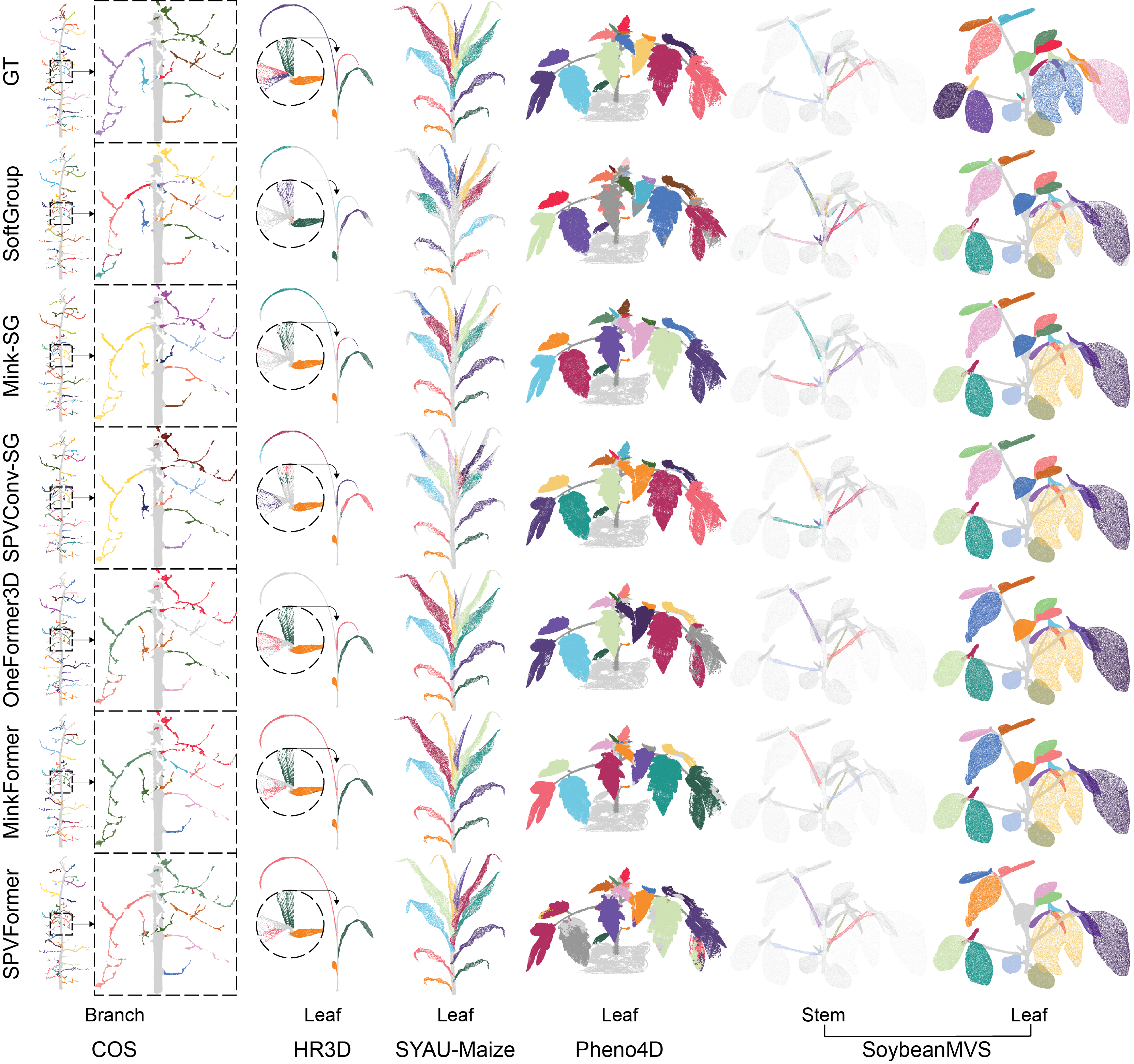}
    \caption{\fontfamily{ptm}\selectfont Qualitative results of instance segmentation.}
    \label{fig:Fig10}
\end{center}
\end{figure*}

\subsubsection{L-TreeGen and deformation for sim2real learning}\label{sec:0-shot test}

We chose SPVFormer to conduct all the experiments in this section given its best overall performance on instance segmentation (Table \ref{tab:Table7}).

\begin{itemize}

    \item \textbf{Robustness of L-TreeGen and deformation in 0-shot segmentation}: 
    Fig. \ref{fig:Fig11} presents 0-shot segmentation results on mIoU, AP25, AP50, and AP across 12 non-overlapping folds using L-TreeGen and deformation synthetic data for sim2real learning. The deformation-based achieved consistently higher mean values for all metrics. In particular, semantic mIoU and instance AP25 improved significantly by an average of 21.12 and 6.23 percentage points compared to L-TreeGen. 

    Regarding robustness, deformation-based segmentation exhibited greater stability in semantic mIoU, with a standard deviation (SD) of 0.88. Conversely, L-TreeGen-based segmentation demonstrated more stable instance-level performance, with SD reductions of 1.01, 3.82, and 2.34 on AP25, AP50, and AP, respectively, compared to the deformation. 

    These differences stem from the generation strategies: L-TreeGen uses the TG module to interpolate geometric features from base trees, producing continuous and complete shapes among all generated data, providing more stable branch instance geometries. In contrast, deformation preserves imperfections from sensor data (e.g., incompleteness or gaps of trunk or branch points) where the instance-level information is less robust (Fig. \ref{fig:Fig7}(b)). 
    
\begin{figure*}
\begin{center}
    \includegraphics[width=0.9\textwidth]{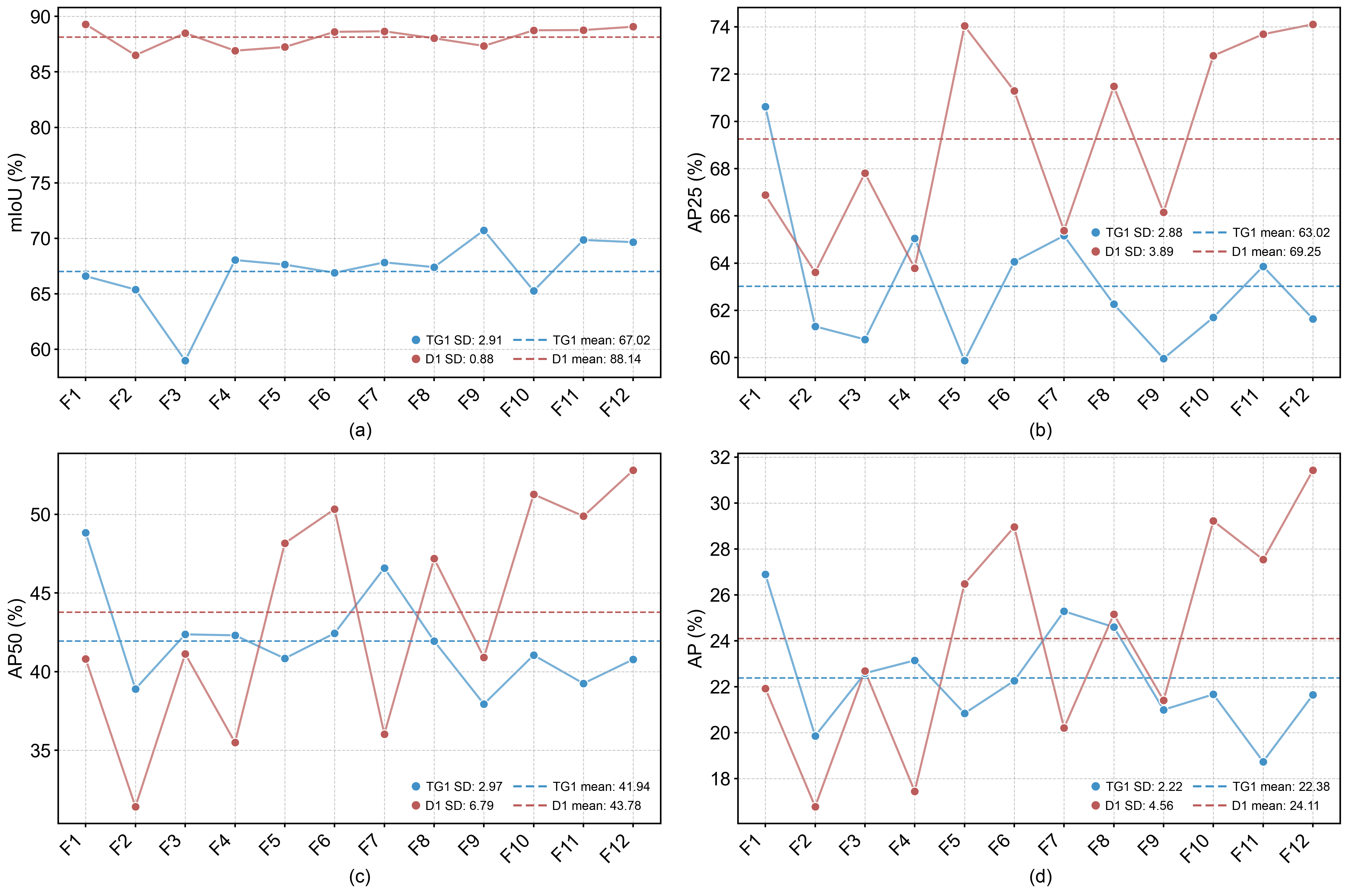}
    \caption{\fontfamily{ptm}\selectfont 0-shot performance of using TG (L-TreeGen) and deformation across 12 non-overlap folds on mIoU (a), AP25 (b), AP50 (c), and AP (d). For each fold, we use 6 base trees to generate synthetic data ($K_{b}=6$), denoted as TG1 and D1 from L-TreeGen-TG module, and deformation, respectively.}
    \label{fig:Fig11}
\end{center}
\end{figure*}

    \item \textbf{0-shot and few-shot performance under varying base tree sizes for synthetic data generation}: Fig. \ref{fig:Fig12} shows the lower (LB) and upper-bound (UB) of 0-shot and few-shot performance as the base tree size ($K_{b}$ = 6, 12, 18, 24) increases. Few-shot results ($K_{b}$-shot) are obtained by fine-tuning the network’s 0-shot weights with all base trees, providing insight into how base tree diversity affects performance under a fixed synthetic dataset size (150 trees).

    In general, 0-shot and few-shot segmentation performance improved with increasing $K_{b}$, and UB performance exceeded LB performance, as expected. However, several L-TreeGen-based 0-shot experiments revealed unexpected trends: at certain $K_{b}$/test set ratios (0.23, 0.46, 0.69), LB configurations achieved higher semantic performance than their UB counterparts. Besides, AP25, AP50, and AP fluctuated as $K_{b}$ increased, with performance at a 0.92 ratio matching that at 0.46, despite doubling the number of base trees. These findings indicate: 
    \begin{enumerate}[i)]
    \item Lower-performing base tree configurations may still provide superior results through feature interpolation. 
    \item L-TreeGen is highly sensitive to base tree composition, and simply adding more base trees does not guarantee linear performance improvements, revealing a quality-diversity trade-off. 
    \end{enumerate}  

    A key observation emerges when comparing pre and post-fine-tuning performance. For L-TreeGen-based, fine-tuning significantly boosted performance, while deformation-based improvements were modest. Table \ref{tab:Table8} lists detailed results, showing that although deformation-based 0-shot consistently outperformed L-TreeGen-based, $K_{b}$-shot fine-tuning enabled L-TreeGen-based to surpass deformation-based performance by a big margin, especially on instance-level metrics. Remarkably, using only 12 fine-tuning trees ($K_{b}=12$, and 0.46 of test size) achieved comparable results to the vanilla baseline (AP=55.49\% achieved by SPVFormer, Table \ref{tab:Table7}), which required 72 real trees. Increasing $K_{b}$ to 18 and 24 further outperformed this baseline. This demonstrates that using $K_{b}$-shot could reduce annotation effort by 83.33\% (from 72 to 12 real trees) while maintaining or exceeding baseline performance, underscoring the potential of PM-based sim2real learning for tackling real-world data scarcity challenges.

\begin{figure*}
\begin{center}
    \includegraphics[width=0.9\textwidth]{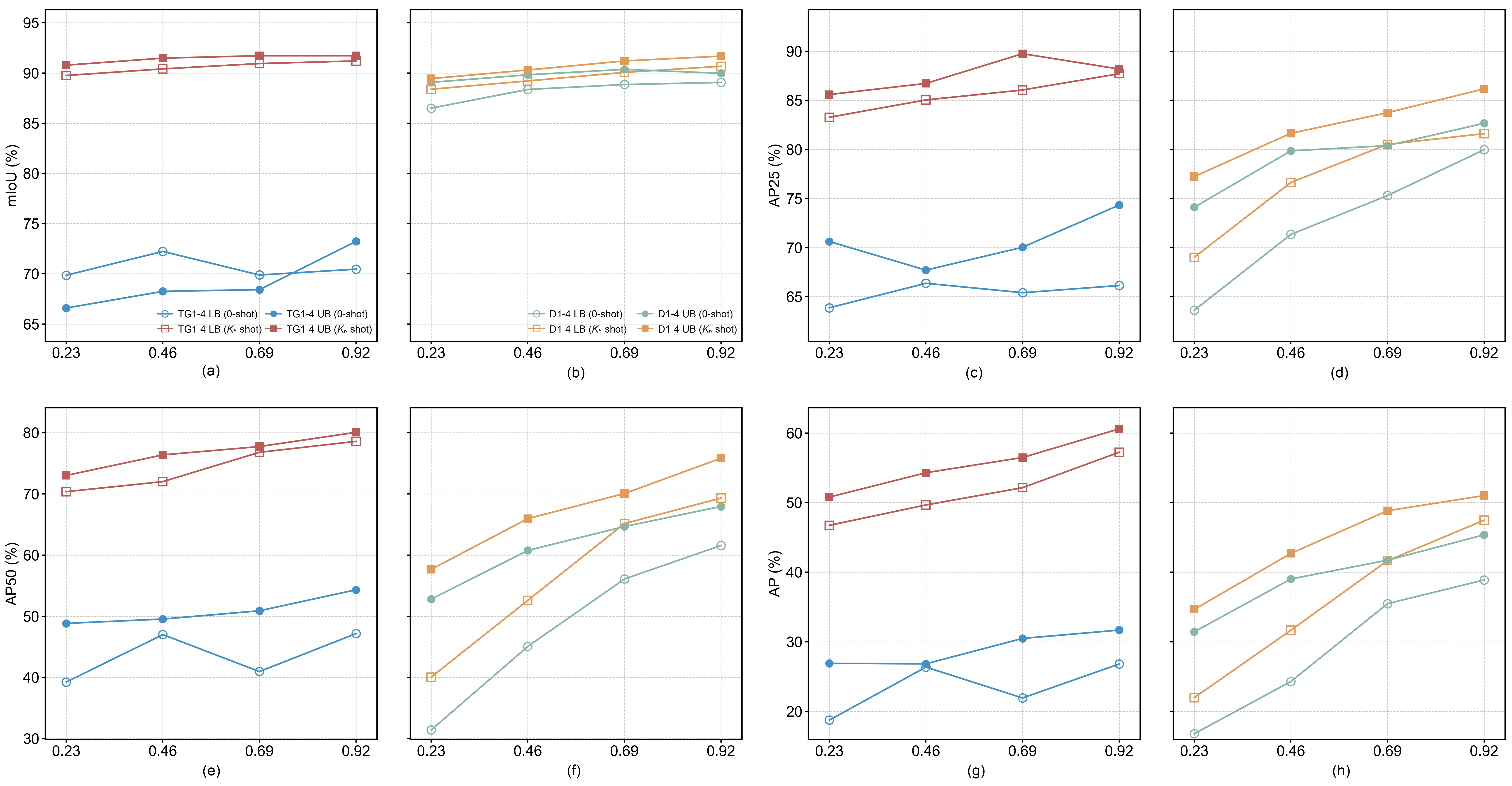}
    \caption{\fontfamily{ptm}\selectfont 0-shot and few-shot performance of using TG and deformation across different base tree size ($K_{b} =6,12,18,24$), denoted as TG1-4 and D1-4 on mIoU (a), (b), AP25 (c), (d), AP50 (e), (f), and AP (g), (h). Lower bound (LB) performance curves are marked with hollow scatters and upper bound (UB) curves are marked with solid scatters.}
    \label{fig:Fig12}
\end{center}
\end{figure*}
    
    \item \textbf{Resolution comparison for L-TreeGen-based sim2real learning}: Table \ref{tab:Table9} compares 0-shot and $K_{b}$-shot ($K_{b}=24$) performance for TG4 under three different VLS resolutions (VLS03, VLS006, VLS003). In 0-shot experiments, coarse (VLS03) and fine-grained (VLS003) scanning reduced instance-level performance, while VLS006 slightly improved AP50 and AP. After fine-tuning, all configurations achieved comparable results, with VLS006 and VLS003 slightly surpassing the TG4 baseline. These findings indicate that selecting an appropriate resolution is critical for effective sim2real learning, and both overly coarse or overly fine sensing simulations can hinder generalization. Besides. fine-tuning remains essential for maximizing performance in PM-based sim2real learning.
\end{itemize}

\begin{table*}
  \caption{\fontfamily{ptm}\selectfont The upper bound results of 0-shot and few-shot performance using TG and deformation with different number of base trees. We only used TG module from L-TreeGen in this experiment, and it's denoted as TG (TG1,2,3,4 refers to using $K_{b}=6,12,18,24$).}
  \label{tab:Table8}
  \centering
  {\fontfamily{ptm}\selectfont
  \setlength{\tabcolsep}{8pt}  % Increased column separation for more even distribution
  \begin{tabular}{c p{2cm} *{8}{c}}  % Using *{8}{c} for uniform numeric columns
    \toprule
    $K_{b}$ & Synthetic Gen. & \multicolumn{2}{c}{mIoU} & \multicolumn{2}{c}{AP25} & \multicolumn{2}{c}{AP50} & \multicolumn{2}{c}{AP} \\
    \cmidrule(lr){3-4} \cmidrule(lr){5-6} \cmidrule(lr){7-8} \cmidrule(lr){9-10}
    & & 0-shot & $K_{b}$-shot & 0-shot & $K_{b}$-shot & 0-shot & $K_{b}$-shot & 0-shot & $K_{b}$-shot \\
    \midrule
    \multirow[t]{2}{*}{6} 
      & TG1 & 66.59 & 90.78 & 70.62 & 85.60 & 48.84 & 73.02 & 26.90 & 50.79 \\
      & D1 & 89.07 & 89.44 & 74.11 & 77.24 & 52.80 & 57.67 & 31.43 & 34.66 \\
    \addlinespace[1.2ex]
    
    \multirow[t]{2}{*}{12}
      & TG2 & 68.25 & 91.48 & 67.70 & 86.73 & 49.55 & 76.38 & 26.83 & 54.29 \\
      & D2 & 89.82 & 90.30 & 79.86 & 81.65 & 60.76 & 65.96 & 39.02 & 42.70 \\
    \addlinespace[1.2ex]
    
    \multirow[t]{2}{*}{18}
      & TG3 & 68.42 & 91.72 & 70.03 & 89.76 & 50.90 & 77.74 & 30.48 & 56.48 \\
      & D3 & 90.36 & 91.20 & 80.39 & 83.75 & 64.67 & 70.06 & 41.72 & 48.83 \\
    \addlinespace[1.2ex]
    
    \multirow[t]{2}{*}{24}
      & TG4 & 73.23 & 91.72 & 74.33 & 88.20 & 54.33 & 80.07 & 31.68 & 60.59 \\
      & D4 & 89.97 & 91.68 & 82.67 & 86.18 & 67.95 & 75.83 & 45.37 & 51.01 \\
    \bottomrule
  \end{tabular}
  }
\end{table*}

\begin{table*}
  \caption{\fontfamily{ptm}\selectfont 0-shot and few-shot (with $K_{b}=24$) performance under different resolution using L-TreeGen (Using both TG and VLS modules).}
  \label{tab:Table9}
  \centering
  {\fontfamily{ptm}\selectfont
  \setlength{\tabcolsep}{8pt}  % Increased column separation for more even distribution
  \begin{tabular}{l *{8}{c}}  % Added spacing between metric groups
    \toprule
    Resolution & \multicolumn{2}{c}{mIoU} & \multicolumn{2}{c}{AP25} & \multicolumn{2}{c}{AP50} & \multicolumn{2}{c}{AP} \\
    \cmidrule(lr){2-3} \cmidrule(lr){4-5} \cmidrule(lr){6-7} \cmidrule(lr){8-9}
    & 0-shot & 24-shot & 0-shot & 24-shot & 0-shot & 24-shot & 0-shot & 24-shot \\
    \midrule
    TG4 & 73.23 & 91.72 & 74.33 & 88.20 & 54.33 & 80.07 & 31.68 & 60.59 \\
    TG4+VLS03 & 75.61 & 91.81 & 58.97 & 89.07 & 30.79 & 81.12 & 17.40 & 59.65 \\
    TG4+VLS006 & 70.96 & 91.91 & 73.01 & 90.78 & 54.49 & 81.45 & 32.86 & 61.99 \\
    TG4+VLS003 & 72.72 & 91.44 & 64.19 & 89.67 & 45.33 & 82.02 & 25.43 & 61.28 \\
    \bottomrule
  \end{tabular}
  }
\end{table*}

\section{Discussion}\label{sec:Discussion}

\subsection{Network design preferences}

Based on comprehensive benchmarks conducted across representative plant point cloud datasets, several network design choices have demonstrated superior performance and are recommended for point cloud segmentation in plant phenotyping. The following subsections summarize these preferences for both semantic and instance segmentation. 

\subsubsection{Semantic segmentation networks}

Our results highlight the superiority of voxel-based methods employing sparse convolutions, particularly SPVConv and MinkUNet. These networks excel at capturing multi-scale contextual information from complex yet dense plant point clouds while maintaining computational efficiency through optimized sparse convolution backends (e.g., Torchsparse, Minkowski Engine). Skip connections in these U-Net-like structures further enhance feature representation across multiple scales, which is critical for processing plants with intricate morphology. SPVConv and MinkUNet also provide a balanced trade-off between segmentation accuracy and computational speed, making them well-suited for diverse downstream tasks. Therefore, SPVConv or MinkUNet are preferred feature extraction backbones when high segmentation accuracy is required.

Although point-based methods exhibit relatively lower performance overall, they remain valuable in scenarios where granule voxelization cannot preserve fine geometric details critical for precise plant morphology characterization. PAConv demonstrated competitive results across several datasets, showing that advanced point convolutions can capture local geometry more effectively and enrich feature representation compared to classic point-MLPs-based backbones. Consequently, point convolution modules are a preferred choice when phenotyping tasks necessitate processing raw point clouds rather than using voxel representations. 

However, a key limitation of point-based networks is their incompatibility with many popular instance segmentation heads due to two inherent design characteristics: fixed input size and block-wise downsampling. Point-based networks require raw point clouds to be downsampled into fixed-size inputs to enable parallel processing with structures like MLPs. To retain geometric fidelity, they employ block-wise sampling, where data is downsampled to a predetermined point size within spatially constrained blocks. While this approach preserves local geometric details, it necessitates a sliding window inference strategy to process individual blocks sequentially and then aggregate the results. This workflow creates a substantial integration barrier with widely adopted voxel-based instance segmentation heads, which operate on dynamically voxelized, full-scale point clouds with direct point-to-voxel mappings.

\subsubsection{Instance segmentation networks}\label{sec:Discussion of instance networks}

Benchmark results indicate that instance segmentation performance is highly dependent on the quality of features extracted by the semantic backbone. Integrating top-performing semantic networks (e.g., SPVConv and MinkUNet) led to substantial improvements for both grouping-based and transformer-based approaches. Networks that combined transformer decoders with strong feature extractors (e.g., SPVFormer, MinkFormer) achieved slightly higher AP across all benchmark datasets, even surpassing leading networks like OneFormer3D. When SPVConv and MinkUNet were integrated into grouping-based methods (e.g., SG variants), the resulting models performed exceptionally well on SoybeanMVS-RS, the most challenging dataset due to its dense stems and leaves. This demonstrates that, for datasets with high object density or severe occlusions, the explicit geometric reasoning of grouping methods can be highly effective in improving segmentation accuracy.

Transformer-based networks generally require less hyperparameter tuning because they involve fewer dataset-specific settings, whereas grouping-based approaches rely heavily on hyperparameters related to point grouping, making careful tuning essential for optimal performance. Among these, the \textbf{grouping radius} ($Gr^{plant}$), the KNN radius threshold for grouping points, and the \textbf{class-specific point count threshold} ($Gnp^{plant}_{i}$, where ${i} \in$ instance classes), which determines whether a cluster contains enough points to be classified as a specific instance, are particularly influential. Given that the HR3D dataset includes a wider diversity of plant species and samples, we recommend initializing these parameters for new plant datasets using $Gr^{HR3D}$ and $Gnp^{HR3D}_{i}$ as references, scaling them proportionally to the mean instance size differences between datasets.

Additionally, semantic mIoU often improved when the semantic task was integrated in an instance network, suggesting that joint training with a powerful decoder enhances the encoder’s semantic understanding. A similar phenomenon was reported in \citep{kolodiazhnyi2024oneformer3d}, where the integration of both semantic and instance queries improved semantic segmentation performance compared to networks trained with semantic-only queries. These findings suggest that incorporating an additional instance segmentation task can mitigate overfitting when using transformer-decoder-based architectures for semantic-only tasks. Two strategies for improving semantic performance are therefore recommended: i) replacing simple linear layers with transformer decoders (architecture-level improvement) and ii) jointly training semantic and instance queries (training-level improvement).

In summary, for general-purpose instance segmentation targeting the best overall performance, transformer decoders (e.g., SPVFormer or MinkFormer) built on top of high-performing sparse convolution backbones (SPVConv or MinkUNet) are the most promising choice. For datasets with particularly dense, complex instances, enhancing grouping-based methods (e.g., SG) with these backbones offers a strong alternative, though careful tuning of key hyperparameters is essential. Lastly, Joint semantic and instance training is recommended in both cases to maximize performance.

\subsection{PM-based and augmentation-based synthetic training data generation for sim2real learning}

\subsubsection{0-shot performance}

The deformation approach, which directly transforms existing real-world point clouds, delivered superior 0-shot performance, particularly in semantic segmentation (mIoU). This result suggests that preserving data noise and sensor-specific imperfections is critical for direct sim2real learning without fine-tuning. This finding aligns with domain randomization principles, where introducing non-realistic variability in synthetic data helps bridge the sim2real reality gap effectively \citep{10.1109/IROS.2017.8202133}. While L-TreeGen-based demonstrated greater stability in 0-shot instance-level metric, likely due to its geometrically cleaner procedural model generation, it suffered from a larger initial reality gap, as L-system-based models may not replicate real-world sensor imperfections and textures. Overall, for 0-shot tasks, augmentation-based approaches not only provide stronger performance but are also generally more cost-effective to develop and deploy.  

\subsubsection{Few-shot performance and annotation efficiency}

A key finding is the remarkable effectiveness of L-TreeGen when combined with few-shot fine-tuning. Networks pre-trained on L-TreeGen data (only use TG module) and fine-tuned with a very small subset of real annotated data (as few as 12 trees, reducing annotation requirements by 83.33\% compared to vanilla setup) achieved performance comparable to or better than those trained entirely on much larger real datasets. This far exceeded the few-shot gains observed with deformation-based learning. These results indicate that although PM-based sim2real learning introduces a larger initial reality gap, it captures essential structural variability and semantics of the target objects, providing a rich foundation for rapid adaptation with minimal real-world supervision.  

\subsubsection{Base tree composition and VLS module in L-TreeGen}

L-TreeGen results revealed sensitivity to base tree composition, showing that simply increasing the number of base trees does not guarantee proportional performance gains. Striking a balance between base tree quality and diversity is crucial for PM-based modeling and sim2real learning with limited base tree size. Resolution comparisons further emphasized the importance of accurately simulating sensor effects in synthetic data to optimize 0-shot performance, as reducing the initial reality gap, while fine-tuning, remained essential to achieving peak results.  

For scenarios lacking robust PM-based synthesis tools or requiring immediate deployment without fine-tuning (pure 0-shot), deformation-based methods or brute-force generation pipelines emphasizing high fidelity and rapid dataset augmentation may be preferred due to their smaller initial reality gap. Conversely, when even a modest amount of annotated real data is available, pre-training on diverse, structurally rich PM-based synthetic data (e.g., L-TreeGen) followed by few-shot fine-tuning offers a highly efficient and effective strategy for achieving state-of-the-art segmentation performance.

\section{Conclusion}\label{sec:Conclusion}

This study provides a comprehensive review of deep learning (DL)-based 3D plant segmentation, integrating perspectives on datasets, algorithms, and computational workflows to address key barriers limiting its scalability and applicability. By systematically reviewing datasets, benchmarking state-of-the-art segmentation networks, and evaluating the value of synthetic data for sim2real learning, we deliver a clear conclusion of current progress while identifying technical challenges and opportunities for advancement. The introduction of the Plant Segmentation Studio (PSS) establishes a reproducible, extensible platform for community-driven benchmarking, helping to accelerate innovation in this domain.

Our findings highlight the efficacy of sparse convolutional backbones (e.g., SPVConv, MinkUNet) for generalized feature extraction across diverse plant point clouds, while identifying the promise and integration challenges of advanced point-based architectures (e.g., PAConv). Transformer-based instance segmentation demonstrates strong adaptability, whereas grouping-based methods exhibit notable potential for handling complex morphological structures. 

Complementary strategies for synthetic data generation further extend these advances: augmentation-based deformation techniques preserve sensing characteristics and are highly effective for 0-shot learning, whereas PM-based L-TreeGen introduces a larger initial reality gap but delivers superior results when used for pre-training and subsequent few-shot fine-tuning. Together, these methods illustrate a practical path toward reducing manual annotation and achieving scalable, high-precision segmentation.

In conclusion, this work bridges a critical gap between cutting-edge algorithmic research and its practical deployment in 3D plant phenotyping. The contributions not only provide immediate tools for domain researchers but also lay the groundwork for future exploration of data-efficient and generalizable DL solutions for 3D plant segmentation. We hope this study will catalyze community efforts to accelerate the success of AI-driven agriculture.

% Figure
% \begin{figure}%[]
%   \centering
% %    \includegraphics{}
%     \caption{}\label{fig1}
% \end{figure}

% \begin{table}%[]
% \caption{}\label{tbl1}
% \begin{tabular*}{\tblwidth}{@{}LL@{}}
% \toprule
%   &  \\ % Table header row
% \midrule
%  & \\
%  & \\
%  & \\
%  & \\
% \bottomrule
% \end{tabular*}
% \end{table}

% Uncomment and use as the case may be
%\begin{theorem} 
%\end{theorem}

% Uncomment and use as the case may be
%\begin{lemma} 
%\end{lemma}

%% The Appendices part is started with the command \appendix;
%% appendix sections are then done as normal sections
%% \appendix

% \section{}\label{}

% To print the credit authorship contribution details
\printcredits

%% Loading bibliography style file
%\bibliographystyle{model1-num-names}
\bibliographystyle{cas-model2-names-nodoiurl}

% Loading bibliography database
\bibliography{cas-refs}

\begin{thebibliography}{201}
\expandafter\ifx\csname natexlab\endcsname\relax\def\natexlab#1{#1}\fi
\providecommand{\url}[1]{\texttt{#1}}
\providecommand{\href}[2]{#2}
\providecommand{\path}[1]{#1}
\providecommand{\DOIprefix}{doi:}
\providecommand{\ArXivprefix}{arXiv:}
\providecommand{\URLprefix}{URL: }
\providecommand{\Pubmedprefix}{pmid:}
\providecommand{\doi}[1]{\href{http://dx.doi.org/#1}{\path{#1}}}
\providecommand{\Pubmed}[1]{\href{pmid:#1}{\path{#1}}}
\providecommand{\bibinfo}[2]{#2}
\ifx\xfnm\relax \def\xfnm[#1]{\unskip,\space#1}\fi
%Type = Article
\bibitem[{Ao et~al.(2022)Ao, Wu, Hu, Sun, Su, Guo and Xin}]{AO20221239}
\bibinfo{author}{Ao, Z.}, \bibinfo{author}{Wu, F.}, \bibinfo{author}{Hu, S.},
  \bibinfo{author}{Sun, Y.}, \bibinfo{author}{Su, Y.}, \bibinfo{author}{Guo,
  Q.}, \bibinfo{author}{Xin, Q.}, \bibinfo{year}{2022}.
\newblock \bibinfo{title}{Automatic segmentation of stem and leaf components
  and individual maize plants in field terrestrial lidar data using
  convolutional neural networks}.
\newblock \bibinfo{journal}{The Crop Journal}
\newblock \bibinfo{note}{Crop phenotyping studies with application to crop
  monitoring}.
%Type = Inproceedings
\bibitem[{Armeni et~al.(2016)Armeni, Sener, Zamir, Jiang, Brilakis, Fischer and
  Savarese}]{Armeni_2016_CVPR}
\bibinfo{author}{Armeni, I.}, \bibinfo{author}{Sener, O.},
  \bibinfo{author}{Zamir, A.R.}, \bibinfo{author}{Jiang, H.},
  \bibinfo{author}{Brilakis, I.}, \bibinfo{author}{Fischer, M.},
  \bibinfo{author}{Savarese, S.}, \bibinfo{year}{2016}.
\newblock \bibinfo{title}{3d semantic parsing of large-scale indoor spaces},
%Type = Article
\bibitem[{Atkinson et~al.(2019)Atkinson, Pound, Bennett and
  Wells}]{ATKINSON20191}
\bibinfo{author}{Atkinson, J.A.}, \bibinfo{author}{Pound, M.P.},
  \bibinfo{author}{Bennett, M.J.}, \bibinfo{author}{Wells, D.M.},
  \bibinfo{year}{2019}.
\newblock \bibinfo{title}{Uncovering the hidden half of plants using new
  advances in root phenotyping}.
\newblock \bibinfo{journal}{Current Opinion in Biotechnology}
\newblock \bibinfo{note}{Analytical Biotechnology}.
%Type = Inproceedings
\bibitem[{Baa et~al.(2020)Baa, Kontogianni and Leibe}]{DPCinproceedings}
\bibinfo{author}{Baa, A.}, \bibinfo{author}{Kontogianni, T.},
  \bibinfo{author}{Leibe, B.}, \bibinfo{year}{2020}.
\newblock \bibinfo{title}{Dilated point convolutions: On the receptive field
  size of point convolutions on 3d point clouds},
%Type = Article
\bibitem[{Bao et~al.(2019)Bao, Tang, Srinivasan and Schnable}]{BAO201986}
\bibinfo{author}{Bao, Y.}, \bibinfo{author}{Tang, L.},
  \bibinfo{author}{Srinivasan, S.}, \bibinfo{author}{Schnable, P.S.},
  \bibinfo{year}{2019}.
\newblock \bibinfo{title}{Field-based architectural traits characterisation of
  maize plant using time-of-flight 3d imaging}.
\newblock \bibinfo{journal}{Biosystems Engineering}
%Type = Article
\bibitem[{Behley et~al.(2021)Behley, Garbade, Milioto, Quenzel, Behnke, Gall
  and Stachniss}]{behley2021ijrr}
\bibinfo{author}{Behley, J.}, \bibinfo{author}{Garbade, M.},
  \bibinfo{author}{Milioto, A.}, \bibinfo{author}{Quenzel, J.},
  \bibinfo{author}{Behnke, S.}, \bibinfo{author}{Gall, J.},
  \bibinfo{author}{Stachniss, C.}, \bibinfo{year}{2021}.
\newblock \bibinfo{title}{{Towards 3D LiDAR-based semantic scene understanding
  of 3D point cloud sequences: The SemanticKITTI Dataset}}.
\newblock \bibinfo{journal}{The International Journal on Robotics Research}
%Type = Article
\bibitem[{Bernotas et~al.(2019)Bernotas, Scorza, Hansen, Hales, Halliday,
  Smith, Smith and McCormick}]{10.1093/gigascience/giz056}
\bibinfo{author}{Bernotas, G.}, \bibinfo{author}{Scorza, L.C.T.},
  \bibinfo{author}{Hansen, M.F.}, \bibinfo{author}{Hales, I.J.},
  \bibinfo{author}{Halliday, K.J.}, \bibinfo{author}{Smith, L.N.},
  \bibinfo{author}{Smith, M.L.}, \bibinfo{author}{McCormick, A.J.},
  \bibinfo{year}{2019}.
\newblock \bibinfo{title}{A photometric stereo-based 3d imaging system using
  computer vision and deep learning for tracking plant growth}.
\newblock \bibinfo{journal}{GigaScience}
%Type = Article
\bibitem[{Bienert et~al.(2021)Bienert, Georgi, Kunz, von Oheimb and
  Maas}]{10.1093/aob/mcab087}
\bibinfo{author}{Bienert, A.}, \bibinfo{author}{Georgi, L.},
  \bibinfo{author}{Kunz, M.}, \bibinfo{author}{von Oheimb, G.},
  \bibinfo{author}{Maas, H.G.}, \bibinfo{year}{2021}.
\newblock \bibinfo{title}{Automatic extraction and measurement of individual
  trees from mobile laser scanning point clouds of forests}.
\newblock \bibinfo{journal}{Annals of Botany}
%Type = Article
\bibitem[{Brede et~al.(2022)Brede, Terryn, Barbier, Bartholomeus, Bartolo,
  Calders, Derroire, {Krishna Moorthy}, Lau, Levick, Raumonen, Verbeeck, Wang,
  Whiteside, {van der Zee} and Herold}]{BREDE2022113180}
\bibinfo{author}{Brede, B.}, \bibinfo{author}{Terryn, L.},
  \bibinfo{author}{Barbier, N.}, \bibinfo{author}{Bartholomeus, H.M.},
  \bibinfo{author}{Bartolo, R.}, \bibinfo{author}{Calders, K.},
  \bibinfo{author}{Derroire, G.}, \bibinfo{author}{{Krishna Moorthy}, S.M.},
  \bibinfo{author}{Lau, A.}, \bibinfo{author}{Levick, S.R.},
  \bibinfo{author}{Raumonen, P.}, \bibinfo{author}{Verbeeck, H.},
  \bibinfo{author}{Wang, D.}, \bibinfo{author}{Whiteside, T.},
  \bibinfo{author}{{van der Zee}, J.}, \bibinfo{author}{Herold, M.},
  \bibinfo{year}{2022}.
\newblock \bibinfo{title}{Non-destructive estimation of individual tree
  biomass: Allometric models, terrestrial and uav laser scanning}.
\newblock \bibinfo{journal}{Remote Sensing of Environment}
%Type = Article
\bibitem[{Van~den Broeck et~al.(2023)Van~den Broeck, Terryn, Cherlet, Cooper
  and Calders}]{isprs-archives-XLVIII-1-W2-2023-765-2023}
\bibinfo{author}{Van~den Broeck, W.A.J.}, \bibinfo{author}{Terryn, L.},
  \bibinfo{author}{Cherlet, W.}, \bibinfo{author}{Cooper, Z.T.},
  \bibinfo{author}{Calders, K.}, \bibinfo{year}{2023}.
\newblock \bibinfo{title}{Three-dimensional deep learning for leaf-wood
  segmentation of tropical tree point clouds}.
\newblock \bibinfo{journal}{The International Archives of the Photogrammetry,
  Remote Sensing and Spatial Information Sciences}
%Type = Article
\bibitem[{Bryson et~al.(2023)Bryson, Wang and Allworth}]{rs15092380}
\bibinfo{author}{Bryson, M.}, \bibinfo{author}{Wang, F.},
  \bibinfo{author}{Allworth, J.}, \bibinfo{year}{2023}.
\newblock \bibinfo{title}{Using synthetic tree data in deep learning-based tree
  segmentation using lidar point clouds}.
\newblock \bibinfo{journal}{Remote Sensing}
%Type = Inproceedings
\bibitem[{\c{C}i\c{c}ek et~al.(2016)\c{C}i\c{c}ek, Abdulkadir, Lienkamp, Brox
  and Ronneberger}]{10.1007/978-3-319-46723-8_49}
\bibinfo{author}{\c{C}i\c{c}ek, O.}, \bibinfo{author}{Abdulkadir, A.},
  \bibinfo{author}{Lienkamp, S.S.}, \bibinfo{author}{Brox, T.},
  \bibinfo{author}{Ronneberger, O.}, \bibinfo{year}{2016}.
\newblock \bibinfo{title}{3d u-net: Learning dense volumetric segmentation from
  sparse annotation}, in: \bibinfo{booktitle}{Medical Image Computing and
  Computer-Assisted Intervention – MICCAI 2016: 19th International
  Conference, Athens, Greece, October 17-21, 2016, Proceedings, Part II},
  \bibinfo{publisher}{Springer-Verlag}, \bibinfo{address}{Berlin, Heidelberg}.
%Type = Inproceedings
\bibitem[{Chaudhury et~al.(2020)Chaudhury, Boudon and Godin}]{Lpy-Arabidopsis}
\bibinfo{author}{Chaudhury, A.}, \bibinfo{author}{Boudon, F.},
  \bibinfo{author}{Godin, C.}, \bibinfo{year}{2020}.
\newblock \bibinfo{title}{3d plant phenotyping: All you need is labelled point
  cloud data}, in: \bibinfo{editor}{Bartoli, A.}, \bibinfo{editor}{Fusiello,
  A.} (Eds.), \bibinfo{booktitle}{Computer Vision -- ECCV 2020 Workshops},
  \bibinfo{publisher}{Springer International Publishing},
  \bibinfo{address}{Cham}.
%Type = Inproceedings
\bibitem[{Chen et~al.(2021)Chen, Fang, Zhang, Liu and
  Wang}]{Chen_HAIS_2021_ICCV}
\bibinfo{author}{Chen, S.}, \bibinfo{author}{Fang, J.}, \bibinfo{author}{Zhang,
  Q.}, \bibinfo{author}{Liu, W.}, \bibinfo{author}{Wang, X.},
  \bibinfo{year}{2021}.
\newblock \bibinfo{title}{Hierarchical aggregation for 3d instance
  segmentation},
%Type = Inproceedings
\bibitem[{Chen et~al.(2015)Chen, Kundu, Zhu, Berneshawi, Ma, Fidler and
  Urtasun}]{NIPS2015_6da37dd3}
\bibinfo{author}{Chen, X.}, \bibinfo{author}{Kundu, K.}, \bibinfo{author}{Zhu,
  Y.}, \bibinfo{author}{Berneshawi, A.G.}, \bibinfo{author}{Ma, H.},
  \bibinfo{author}{Fidler, S.}, \bibinfo{author}{Urtasun, R.},
  \bibinfo{year}{2015}.
\newblock \bibinfo{title}{3d object proposals for accurate object class
  detection}, in: \bibinfo{editor}{Cortes, C.}, \bibinfo{editor}{Lawrence, N.},
  \bibinfo{editor}{Lee, D.}, \bibinfo{editor}{Sugiyama, M.},
  \bibinfo{editor}{Garnett, R.} (Eds.), \bibinfo{booktitle}{Advances in Neural
  Information Processing Systems},
%Type = Inproceedings
\bibitem[{Cheng et~al.(2022)Cheng, Misra, Schwing, Kirillov and
  Girdhar}]{cheng2021mask2former}
\bibinfo{author}{Cheng, B.}, \bibinfo{author}{Misra, I.},
  \bibinfo{author}{Schwing, A.G.}, \bibinfo{author}{Kirillov, A.},
  \bibinfo{author}{Girdhar, R.}, \bibinfo{year}{2022}.
\newblock
%Type = Inproceedings
\bibitem[{Choy et~al.(2019)Choy, Gwak and Savarese}]{choy20194d}
\bibinfo{author}{Choy, C.}, \bibinfo{author}{Gwak, J.},
  \bibinfo{author}{Savarese, S.}, \bibinfo{year}{2019}.
\newblock \bibinfo{title}{4d spatio-temporal convnets: Minkowski convolutional
  neural networks}, in: \bibinfo{booktitle}{Proceedings of the IEEE Conference
  on Computer Vision and Pattern Recognition},
%Type = Article
\bibitem[{Chéné et~al.(2012)Chéné, Rousseau, Lucidarme, Bertheloot,
  Caffier, Morel, Étienne Belin and Chapeau-Blondeau}]{CHENE2012122}
\bibinfo{author}{Chéné, Y.}, \bibinfo{author}{Rousseau, D.},
  \bibinfo{author}{Lucidarme, P.}, \bibinfo{author}{Bertheloot, J.},
  \bibinfo{author}{Caffier, V.}, \bibinfo{author}{Morel, P.},
  \bibinfo{author}{Étienne Belin}, \bibinfo{author}{Chapeau-Blondeau, F.},
  \bibinfo{year}{2012}.
\newblock \bibinfo{title}{On the use of depth camera for 3d phenotyping of
  entire plants}.
\newblock \bibinfo{journal}{Computers and Electronics in Agriculture}
%Type = Article
\bibitem[{Conn et~al.(2017a)Conn, Pedmale, Chory and Navlakha}]{CONNa}
\bibinfo{author}{Conn, A.}, \bibinfo{author}{Pedmale, U.V.},
  \bibinfo{author}{Chory, J.}, \bibinfo{author}{Navlakha, S.},
  \bibinfo{year}{2017}a.
\newblock \bibinfo{title}{High-resolution laser scanning reveals plant
  architectures that reflect universal network design principles}.
\newblock \bibinfo{journal}{Cell Systems}
%Type = Article
\bibitem[{Conn et~al.(2017b)Conn, Pedmale, Chory, Stevens and Navlakha}]{CONNb}
\bibinfo{author}{Conn, A.}, \bibinfo{author}{Pedmale, U.V.},
  \bibinfo{author}{Chory, J.}, \bibinfo{author}{Stevens, C.F.},
  \bibinfo{author}{Navlakha, S.}, \bibinfo{year}{2017}b.
\newblock \bibinfo{title}{A statistical description of plant shoot
  architecture}.
\newblock \bibinfo{journal}{Current Biology}
%Type = Misc
\bibitem[{Contributors(2020)}]{mmdet3d2020}
\bibinfo{author}{Contributors, M.}, \bibinfo{year}{2020}.
\newblock \bibinfo{title}{{MMDetection3D: OpenMMLab} next-generation platform
  for general {3D} object detection}.
\newblock
%Type = Misc
\bibitem[{Contributors(2022)}]{Spconv2022}
\bibinfo{author}{Contributors, S.}, \bibinfo{year}{2022}.
\newblock \bibinfo{title}{Spconv: Spatially sparse convolution library}.
\newblock
%Type = Article
\bibitem[{{Cournède, P.-H.} et~al.(2011){Cournède, P.-H.}, {Letort, V.},
  {Mathieu, A.}, {Kang, M. Z.}, {Lemaire, S.}, {Trevezas, S.}, {Houllier, F.}
  and {de Reffye, P.}}]{refId0}
\bibinfo{author}{{Cournède, P.-H.}}, \bibinfo{author}{{Letort, V.}},
  \bibinfo{author}{{Mathieu, A.}}, \bibinfo{author}{{Kang, M. Z.}},
  \bibinfo{author}{{Lemaire, S.}}, \bibinfo{author}{{Trevezas, S.}},
  \bibinfo{author}{{Houllier, F.}}, \bibinfo{author}{{de Reffye, P.}},
  \bibinfo{year}{2011}.
\newblock \bibinfo{title}{Some parameter estimation issues in
  functional-structural plant modelling}.
\newblock \bibinfo{journal}{Math. Model. Nat. Phenom.}
%Type = Article
\bibitem[{Cui et~al.(2025)Cui, Liu, Liu, Zhao and Feng}]{agriculture15020175}
\bibinfo{author}{Cui, D.}, \bibinfo{author}{Liu, P.}, \bibinfo{author}{Liu,
  Y.}, \bibinfo{author}{Zhao, Z.}, \bibinfo{author}{Feng, J.},
  \bibinfo{year}{2025}.
\newblock \bibinfo{title}{Automated phenotypic analysis of mature soybean using
  multi-view stereo 3d reconstruction and point cloud segmentation}.
\newblock \bibinfo{journal}{Agriculture}
%Type = Inproceedings
\bibitem[{Dai et~al.(2017)Dai, Chang, Savva, Halber, Funkhouser and
  Nie{\ss}ner}]{dai2017scannet}
\bibinfo{author}{Dai, A.}, \bibinfo{author}{Chang, A.X.},
  \bibinfo{author}{Savva, M.}, \bibinfo{author}{Halber, M.},
  \bibinfo{author}{Funkhouser, T.}, \bibinfo{author}{Nie{\ss}ner, M.},
  \bibinfo{year}{2017}.
\newblock \bibinfo{title}{Scannet: Richly-annotated 3d reconstructions of
  indoor scenes},
%Type = Article
\bibitem[{Demol et~al.(2022)Demol, Verbeeck, Gielen, Armston, Burt, Disney,
  Duncanson, Hackenberg, Kükenbrink, Lau, Ploton, Sewdien, Stovall, Takoudjou,
  Volkova, Weston, Wortel and
  Calders}]{https://doi.org/10.1111/2041-210X.13906}
\bibinfo{author}{Demol, M.}, \bibinfo{author}{Verbeeck, H.},
  \bibinfo{author}{Gielen, B.}, \bibinfo{author}{Armston, J.},
  \bibinfo{author}{Burt, A.}, \bibinfo{author}{Disney, M.},
  \bibinfo{author}{Duncanson, L.}, \bibinfo{author}{Hackenberg, J.},
  \bibinfo{author}{Kükenbrink, D.}, \bibinfo{author}{Lau, A.},
  \bibinfo{author}{Ploton, P.}, \bibinfo{author}{Sewdien, A.},
  \bibinfo{author}{Stovall, A.}, \bibinfo{author}{Takoudjou, S.M.},
  \bibinfo{author}{Volkova, L.}, \bibinfo{author}{Weston, C.},
  \bibinfo{author}{Wortel, V.}, \bibinfo{author}{Calders, K.},
  \bibinfo{year}{2022}.
\newblock \bibinfo{title}{Estimating forest above-ground biomass with
  terrestrial laser scanning: Current status and future directions}.
\newblock \bibinfo{journal}{Methods in Ecology and Evolution}
%Type = Article
\bibitem[{Du et~al.(2016)Du, Zhang, Guo, Ma, Shao, Pan and Zhao}]{du2016micron}
\bibinfo{author}{Du, J.}, \bibinfo{author}{Zhang, Y.}, \bibinfo{author}{Guo,
  X.}, \bibinfo{author}{Ma, L.}, \bibinfo{author}{Shao, M.},
  \bibinfo{author}{Pan, X.}, \bibinfo{author}{Zhao, C.}, \bibinfo{year}{2016}.
\newblock \bibinfo{title}{Micron-scale phenotyping quantification and
  three-dimensional microstructure reconstruction of vascular bundles within
  maize stalks based on micro-ct scanning}.
\newblock \bibinfo{journal}{Functional Plant Biology}
%Type = Article
\bibitem[{Du et~al.(2023)Du, Ma, Xie, He and Cen}]{DU2023380}
\bibinfo{author}{Du, R.}, \bibinfo{author}{Ma, Z.}, \bibinfo{author}{Xie, P.},
  \bibinfo{author}{He, Y.}, \bibinfo{author}{Cen, H.}, \bibinfo{year}{2023}.
\newblock \bibinfo{title}{Pst: Plant segmentation transformer for 3d point
  clouds of rapeseed plants at the podding stage}.
\newblock \bibinfo{journal}{ISPRS Journal of Photogrammetry and Remote Sensing}
%Type = Inproceedings
\bibitem[{Du et~al.(2024)Du, Qiu, Xu and Jiang}]{DUASABEinproceedings}
\bibinfo{author}{Du, R.}, \bibinfo{author}{Qiu, T.}, \bibinfo{author}{Xu, K.},
  \bibinfo{author}{Jiang, Y.}, \bibinfo{year}{2024}.
\newblock \bibinfo{title}{Simulated data enhances three-dimensional
  segmentation-based characterization of real apple trees},
%Type = Article
\bibitem[{Dutağacı et~al.(2020)Dutağacı, Rasti, Galopin and
  Rousseau}]{RoseX}
\bibinfo{author}{Dutağacı, H.}, \bibinfo{author}{Rasti, P.},
  \bibinfo{author}{Galopin, G.}, \bibinfo{author}{Rousseau, D.},
  \bibinfo{year}{2020}.
\newblock \bibinfo{title}{Rose-x: an annotated data set for evaluation of 3d
  plant organ segmentation methods}.
\newblock \bibinfo{journal}{Plant Methods}
%Type = Article
\bibitem[{Easlon and Bloom(2014)}]{https://doi.org/10.3732/apps.1400033}
\bibinfo{author}{Easlon, H.M.}, \bibinfo{author}{Bloom, A.J.},
  \bibinfo{year}{2014}.
\newblock \bibinfo{title}{Easy leaf area: Automated digital image analysis for
  rapid and accurate measurement of leaf area}.
\newblock \bibinfo{journal}{Applications in Plant Sciences}
%Type = Inproceedings
\bibitem[{Engelmann et~al.(2020)Engelmann, Bokeloh, Fathi, Leibe and
  Nie{\ss}ner}]{Engelmann20CVPR}
\bibinfo{author}{Engelmann, F.}, \bibinfo{author}{Bokeloh, M.},
  \bibinfo{author}{Fathi, A.}, \bibinfo{author}{Leibe, B.},
  \bibinfo{author}{Nie{\ss}ner, M.}, \bibinfo{year}{2020}.
\newblock \bibinfo{title}{{3D-MPA: Multi Proposal Aggregation for 3D Semantic
  Instance Segmentation}},
%Type = Inproceedings
\bibitem[{Engelmann et~al.(2018)Engelmann, Kontogianni, Schult and
  Leibe}]{10.1007/978-3-030-11015-4_29}
\bibinfo{author}{Engelmann, F.}, \bibinfo{author}{Kontogianni, T.},
  \bibinfo{author}{Schult, J.}, \bibinfo{author}{Leibe, B.},
  \bibinfo{year}{2018}.
\newblock \bibinfo{title}{Know what your neighbors do: 3d semantic segmentation
  of point clouds}, in: \bibinfo{booktitle}{Computer Vision – ECCV 2018
  Workshops: Munich, Germany, September 8-14, 2018, Proceedings, Part III},
  \bibinfo{publisher}{Springer-Verlag}, \bibinfo{address}{Berlin, Heidelberg}.
%Type = Article
\bibitem[{Esser et~al.(2023)Esser, Klingbeil, Zabawa and Kuhlmann}]{rs15041117}
\bibinfo{author}{Esser, F.}, \bibinfo{author}{Klingbeil, L.},
  \bibinfo{author}{Zabawa, L.}, \bibinfo{author}{Kuhlmann, H.},
  \bibinfo{year}{2023}.
\newblock \bibinfo{title}{Quality analysis of a high-precision kinematic laser
  scanning system for the use of spatio-temporal plant and organ-level
  phenotyping in the field}.
\newblock \bibinfo{journal}{Remote Sensing}
%Type = Article
\bibitem[{Fang et~al.(2019)Fang, Baret, Plummer and
  Schaepman-Strub}]{https://doi.org/10.1029/2018RG000608}
\bibinfo{author}{Fang, H.}, \bibinfo{author}{Baret, F.},
  \bibinfo{author}{Plummer, S.}, \bibinfo{author}{Schaepman-Strub, G.},
  \bibinfo{year}{2019}.
\newblock \bibinfo{title}{An overview of global leaf area index (lai): Methods,
  products, validation, and applications}.
\newblock \bibinfo{journal}{Reviews of Geophysics}
%Type = Article
\bibitem[{Freschet et~al.(2021)Freschet, Pag{\`e}s, Iversen, Comas, Rewald,
  Roumet, Klime{\v{s}}ov{\'a}, Zadworny, Poorter, Postma
  et~al.}]{freschet2021starting}
\bibinfo{author}{Freschet, G.T.}, \bibinfo{author}{Pag{\`e}s, L.},
  \bibinfo{author}{Iversen, C.M.}, \bibinfo{author}{Comas, L.H.},
  \bibinfo{author}{Rewald, B.}, \bibinfo{author}{Roumet, C.},
  \bibinfo{author}{Klime{\v{s}}ov{\'a}, J.}, \bibinfo{author}{Zadworny, M.},
  \bibinfo{author}{Poorter, H.}, \bibinfo{author}{Postma, J.A.}, et~al.,
  \bibinfo{year}{2021}.
\newblock \bibinfo{title}{A starting guide to root ecology: strengthening
  ecological concepts and standardising root classification, sampling,
  processing and trait measurements}.
\newblock \bibinfo{journal}{New Phytologist}
%Type = Inproceedings
\bibitem[{Gaillard et~al.(2020)Gaillard, Miao, Schnable and
  Benes}]{10.1007/978-3-030-65414-6_21}
\bibinfo{author}{Gaillard, M.}, \bibinfo{author}{Miao, C.},
  \bibinfo{author}{Schnable, J.}, \bibinfo{author}{Benes, B.},
  \bibinfo{year}{2020}.
\newblock \bibinfo{title}{Sorghum segmentation by skeleton extraction}, in:
  \bibinfo{editor}{Bartoli, A.}, \bibinfo{editor}{Fusiello, A.} (Eds.),
  \bibinfo{booktitle}{Computer Vision -- ECCV 2020 Workshops},
  \bibinfo{publisher}{Springer International Publishing},
  \bibinfo{address}{Cham}.
%Type = Article
\bibitem[{Gao et~al.(2024)Gao, Liu, Chen, Liu, Yan and Zhang}]{rs16061079}
\bibinfo{author}{Gao, L.}, \bibinfo{author}{Liu, Y.}, \bibinfo{author}{Chen,
  X.}, \bibinfo{author}{Liu, Y.}, \bibinfo{author}{Yan, S.},
  \bibinfo{author}{Zhang, M.}, \bibinfo{year}{2024}.
\newblock \bibinfo{title}{Cus3d: A new comprehensive urban-scale
  semantic-segmentation 3d benchmark dataset}.
\newblock \bibinfo{journal}{Remote Sensing}
%Type = Article
\bibitem[{Gao et~al.(2021)Gao, Zhu, Paul, Sandhu, Doku, Sun, Pan, Staswick,
  Walia and Yu}]{rs13112113}
\bibinfo{author}{Gao, T.}, \bibinfo{author}{Zhu, F.}, \bibinfo{author}{Paul,
  P.}, \bibinfo{author}{Sandhu, J.}, \bibinfo{author}{Doku, H.A.},
  \bibinfo{author}{Sun, J.}, \bibinfo{author}{Pan, Y.},
  \bibinfo{author}{Staswick, P.}, \bibinfo{author}{Walia, H.},
  \bibinfo{author}{Yu, H.}, \bibinfo{year}{2021}.
\newblock \bibinfo{title}{Novel 3d imaging systems for high-throughput
  phenotyping of plants}.
\newblock \bibinfo{journal}{Remote Sensing}
%Type = Article
\bibitem[{Ge et~al.(2025)Ge, Wu, Wen, Shen, Xiao, Lu, Liu, Zhang and
  Guo}]{GE202573}
\bibinfo{author}{Ge, X.}, \bibinfo{author}{Wu, S.}, \bibinfo{author}{Wen, W.},
  \bibinfo{author}{Shen, F.}, \bibinfo{author}{Xiao, P.}, \bibinfo{author}{Lu,
  X.}, \bibinfo{author}{Liu, H.}, \bibinfo{author}{Zhang, M.},
  \bibinfo{author}{Guo, X.}, \bibinfo{year}{2025}.
\newblock \bibinfo{title}{Lettucep3d: A tool for analysing 3d phenotypes of
  individual lettuce plants}.
\newblock \bibinfo{journal}{Biosystems Engineering}
%Type = Article
\bibitem[{Gené-Mola et~al.(2021)Gené-Mola, Sanz-Cortiella, Rosell-Polo,
  Escolà and Gregorio}]{GENEMOLA2021107629}
\bibinfo{author}{Gené-Mola, J.}, \bibinfo{author}{Sanz-Cortiella, R.},
  \bibinfo{author}{Rosell-Polo, J.R.}, \bibinfo{author}{Escolà, A.},
  \bibinfo{author}{Gregorio, E.}, \bibinfo{year}{2021}.
\newblock \bibinfo{title}{Pfuji-size dataset: A collection of images and
  photogrammetry-derived 3d point clouds with ground truth annotations for fuji
  apple detection and size estimation in field conditions}.
\newblock \bibinfo{journal}{Data in Brief}
%Type = Article
\bibitem[{Gong et~al.(2021)Gong, Du, Zhu, Lin, Lou, Yuan, Huang and
  Liu}]{Panicle3D}
\bibinfo{author}{Gong, L.}, \bibinfo{author}{Du, X.}, \bibinfo{author}{Zhu,
  K.}, \bibinfo{author}{Lin, K.}, \bibinfo{author}{Lou, Q.},
  \bibinfo{author}{Yuan, Z.}, \bibinfo{author}{Huang, G.},
  \bibinfo{author}{Liu, C.}, \bibinfo{year}{2021}.
\newblock \bibinfo{title}{Panicle-3d: Efficient phenotyping tool for precise
  semantic segmentation of rice panicle point cloud}.
\newblock \bibinfo{journal}{Plant Phenomics}
%Type = Article
\bibitem[{Graham(2015)}]{graham2015sparse}
\bibinfo{author}{Graham, B.}, \bibinfo{year}{2015}.
\newblock \bibinfo{title}{Sparse 3d convolutional neural networks}.
\newblock
%Type = Article
\bibitem[{Graham et~al.(2018)Graham, Engelcke and van~der
  Maaten}]{3DSemanticSegmentationWithSubmanifoldSparseConvNet}
\bibinfo{author}{Graham, B.}, \bibinfo{author}{Engelcke, M.},
  \bibinfo{author}{van~der Maaten, L.}, \bibinfo{year}{2018}.
\newblock \bibinfo{title}{3d semantic segmentation with submanifold sparse
  convolutional networks}.
\newblock
%Type = Article
\bibitem[{Guo et~al.(2021)Guo, Su, Hu, Guan, Jin, Zhang, Zhao, Xu, Wei, Kelly
  and Coops}]{9309060}
\bibinfo{author}{Guo, Q.}, \bibinfo{author}{Su, Y.}, \bibinfo{author}{Hu, T.},
  \bibinfo{author}{Guan, H.}, \bibinfo{author}{Jin, S.},
  \bibinfo{author}{Zhang, J.}, \bibinfo{author}{Zhao, X.}, \bibinfo{author}{Xu,
  K.}, \bibinfo{author}{Wei, D.}, \bibinfo{author}{Kelly, M.},
  \bibinfo{author}{Coops, N.C.}, \bibinfo{year}{2021}.
\newblock \bibinfo{title}{Lidar boosts 3d ecological observations and
  modelings: A review and perspective}.
\newblock \bibinfo{journal}{IEEE Geoscience and Remote Sensing Magazine}
%Type = Article
\bibitem[{Guo et~al.(2020)Guo, Wang, Hu, Liu, Liu and
  Bennamoun}]{ASurveyarticle}
\bibinfo{author}{Guo, Y.}, \bibinfo{author}{Wang, H.}, \bibinfo{author}{Hu,
  Q.}, \bibinfo{author}{Liu, H.}, \bibinfo{author}{Liu, L.},
  \bibinfo{author}{Bennamoun, M.}, \bibinfo{year}{2020}.
\newblock \bibinfo{title}{Deep learning for 3d point clouds: A survey}.
\newblock \bibinfo{journal}{IEEE Transactions on Pattern Analysis and Machine
  Intelligence}
%Type = Article
\bibitem[{Hackel et~al.(2017)Hackel, Savinov, Ladicky, Wegner, Schindler and
  Pollefeys}]{isprs-annals-IV-1-W1-91-2017}
\bibinfo{author}{Hackel, T.}, \bibinfo{author}{Savinov, N.},
  \bibinfo{author}{Ladicky, L.}, \bibinfo{author}{Wegner, J.D.},
  \bibinfo{author}{Schindler, K.}, \bibinfo{author}{Pollefeys, M.},
  \bibinfo{year}{2017}.
\newblock \bibinfo{title}{Semantic3d.net: A new large-scale point cloud
  classification benchmark}.
\newblock \bibinfo{journal}{ISPRS Annals of the Photogrammetry, Remote Sensing
  and Spatial Information Sciences}
%Type = Misc
\bibitem[{Hahner et~al.(2022)Hahner, Dai, Liniger and
  Gool}]{hahner2022quantifyingdataaugmentationlidar}
\bibinfo{author}{Hahner, M.}, \bibinfo{author}{Dai, D.},
  \bibinfo{author}{Liniger, A.}, \bibinfo{author}{Gool, L.V.},
  \bibinfo{year}{2022}.
\newblock
%Type = Inproceedings
\bibitem[{Hamamoto et~al.(2020)Hamamoto, Uchiyama, Shimada and
  Taniguchi}]{visapp20}
\bibinfo{author}{Hamamoto, T.}, \bibinfo{author}{Uchiyama, H.},
  \bibinfo{author}{Shimada, A.}, \bibinfo{author}{Taniguchi, R.},
  \bibinfo{year}{2020}.
\newblock \bibinfo{title}{3d plant growth prediction via image-to-image
  translation}, in: \bibinfo{booktitle}{Proceedings of the 15th International
  Joint Conference on Computer Vision, Imaging and Computer Graphics Theory and
  Applications (VISIGRAPP 2020) - Volume 5: VISAPP},
  \bibinfo{organization}{INSTICC}. \bibinfo{publisher}{SciTePress}.
%Type = Inproceedings
\bibitem[{Han et~al.(2020)Han, Zheng, Xu and Fang}]{9157103}
\bibinfo{author}{Han, L.}, \bibinfo{author}{Zheng, T.}, \bibinfo{author}{Xu,
  L.}, \bibinfo{author}{Fang, L.}, \bibinfo{year}{2020}.
\newblock \bibinfo{title}{Occuseg: Occupancy-aware 3d instance segmentation},
  in: \bibinfo{booktitle}{2020 IEEE/CVF Conference on Computer Vision and
  Pattern Recognition (CVPR)},
%Type = Article
\bibitem[{Han et~al.(2021)Han, Dong and Yang}]{HAN2021199}
\bibinfo{author}{Han, X.}, \bibinfo{author}{Dong, Z.}, \bibinfo{author}{Yang,
  B.}, \bibinfo{year}{2021}.
\newblock \bibinfo{title}{A point-based deep learning network for semantic
  segmentation of mls point clouds}.
\newblock \bibinfo{journal}{ISPRS Journal of Photogrammetry and Remote Sensing}
%Type = Inproceedings
\bibitem[{He et~al.(2017)He, Gkioxari, Doll{\'a}r and Girshick}]{he2017mask}
\bibinfo{author}{He, K.}, \bibinfo{author}{Gkioxari, G.},
  \bibinfo{author}{Doll{\'a}r, P.}, \bibinfo{author}{Girshick, R.},
  \bibinfo{year}{2017}.
\newblock \bibinfo{title}{Mask r-cnn}, in: \bibinfo{booktitle}{Proceedings of
  the IEEE international conference on computer vision},
%Type = Inproceedings
\bibitem[{He et~al.(2020)He, Liu, Shen, Wang and
  Sun}]{10.1007/978-3-030-58577-8_16}
\bibinfo{author}{He, T.}, \bibinfo{author}{Liu, Y.}, \bibinfo{author}{Shen,
  C.}, \bibinfo{author}{Wang, X.}, \bibinfo{author}{Sun, C.},
  \bibinfo{year}{2020}.
\newblock \bibinfo{title}{Instance-aware embedding for point cloud instance
  segmentation}, in: \bibinfo{booktitle}{Computer Vision – ECCV 2020: 16th
  European Conference, Glasgow, UK, August 23–28, 2020, Proceedings, Part
  XXX}, \bibinfo{publisher}{Springer-Verlag}, \bibinfo{address}{Berlin,
  Heidelberg}.
%Type = Inproceedings
\bibitem[{He et~al.(2021)He, Shen and van~den Hengel}]{9577908}
\bibinfo{author}{He, T.}, \bibinfo{author}{Shen, C.}, \bibinfo{author}{van~den
  Hengel, A.}, \bibinfo{year}{2021}.
\newblock \bibinfo{title}{{ DyCo3D: Robust Instance Segmentation of 3D Point
  Clouds through Dynamic Convolution }}, in: \bibinfo{booktitle}{2021 IEEE/CVF
  Conference on Computer Vision and Pattern Recognition (CVPR)},
  \bibinfo{publisher}{IEEE Computer Society}, \bibinfo{address}{Los Alamitos,
  CA, USA}.
%Type = Article
\bibitem[{Henrich et~al.(2024)Henrich, {van Delden}, Seidel, Kneib and
  Ecker}]{HENRICH2024102888}
\bibinfo{author}{Henrich, J.}, \bibinfo{author}{{van Delden}, J.},
  \bibinfo{author}{Seidel, D.}, \bibinfo{author}{Kneib, T.},
  \bibinfo{author}{Ecker, A.S.}, \bibinfo{year}{2024}.
\newblock \bibinfo{title}{Treelearn: A deep learning method for segmenting
  individual trees from ground-based lidar forest point clouds}.
\newblock \bibinfo{journal}{Ecological Informatics}
%Type = Article
\bibitem[{Hornik(1991)}]{HORNIK1991251}
\bibinfo{author}{Hornik, K.}, \bibinfo{year}{1991}.
\newblock \bibinfo{title}{Approximation capabilities of multilayer feedforward
  networks}.
\newblock \bibinfo{journal}{Neural Networks}
%Type = Inproceedings
\bibitem[{Hu et~al.(2021)Hu, Yang, Khalid, Xiao, Trigoni and
  Markham}]{Hu_2021_CVPR}
\bibinfo{author}{Hu, Q.}, \bibinfo{author}{Yang, B.}, \bibinfo{author}{Khalid,
  S.}, \bibinfo{author}{Xiao, W.}, \bibinfo{author}{Trigoni, N.},
  \bibinfo{author}{Markham, A.}, \bibinfo{year}{2021}.
\newblock \bibinfo{title}{Towards semantic segmentation of urban-scale 3d point
  clouds: A dataset, benchmarks and challenges}, in:
  \bibinfo{booktitle}{Proceedings of the IEEE/CVF Conference on Computer Vision
  and Pattern Recognition (CVPR)},
%Type = Article
\bibitem[{Hu et~al.(2020)Hu, Yang, Xie, Rosa, Guo, Wang, Trigoni and
  Markham}]{hu2019randla}
\bibinfo{author}{Hu, Q.}, \bibinfo{author}{Yang, B.}, \bibinfo{author}{Xie,
  L.}, \bibinfo{author}{Rosa, S.}, \bibinfo{author}{Guo, Y.},
  \bibinfo{author}{Wang, Z.}, \bibinfo{author}{Trigoni, N.},
  \bibinfo{author}{Markham, A.}, \bibinfo{year}{2020}.
\newblock \bibinfo{title}{Randla-net: Efficient semantic segmentation of
  large-scale point clouds}.
\newblock
%Type = Inproceedings
\bibitem[{Huang and You(2016)}]{Dense3DCNNinproceedings}
\bibinfo{author}{Huang, J.}, \bibinfo{author}{You, S.}, \bibinfo{year}{2016}.
\newblock
%Type = Article
\bibitem[{Hui et~al.(2021)Hui, Jin, Xia, Wang, {Yevenyo Ziggah} and
  Cheng}]{HUI2021219}
\bibinfo{author}{Hui, Z.}, \bibinfo{author}{Jin, S.}, \bibinfo{author}{Xia,
  Y.}, \bibinfo{author}{Wang, L.}, \bibinfo{author}{{Yevenyo Ziggah}, Y.},
  \bibinfo{author}{Cheng, P.}, \bibinfo{year}{2021}.
\newblock \bibinfo{title}{Wood and leaf separation from terrestrial lidar point
  clouds based on mode points evolution}.
\newblock \bibinfo{journal}{ISPRS Journal of Photogrammetry and Remote Sensing}
%Type = Article
\bibitem[{Itakura and Hosoi(2018)}]{s18103576}
\bibinfo{author}{Itakura, K.}, \bibinfo{author}{Hosoi, F.},
  \bibinfo{year}{2018}.
\newblock \bibinfo{title}{Automatic leaf segmentation for estimating leaf area
  and leaf inclination angle in 3d plant images}.
\newblock \bibinfo{journal}{Sensors}
%Type = Article
\bibitem[{James et~al.(2018)James, Wohlhart, Kalakrishnan, Kalashnikov, Irpan,
  Ibarz, Levine, Hadsell and Bousmalis}]{James2018SimToRealVS}
\bibinfo{author}{James, S.}, \bibinfo{author}{Wohlhart, P.},
  \bibinfo{author}{Kalakrishnan, M.}, \bibinfo{author}{Kalashnikov, D.},
  \bibinfo{author}{Irpan, A.}, \bibinfo{author}{Ibarz, J.},
  \bibinfo{author}{Levine, S.}, \bibinfo{author}{Hadsell, R.},
  \bibinfo{author}{Bousmalis, K.}, \bibinfo{year}{2018}.
\newblock \bibinfo{title}{Sim-to-real via sim-to-sim: Data-efficient robotic
  grasping via randomized-to-canonical adaptation networks}.
\newblock
%Type = Article
\bibitem[{Jiang et~al.(2020)Jiang, Zhao, Shi, Liu, Fu and
  Jia}]{jiang2020pointgroup}
\bibinfo{author}{Jiang, L.}, \bibinfo{author}{Zhao, H.}, \bibinfo{author}{Shi,
  S.}, \bibinfo{author}{Liu, S.}, \bibinfo{author}{Fu, C.W.},
  \bibinfo{author}{Jia, J.}, \bibinfo{year}{2020}.
\newblock \bibinfo{title}{Pointgroup: Dual-set point grouping for 3d instance
  segmentation}.
\newblock
%Type = Misc
\bibitem[{Jiang et~al.(2018)Jiang, Wu, Zhao, Zhao and
  Lu}]{jiang2018pointsiftsiftlikenetworkmodule}
\bibinfo{author}{Jiang, M.}, \bibinfo{author}{Wu, Y.}, \bibinfo{author}{Zhao,
  T.}, \bibinfo{author}{Zhao, Z.}, \bibinfo{author}{Lu, C.},
  \bibinfo{year}{2018}.
\newblock
%Type = Article
\bibitem[{Jiang and Li(2020)}]{doi:10.34133/2020/4152816}
\bibinfo{author}{Jiang, Y.}, \bibinfo{author}{Li, C.}, \bibinfo{year}{2020}.
\newblock \bibinfo{title}{Convolutional neural networks for image-based
  high-throughput plant phenotyping: A review}.
\newblock \bibinfo{journal}{Plant Phenomics}
%Type = Article
\bibitem[{Jin et~al.(2020)Jin, Su, Gao, Wu, Ma, Xu, Ma, Hu, Liu, Pang, Guan,
  Zhang and Guo}]{8931235}
\bibinfo{author}{Jin, S.}, \bibinfo{author}{Su, Y.}, \bibinfo{author}{Gao, S.},
  \bibinfo{author}{Wu, F.}, \bibinfo{author}{Ma, Q.}, \bibinfo{author}{Xu, K.},
  \bibinfo{author}{Ma, Q.}, \bibinfo{author}{Hu, T.}, \bibinfo{author}{Liu,
  J.}, \bibinfo{author}{Pang, S.}, \bibinfo{author}{Guan, H.},
  \bibinfo{author}{Zhang, J.}, \bibinfo{author}{Guo, Q.}, \bibinfo{year}{2020}.
\newblock \bibinfo{title}{Separating the structural components of maize for
  field phenotyping using terrestrial lidar data and deep convolutional neural
  networks}.
\newblock \bibinfo{journal}{IEEE Transactions on Geoscience and Remote Sensing}
%Type = Article
\bibitem[{Jin et~al.(2019)Jin, Su, Wu, Pang, Gao, Hu, Liu and Guo}]{8468204}
\bibinfo{author}{Jin, S.}, \bibinfo{author}{Su, Y.}, \bibinfo{author}{Wu, F.},
  \bibinfo{author}{Pang, S.}, \bibinfo{author}{Gao, S.}, \bibinfo{author}{Hu,
  T.}, \bibinfo{author}{Liu, J.}, \bibinfo{author}{Guo, Q.},
  \bibinfo{year}{2019}.
\newblock \bibinfo{title}{Stem–leaf segmentation and phenotypic trait
  extraction of individual maize using terrestrial lidar data}.
\newblock \bibinfo{journal}{IEEE Transactions on Geoscience and Remote Sensing}
%Type = Article
\bibitem[{Jin et~al.(2021)Jin, Sun, Wu, Su, Li, Song, Xu, Ma, Baret, Jiang,
  Ding and Guo}]{JIN2021202}
\bibinfo{author}{Jin, S.}, \bibinfo{author}{Sun, X.}, \bibinfo{author}{Wu, F.},
  \bibinfo{author}{Su, Y.}, \bibinfo{author}{Li, Y.}, \bibinfo{author}{Song,
  S.}, \bibinfo{author}{Xu, K.}, \bibinfo{author}{Ma, Q.},
  \bibinfo{author}{Baret, F.}, \bibinfo{author}{Jiang, D.},
  \bibinfo{author}{Ding, Y.}, \bibinfo{author}{Guo, Q.}, \bibinfo{year}{2021}.
\newblock \bibinfo{title}{Lidar sheds new light on plant phenomics for plant
  breeding and management: Recent advances and future prospects}.
\newblock \bibinfo{journal}{ISPRS Journal of Photogrammetry and Remote Sensing}
%Type = Article
\bibitem[{Khanna et~al.(2019)Khanna, Schmid, Walter, Nieto, Siegwart and
  Liebisch}]{EschikonPlantStress}
\bibinfo{author}{Khanna, R.}, \bibinfo{author}{Schmid, L.},
  \bibinfo{author}{Walter, A.}, \bibinfo{author}{Nieto, J.},
  \bibinfo{author}{Siegwart, R.}, \bibinfo{author}{Liebisch, F.},
  \bibinfo{year}{2019}.
\newblock \bibinfo{title}{A spatio temporal spectral framework for plant stress
  phenotyping}.
\newblock \bibinfo{journal}{Plant Methods}
%Type = Inproceedings
\bibitem[{Kolodiazhnyi et~al.(2024a)Kolodiazhnyi, Vorontsova and
  Konushin}]{Top-DownBeatsBottom-Up}
\bibinfo{author}{Kolodiazhnyi, M.}, \bibinfo{author}{Vorontsova, A.},
  \bibinfo{author}{Konushin, A.}, \bibinfo{year}{2024}a.
\newblock \bibinfo{title}{Top-down beats bottom-up in 3d instance
  segmentation},
%Type = Inproceedings
\bibitem[{Kolodiazhnyi et~al.(2024b)Kolodiazhnyi, Vorontsova, Konushin and
  Rukhovich}]{kolodiazhnyi2024oneformer3d}
\bibinfo{author}{Kolodiazhnyi, M.}, \bibinfo{author}{Vorontsova, A.},
  \bibinfo{author}{Konushin, A.}, \bibinfo{author}{Rukhovich, D.},
  \bibinfo{year}{2024}b.
\newblock \bibinfo{title}{Oneformer3d: One transformer for unified point cloud
  segmentation}, in: \bibinfo{booktitle}{Proceedings of the IEEE/CVF Conference
  on Computer Vision and Pattern Recognition},
%Type = Inproceedings
\bibitem[{Lambourne et~al.(2021)Lambourne, Willis, Jayaraman, Sanghi, Meltzer
  and Shayani}]{lambourne2021brepnet}
\bibinfo{author}{Lambourne, J.G.}, \bibinfo{author}{Willis, K.D.},
  \bibinfo{author}{Jayaraman, P.K.}, \bibinfo{author}{Sanghi, A.},
  \bibinfo{author}{Meltzer, P.}, \bibinfo{author}{Shayani, H.},
  \bibinfo{year}{2021}.
\newblock \bibinfo{title}{Brepnet: A topological message passing system for
  solid models}, in: \bibinfo{booktitle}{Proceedings of the IEEE/CVF Conference
  on Computer Vision and Pattern Recognition (CVPR)},
%Type = Inproceedings
\bibitem[{Landrieu and Boussaha(2019)}]{Landrieu2019inproceedings}
\bibinfo{author}{Landrieu, L.}, \bibinfo{author}{Boussaha, M.},
  \bibinfo{year}{2019}.
\newblock
%Type = Inproceedings
\bibitem[{Landrieu and Simonovsky(2018)}]{SuperpointGraphsinproceedings}
\bibinfo{author}{Landrieu, L.}, \bibinfo{author}{Simonovsky, M.},
  \bibinfo{year}{2018}.
\newblock
%Type = Article
\bibitem[{Lee et~al.(2023)Lee, Li and Benes}]{10.1145/3627101}
\bibinfo{author}{Lee, J.J.}, \bibinfo{author}{Li, B.}, \bibinfo{author}{Benes,
  B.}, \bibinfo{year}{2023}.
\newblock \bibinfo{title}{Latent l-systems: Transformer-based tree generator}.
\newblock \bibinfo{journal}{ACM Trans. Graph.}
%Type = Article
\bibitem[{Li et~al.(2025a)Li, Ahmed and Wang}]{LI2025100002}
\bibinfo{author}{Li, D.}, \bibinfo{author}{Ahmed, F.}, \bibinfo{author}{Wang,
  Z.}, \bibinfo{year}{2025}a.
\newblock \bibinfo{title}{3d-nod: 3d new organ detection in plant growth by a
  spatiotemporal point cloud deep segmentation framework}.
\newblock \bibinfo{journal}{Plant Phenomics}
%Type = Article
\bibitem[{Li et~al.(2022a)Li, Li, Xiang and Pan}]{doi:10.34133/2022/9787643}
\bibinfo{author}{Li, D.}, \bibinfo{author}{Li, J.}, \bibinfo{author}{Xiang,
  S.}, \bibinfo{author}{Pan, A.}, \bibinfo{year}{2022}a.
\newblock \bibinfo{title}{Psegnet: Simultaneous semantic and instance
  segmentation for point clouds of plants}.
\newblock \bibinfo{journal}{Plant Phenomics}
%Type = Article
\bibitem[{Li et~al.(2024a)Li, Liu, Xu and Jin}]{LI2024109435}
\bibinfo{author}{Li, D.}, \bibinfo{author}{Liu, L.}, \bibinfo{author}{Xu, S.},
  \bibinfo{author}{Jin, S.}, \bibinfo{year}{2024}a.
\newblock \bibinfo{title}{Trackplant3d: 3d organ growth tracking framework for
  organ-level dynamic phenotyping}.
\newblock \bibinfo{journal}{Computers and Electronics in Agriculture}
%Type = Article
\bibitem[{Li et~al.(2022b)Li, Shi, Li, Chen, Zhang, Xiang and Jin}]{Plantnet}
\bibinfo{author}{Li, D.}, \bibinfo{author}{Shi, G.}, \bibinfo{author}{Li, J.},
  \bibinfo{author}{Chen, Y.}, \bibinfo{author}{Zhang, S.},
  \bibinfo{author}{Xiang, S.}, \bibinfo{author}{Jin, S.},
  \bibinfo{year}{2022}b.
\newblock \bibinfo{title}{Plantnet: A dual-function point cloud segmentation
  network for multiple plant species}.
\newblock \bibinfo{journal}{ISPRS Journal of Photogrammetry and Remote Sensing}
%Type = Article
\bibitem[{Li et~al.(2023)Li, Wei and Zhu}]{LiDaweiDownSamplingArticle}
\bibinfo{author}{Li, D.}, \bibinfo{author}{Wei, Y.}, \bibinfo{author}{Zhu, R.},
  \bibinfo{year}{2023}.
\newblock \bibinfo{title}{A comparative study on point cloud down-sampling
  strategies for deep learning-based crop organ segmentation}.
\newblock \bibinfo{journal}{Plant Methods}
%Type = Article
\bibitem[{Li et~al.(2024b)Li, Zhou and Wei}]{LI2024172}
\bibinfo{author}{Li, D.}, \bibinfo{author}{Zhou, Z.}, \bibinfo{author}{Wei,
  Y.}, \bibinfo{year}{2024}b.
\newblock \bibinfo{title}{Unsupervised shape-aware som down-sampling for plant
  point clouds}.
\newblock \bibinfo{journal}{ISPRS Journal of Photogrammetry and Remote Sensing}
%Type = Inproceedings
\bibitem[{Li et~al.(2024c)Li, Park, Reberg-Horton, Mirsky, Lobaton and
  Xiang}]{10678093}
\bibinfo{author}{Li, X.}, \bibinfo{author}{Park, J.},
  \bibinfo{author}{Reberg-Horton, C.}, \bibinfo{author}{Mirsky, S.},
  \bibinfo{author}{Lobaton, E.}, \bibinfo{author}{Xiang, L.},
  \bibinfo{year}{2024}c.
\newblock \bibinfo{title}{Photorealistic arm robot simulation for 3d plant
  reconstruction and automatic annotation using unreal engine 5}, in:
  \bibinfo{booktitle}{2024 IEEE/CVF Conference on Computer Vision and Pattern
  Recognition Workshops (CVPRW)},
%Type = Article
\bibitem[{Li et~al.(2022c)Li, Wen, Miao, Wu, Yu, Wang, Guo and
  Zhao}]{LI2022106702}
\bibinfo{author}{Li, Y.}, \bibinfo{author}{Wen, W.}, \bibinfo{author}{Miao,
  T.}, \bibinfo{author}{Wu, S.}, \bibinfo{author}{Yu, Z.},
  \bibinfo{author}{Wang, X.}, \bibinfo{author}{Guo, X.}, \bibinfo{author}{Zhao,
  C.}, \bibinfo{year}{2022}c.
\newblock \bibinfo{title}{Automatic organ-level point cloud segmentation of
  maize shoots by integrating high-throughput data acquisition and deep
  learning}.
\newblock \bibinfo{journal}{Computers and Electronics in Agriculture}
%Type = Inproceedings
\bibitem[{Li and Su(2025)}]{10.1117/12.3061446}
\bibinfo{author}{Li, Z.}, \bibinfo{author}{Su, Y.}, \bibinfo{year}{2025}.
\newblock \bibinfo{title}{{A 3D point cloud instance segmentation method for
  strawberry based on SGC}}, in: \bibinfo{editor}{Xu, X.}, \bibinfo{editor}{bin
  Mohd~Zain, A.} (Eds.), \bibinfo{booktitle}{International Conference on
  Computer Graphics, Artificial Intelligence, and Data Processing (ICCAID
  2024)}, \bibinfo{organization}{International Society for Optics and
  Photonics}. \bibinfo{publisher}{SPIE}.
%Type = Article
\bibitem[{Li et~al.(2025b)Li, Wang, Su and Yu}]{10872974}
\bibinfo{author}{Li, Z.}, \bibinfo{author}{Wang, S.}, \bibinfo{author}{Su, Y.},
  \bibinfo{author}{Yu, D.}, \bibinfo{year}{2025}b.
\newblock \bibinfo{title}{A method for measuring strawberry leaf area based on
  three-dimensional point cloud instance segmentation}.
\newblock \bibinfo{journal}{IEEE Access}
%Type = Article
\bibitem[{Liang et~al.(2016)Liang, Kankare, Hyyppä, Wang, Kukko, Haggrén, Yu,
  Kaartinen, Jaakkola, Guan, Holopainen and Vastaranta}]{LIANG201663}
\bibinfo{author}{Liang, X.}, \bibinfo{author}{Kankare, V.},
  \bibinfo{author}{Hyyppä, J.}, \bibinfo{author}{Wang, Y.},
  \bibinfo{author}{Kukko, A.}, \bibinfo{author}{Haggrén, H.},
  \bibinfo{author}{Yu, X.}, \bibinfo{author}{Kaartinen, H.},
  \bibinfo{author}{Jaakkola, A.}, \bibinfo{author}{Guan, F.},
  \bibinfo{author}{Holopainen, M.}, \bibinfo{author}{Vastaranta, M.},
  \bibinfo{year}{2016}.
\newblock \bibinfo{title}{Terrestrial laser scanning in forest inventories}.
\newblock \bibinfo{journal}{ISPRS Journal of Photogrammetry and Remote Sensing}
\newblock \bibinfo{note}{Theme issue 'State-of-the-art in photogrammetry,
  remote sensing and spatial information science'}.
%Type = Inproceedings
\bibitem[{Liang et~al.(2021)Liang, Li, Xu, Tan and Jia}]{liang2021instance}
\bibinfo{author}{Liang, Z.}, \bibinfo{author}{Li, Z.}, \bibinfo{author}{Xu,
  S.}, \bibinfo{author}{Tan, M.}, \bibinfo{author}{Jia, K.},
  \bibinfo{year}{2021}.
\newblock \bibinfo{title}{Instance segmentation in 3d scenes using semantic
  superpoint tree networks}, in: \bibinfo{booktitle}{Proceedings of the
  IEEE/CVF International Conference on Computer Vision},
%Type = Article
\bibitem[{Liao et~al.(2021)Liao, Xie and Geiger}]{Liao2021ARXIV}
\bibinfo{author}{Liao, Y.}, \bibinfo{author}{Xie, J.}, \bibinfo{author}{Geiger,
  A.}, \bibinfo{year}{2021}.
\newblock \bibinfo{title}{{KITTI}-360: A novel dataset and benchmarks for urban
  scene understanding in 2d and 3d}.
\newblock
%Type = Article
\bibitem[{Lindenmayer(1968)}]{LINDENMAYER1968280}
\bibinfo{author}{Lindenmayer, A.}, \bibinfo{year}{1968}.
\newblock \bibinfo{title}{Mathematical models for cellular interactions in
  development i. filaments with one-sided inputs}.
\newblock \bibinfo{journal}{Journal of Theoretical Biology}
%Type = Inproceedings
\bibitem[{Liu et~al.(2019)Liu, Tang, Lin and Han}]{liu2019pvcnn}
\bibinfo{author}{Liu, Z.}, \bibinfo{author}{Tang, H.}, \bibinfo{author}{Lin,
  Y.}, \bibinfo{author}{Han, S.}, \bibinfo{year}{2019}.
\newblock \bibinfo{title}{Point-voxel cnn for efficient 3d deep learning},
%Type = Inproceedings
\bibitem[{Long et~al.(2015)Long, Shelhamer and Darrell}]{7298965}
\bibinfo{author}{Long, J.}, \bibinfo{author}{Shelhamer, E.},
  \bibinfo{author}{Darrell, T.}, \bibinfo{year}{2015}.
\newblock \bibinfo{title}{{ Fully convolutional networks for semantic
  segmentation }}, in: \bibinfo{booktitle}{2015 IEEE Conference on Computer
  Vision and Pattern Recognition (CVPR)}, \bibinfo{publisher}{IEEE Computer
  Society}, \bibinfo{address}{Los Alamitos, CA, USA}.
%Type = Article
\bibitem[{Lowry et~al.(2024)Lowry, Giraldo, Steinmetz, Avellan, Demirer,
  Ristroph, Wang, Hendren, Alabi, Caparco et~al.}]{lowry2024towards}
\bibinfo{author}{Lowry, G.V.}, \bibinfo{author}{Giraldo, J.P.},
  \bibinfo{author}{Steinmetz, N.F.}, \bibinfo{author}{Avellan, A.},
  \bibinfo{author}{Demirer, G.S.}, \bibinfo{author}{Ristroph, K.D.},
  \bibinfo{author}{Wang, G.J.}, \bibinfo{author}{Hendren, C.O.},
  \bibinfo{author}{Alabi, C.A.}, \bibinfo{author}{Caparco, A.}, et~al.,
  \bibinfo{year}{2024}.
\newblock \bibinfo{title}{Towards realizing nano-enabled precision delivery in
  plants}.
\newblock \bibinfo{journal}{Nature Nanotechnology}
%Type = Article
\bibitem[{Luo et~al.(2023)Luo, Jiang, Yang, Samy, Lefsrud, Hoyos-Villegas and
  Sun}]{doi:10.34133/plantphenomics.0080}
\bibinfo{author}{Luo, L.}, \bibinfo{author}{Jiang, X.}, \bibinfo{author}{Yang,
  Y.}, \bibinfo{author}{Samy, E.R.A.}, \bibinfo{author}{Lefsrud, M.},
  \bibinfo{author}{Hoyos-Villegas, V.}, \bibinfo{author}{Sun, S.},
  \bibinfo{year}{2023}.
\newblock \bibinfo{title}{Eff-3dpseg: 3d organ-level plant shoot segmentation
  using annotation-efficient deep learning}.
\newblock \bibinfo{journal}{Plant Phenomics}
%Type = Article
\bibitem[{Ma et~al.(2019)Ma, Zhu, Guan, Feng, Yu and Liu}]{rs11091085}
\bibinfo{author}{Ma, X.}, \bibinfo{author}{Zhu, K.}, \bibinfo{author}{Guan,
  H.}, \bibinfo{author}{Feng, J.}, \bibinfo{author}{Yu, S.},
  \bibinfo{author}{Liu, G.}, \bibinfo{year}{2019}.
\newblock \bibinfo{title}{High-throughput phenotyping analysis of potted
  soybean plants using colorized depth images based on a proximal platform}.
\newblock \bibinfo{journal}{Remote Sensing}
%Type = Article
\bibitem[{Ma et~al.(2023)Ma, Du, Xie, Sun, Fang, Jiang and
  Cen}]{doi:10.34133/plantphenomics.0027}
\bibinfo{author}{Ma, Z.}, \bibinfo{author}{Du, R.}, \bibinfo{author}{Xie, J.},
  \bibinfo{author}{Sun, D.}, \bibinfo{author}{Fang, H.},
  \bibinfo{author}{Jiang, L.}, \bibinfo{author}{Cen, H.}, \bibinfo{year}{2023}.
\newblock \bibinfo{title}{Phenotyping of silique morphology in oilseed rape
  using skeletonization with hierarchical segmentation}.
\newblock \bibinfo{journal}{Plant Phenomics}
%Type = Inproceedings
\bibitem[{Marks et~al.(2024)Marks, Bömer, Magistri, Sag, Behley and
  Stachniss}]{10802820}
\bibinfo{author}{Marks, E.}, \bibinfo{author}{Bömer, J.},
  \bibinfo{author}{Magistri, F.}, \bibinfo{author}{Sag, A.},
  \bibinfo{author}{Behley, J.}, \bibinfo{author}{Stachniss, C.},
  \bibinfo{year}{2024}.
\newblock \bibinfo{title}{Bonnbeetclouds3d: A dataset towards point cloud-based
  organ-level phenotyping of sugar beet plants under real field conditions},
  in: \bibinfo{booktitle}{2024 IEEE/RSJ International Conference on Intelligent
  Robots and Systems (IROS)},
%Type = Inproceedings
\bibitem[{Maturana and Scherer(2015)}]{7353481}
\bibinfo{author}{Maturana, D.}, \bibinfo{author}{Scherer, S.},
  \bibinfo{year}{2015}.
\newblock \bibinfo{title}{Voxnet: A 3d convolutional neural network for
  real-time object recognition}, in: \bibinfo{booktitle}{2015 IEEE/RSJ
  International Conference on Intelligent Robots and Systems (IROS)},
%Type = Inproceedings
\bibitem[{M{\v{e}}ch and Prusinkiewicz(1996)}]{mvech1996visual}
\bibinfo{author}{M{\v{e}}ch, R.}, \bibinfo{author}{Prusinkiewicz, P.},
  \bibinfo{year}{1996}.
\newblock \bibinfo{title}{Visual models of plants interacting with their
  environment}, in: \bibinfo{booktitle}{Proceedings of the 23rd annual
  conference on Computer graphics and interactive techniques},
%Type = Inproceedings
\bibitem[{Mertoğlu et~al.(2023)Mertoğlu, Şalk, Sarıkaya, Turgut,
  Evrenosoğlu, Çevikalp, Gerek, Dutağacı and Rousseau}]{10223838}
\bibinfo{author}{Mertoğlu, K.}, \bibinfo{author}{Şalk, Y.},
  \bibinfo{author}{Sarıkaya, S.K.}, \bibinfo{author}{Turgut, K.},
  \bibinfo{author}{Evrenosoğlu, Y.}, \bibinfo{author}{Çevikalp, H.},
  \bibinfo{author}{Gerek, {\"O}.N.}, \bibinfo{author}{Dutağacı, H.},
  \bibinfo{author}{Rousseau, D.}, \bibinfo{year}{2023}.
\newblock \bibinfo{title}{Planest-3d: A new annotated data set of 3d color
  point clouds of plants}, in: \bibinfo{booktitle}{2023 31st Signal Processing
  and Communications Applications Conference (SIU)},
%Type = Article
\bibitem[{Mirande et~al.(2022)Mirande, Godin, Tisserand, Charlaix, Besnard and
  Hétroy-Wheeler}]{10.3389/fpls.2022.1012669}
\bibinfo{author}{Mirande, K.}, \bibinfo{author}{Godin, C.},
  \bibinfo{author}{Tisserand, M.}, \bibinfo{author}{Charlaix, J.},
  \bibinfo{author}{Besnard, F.}, \bibinfo{author}{Hétroy-Wheeler, F.},
  \bibinfo{year}{2022}.
\newblock \bibinfo{title}{A graph-based approach for simultaneous semantic and
  instance segmentation of plant 3d point clouds}.
\newblock \bibinfo{journal}{Frontiers in Plant Science}
%Type = Article
\bibitem[{Mkaouar et~al.(2021)Mkaouar, Kallel, Rabah and Chahed}]{9576621}
\bibinfo{author}{Mkaouar, A.}, \bibinfo{author}{Kallel, A.},
  \bibinfo{author}{Rabah, Z.B.}, \bibinfo{author}{Chahed, T.S.},
  \bibinfo{year}{2021}.
\newblock \bibinfo{title}{Joint estimation of leaf area density and leaf angle
  distribution using tls point cloud for forest stands}.
\newblock \bibinfo{journal}{IEEE Journal of Selected Topics in Applied Earth
  Observations and Remote Sensing}
%Type = Inproceedings
\bibitem[{Mo et~al.(2019)Mo, Zhu, Chang, Yi, Tripathi, Guibas and
  Su}]{Mo_2019_CVPR}
\bibinfo{author}{Mo, K.}, \bibinfo{author}{Zhu, S.}, \bibinfo{author}{Chang,
  A.X.}, \bibinfo{author}{Yi, L.}, \bibinfo{author}{Tripathi, S.},
  \bibinfo{author}{Guibas, L.J.}, \bibinfo{author}{Su, H.},
  \bibinfo{year}{2019}.
\newblock \bibinfo{title}{{PartNet}: A large-scale benchmark for fine-grained
  and hierarchical part-level {3D} object understanding},
%Type = Article
\bibitem[{Niemeyer et~al.(2012)Niemeyer, Rottensteiner and
  Soergel}]{Niemeyerarticle}
\bibinfo{author}{Niemeyer, J.}, \bibinfo{author}{Rottensteiner, F.},
  \bibinfo{author}{Soergel, U.}, \bibinfo{year}{2012}.
\newblock \bibinfo{title}{Conditional random fields for lidar point cloud
  classification in complex urban areas}.
\newblock \bibinfo{journal}{ISPRS Annals of Photogrammetry, Remote Sensing and
  Spatial Information Sciences}
%Type = Article
\bibitem[{Patel et~al.(2023)Patel, Park, Lee, Priya, Kim, Joshi, Arief, Kim,
  Baek and Cho}]{10243159}
\bibinfo{author}{Patel, A.K.}, \bibinfo{author}{Park, E.S.},
  \bibinfo{author}{Lee, H.}, \bibinfo{author}{Priya, G.G.L.},
  \bibinfo{author}{Kim, H.}, \bibinfo{author}{Joshi, R.},
  \bibinfo{author}{Arief, M.A.A.}, \bibinfo{author}{Kim, M.S.},
  \bibinfo{author}{Baek, I.}, \bibinfo{author}{Cho, B.K.},
  \bibinfo{year}{2023}.
\newblock \bibinfo{title}{Deep learning-based plant organ segmentation and
  phenotyping of sorghum plants using lidar point cloud}.
\newblock \bibinfo{journal}{IEEE Journal of Selected Topics in Applied Earth
  Observations and Remote Sensing}
%Type = Inproceedings
\bibitem[{Prusinkiewicz(1986)}]{10.5555/16564.16608}
\bibinfo{author}{Prusinkiewicz, P.}, \bibinfo{year}{1986}.
\newblock \bibinfo{title}{Graphical applications of l-systems}, in:
  \bibinfo{booktitle}{Proceedings on Graphics Interface '86/Vision Interface
  '86}, \bibinfo{publisher}{Canadian Information Processing Society},
  \bibinfo{address}{CAN}.
%Type = Inproceedings
\bibitem[{Prusinkiewicz et~al.(1993)Prusinkiewicz, Hammel and
  Mjolsness}]{10.1145/166117.166161}
\bibinfo{author}{Prusinkiewicz, P.}, \bibinfo{author}{Hammel, M.S.},
  \bibinfo{author}{Mjolsness, E.}, \bibinfo{year}{1993}.
\newblock \bibinfo{title}{Animation of plant development}, in:
  \bibinfo{booktitle}{Proceedings of the 20th Annual Conference on Computer
  Graphics and Interactive Techniques}, \bibinfo{publisher}{Association for
  Computing Machinery}, \bibinfo{address}{New York, NY, USA}.
%Type = Book
\bibitem[{Prusinkiewicz and Lindenmayer(1990)}]{10.5555/83596}
\bibinfo{author}{Prusinkiewicz, P.}, \bibinfo{author}{Lindenmayer, A.},
  \bibinfo{year}{1990}.
\newblock \bibinfo{title}{The algorithmic beauty of plants}.
\newblock
%Type = Inproceedings
\bibitem[{Qi et~al.(2017a)Qi, Su, Mo and Guibas}]{qi2017pointnet}
\bibinfo{author}{Qi, C.R.}, \bibinfo{author}{Su, H.}, \bibinfo{author}{Mo, K.},
  \bibinfo{author}{Guibas, L.J.}, \bibinfo{year}{2017}a.
\newblock \bibinfo{title}{Pointnet: Deep learning on point sets for 3d
  classification and segmentation}, in: \bibinfo{booktitle}{Proceedings of the
  IEEE conference on computer vision and pattern recognition},
%Type = Article
\bibitem[{Qi et~al.(2017b)Qi, Yi, Su and Guibas}]{qi2017PointNetplusplus}
\bibinfo{author}{Qi, C.R.}, \bibinfo{author}{Yi, L.}, \bibinfo{author}{Su, H.},
  \bibinfo{author}{Guibas, L.J.}, \bibinfo{year}{2017}b.
\newblock \bibinfo{title}{Pointnet++: Deep hierarchical feature learning on
  point sets in a metric space}.
\newblock
%Type = Article
\bibitem[{Qian et~al.(2023)Qian, Huang, Xie, Ye, Guo, Pan, Jin, Xie, Jiao,
  Zhang, Ruan, Xu, Zhang and Nie}]{QIAN2023108124}
\bibinfo{author}{Qian, B.}, \bibinfo{author}{Huang, W.}, \bibinfo{author}{Xie,
  D.}, \bibinfo{author}{Ye, H.}, \bibinfo{author}{Guo, A.},
  \bibinfo{author}{Pan, Y.}, \bibinfo{author}{Jin, Y.}, \bibinfo{author}{Xie,
  Q.}, \bibinfo{author}{Jiao, Q.}, \bibinfo{author}{Zhang, B.},
  \bibinfo{author}{Ruan, C.}, \bibinfo{author}{Xu, T.}, \bibinfo{author}{Zhang,
  Y.}, \bibinfo{author}{Nie, T.}, \bibinfo{year}{2023}.
\newblock \bibinfo{title}{Coupled maize model: A 4d maize growth model based on
  growing degree days}.
\newblock \bibinfo{journal}{Computers and Electronics in Agriculture}
%Type = Article
\bibitem[{Qiu et~al.(2019)Qiu, Sun, Bai, Wang, Fan, Wang, Meng, Li and
  Cong}]{10.3389/fpls.2019.00554}
\bibinfo{author}{Qiu, Q.}, \bibinfo{author}{Sun, N.}, \bibinfo{author}{Bai,
  H.}, \bibinfo{author}{Wang, N.}, \bibinfo{author}{Fan, Z.},
  \bibinfo{author}{Wang, Y.}, \bibinfo{author}{Meng, Z.}, \bibinfo{author}{Li,
  B.}, \bibinfo{author}{Cong, Y.}, \bibinfo{year}{2019}.
\newblock \bibinfo{title}{Field-based high-throughput phenotyping for maize
  plant using 3d lidar point cloud generated with a “phenomobile”}.
\newblock \bibinfo{journal}{Frontiers in Plant Science}
%Type = Article
\bibitem[{Qiu et~al.(2021)Qiu, Miao, Zhang and Li}]{QIU2021106551}
\bibinfo{author}{Qiu, R.}, \bibinfo{author}{Miao, Y.}, \bibinfo{author}{Zhang,
  M.}, \bibinfo{author}{Li, H.}, \bibinfo{year}{2021}.
\newblock \bibinfo{title}{Detection of the 3d temperature characteristics of
  maize under water stress using thermal and rgb-d cameras}.
\newblock \bibinfo{journal}{Computers and Electronics in Agriculture}
%Type = Misc
\bibitem[{Qiu et~al.(2025)Qiu, Du, Spine, Cheng and
  Jiang}]{qiu2025joint3dpointcloud}
\bibinfo{author}{Qiu, T.}, \bibinfo{author}{Du, R.}, \bibinfo{author}{Spine,
  N.}, \bibinfo{author}{Cheng, L.}, \bibinfo{author}{Jiang, Y.},
  \bibinfo{year}{2025}.
\newblock
%Type = Article
\bibitem[{Qiu et~al.(2024a)Qiu, Wang, Han, Kuehn, Cheng, Meng, Xu, Xu and
  Yu}]{doi:10.34133/plantphenomics.0179}
\bibinfo{author}{Qiu, T.}, \bibinfo{author}{Wang, T.}, \bibinfo{author}{Han,
  T.}, \bibinfo{author}{Kuehn, K.}, \bibinfo{author}{Cheng, L.},
  \bibinfo{author}{Meng, C.}, \bibinfo{author}{Xu, X.}, \bibinfo{author}{Xu,
  K.}, \bibinfo{author}{Yu, J.}, \bibinfo{year}{2024}a.
\newblock \bibinfo{title}{Appleqsm: Geometry-based 3d characterization of apple
  tree architecture in orchards}.
\newblock \bibinfo{journal}{Plant Phenomics}
%Type = Inproceedings
\bibitem[{Qiu et~al.(2024b)Qiu, Zoubi, Spine, Cheng and Jiang}]{10803058}
\bibinfo{author}{Qiu, T.}, \bibinfo{author}{Zoubi, A.}, \bibinfo{author}{Spine,
  N.}, \bibinfo{author}{Cheng, L.}, \bibinfo{author}{Jiang, Y.},
  \bibinfo{year}{2024}b.
\newblock \bibinfo{title}{(real2sim)-1: 3d branch point cloud completion for
  robotic pruning in apple orchards}, in: \bibinfo{booktitle}{2024 IEEE/RSJ
  International Conference on Intelligent Robots and Systems (IROS)},
%Type = Article
\bibitem[{Rabbani et~al.(2006)Rabbani, Heuvel and Vosselman}]{Rabbaniarticle}
\bibinfo{author}{Rabbani, T.}, \bibinfo{author}{Heuvel, F.},
  \bibinfo{author}{Vosselman, G.}, \bibinfo{year}{2006}.
\newblock \bibinfo{title}{Segmentation of point clouds using smoothness
  constraint}.
\newblock \bibinfo{journal}{International Archives of Photogrammetry, Remote
  Sensing and Spatial Information Sciences}
%Type = Inproceedings
\bibitem[{Ren and Sudderth(2016)}]{7780538}
\bibinfo{author}{Ren, Z.}, \bibinfo{author}{Sudderth, E.B.},
  \bibinfo{year}{2016}.
\newblock \bibinfo{title}{Three-dimensional object detection and layout
  prediction using clouds of oriented gradients}, in: \bibinfo{booktitle}{2016
  IEEE Conference on Computer Vision and Pattern Recognition (CVPR)},
%Type = Inproceedings
\bibitem[{Rusu et~al.(2008)Rusu, Marton, Blodow and Beetz}]{Rusuinproceedings}
\bibinfo{author}{Rusu, R.}, \bibinfo{author}{Marton, Z.},
  \bibinfo{author}{Blodow, N.}, \bibinfo{author}{Beetz, M.},
  \bibinfo{year}{2008}.
\newblock \bibinfo{title}{Learning informative point classes for the
  acquisition of object model maps},
%Type = Article
\bibitem[{Saeed et~al.(2023)Saeed, Sun, Rodriguez-Sanchez, Snider, Liu and
  Li}]{Cotton3D}
\bibinfo{author}{Saeed, F.}, \bibinfo{author}{Sun, S.},
  \bibinfo{author}{Rodriguez-Sanchez, J.}, \bibinfo{author}{Snider, J.},
  \bibinfo{author}{Liu, T.}, \bibinfo{author}{Li, C.}, \bibinfo{year}{2023}.
\newblock \bibinfo{title}{Cotton plant part 3d segmentation and architectural
  trait extraction using point voxel convolutional neural networks}.
\newblock \bibinfo{journal}{Plant methods}
%Type = Article
\bibitem[{Schult et~al.(2023)Schult, Engelmann, Hermans, Litany, Tang and
  Leibe}]{Schult23ICRA}
\bibinfo{author}{Schult, J.}, \bibinfo{author}{Engelmann, F.},
  \bibinfo{author}{Hermans, A.}, \bibinfo{author}{Litany, O.},
  \bibinfo{author}{Tang, S.}, \bibinfo{author}{Leibe, B.},
  \bibinfo{year}{2023}.
\newblock
%Type = Article
\bibitem[{Schunck et~al.(2021)Schunck, Magistri, Rosu, Cornelißen, Chebrolu,
  Paulus, Léon, Behnke, Stachniss, Kuhlmann and Klingbeil}]{Pheno4D}
\bibinfo{author}{Schunck, D.}, \bibinfo{author}{Magistri, F.},
  \bibinfo{author}{Rosu, R.A.}, \bibinfo{author}{Cornelißen, A.},
  \bibinfo{author}{Chebrolu, N.}, \bibinfo{author}{Paulus, S.},
  \bibinfo{author}{Léon, J.}, \bibinfo{author}{Behnke, S.},
  \bibinfo{author}{Stachniss, C.}, \bibinfo{author}{Kuhlmann, H.},
  \bibinfo{author}{Klingbeil, L.}, \bibinfo{year}{2021}.
\newblock \bibinfo{title}{Pheno4d: A spatio-temporal dataset of maize and
  tomato plant point clouds for phenotyping and advanced plant analysis}.
\newblock \bibinfo{journal}{PLOS ONE}
%Type = Article
\bibitem[{Shen et~al.(2022)Shen, Huang, Wang, Li and Xi}]{rs14153842}
\bibinfo{author}{Shen, X.}, \bibinfo{author}{Huang, Q.}, \bibinfo{author}{Wang,
  X.}, \bibinfo{author}{Li, J.}, \bibinfo{author}{Xi, B.},
  \bibinfo{year}{2022}.
\newblock \bibinfo{title}{A deep learning-based method for extracting standing
  wood feature parameters from terrestrial laser scanning point clouds of
  artificially planted forest}.
\newblock \bibinfo{journal}{Remote Sensing}
%Type = Inproceedings
\bibitem[{Shen et~al.(2018)Shen, Feng, Yang and
  Tian}]{MiningPointCloudinproceedings}
\bibinfo{author}{Shen, Y.}, \bibinfo{author}{Feng, C.}, \bibinfo{author}{Yang,
  Y.}, \bibinfo{author}{Tian, D.}, \bibinfo{year}{2018}.
\newblock \bibinfo{title}{Mining point cloud local structures by kernel
  correlation and graph pooling},
%Type = Incollection
\bibitem[{Simpson(2019)}]{SIMPSON2019469}
\bibinfo{author}{Simpson, M.G.}, \bibinfo{year}{2019}.
\newblock \bibinfo{title}{9 - plant morphology}, in: \bibinfo{editor}{Simpson,
  M.G.} (Ed.), \bibinfo{booktitle}{Plant Systematics (Third Edition)}.
  \bibinfo{edition}{third edition} ed.. \bibinfo{publisher}{Academic Press},
%Type = Article
\bibitem[{Smith(1984)}]{10.1145/964965.808571}
\bibinfo{author}{Smith, A.R.}, \bibinfo{year}{1984}.
\newblock \bibinfo{title}{Plants, fractals, and formal languages}.
\newblock \bibinfo{journal}{SIGGRAPH Comput. Graph.}
%Type = Article
\bibitem[{Song et~al.(2025)Song, Wen, Wu and Guo}]{SONG2025296}
\bibinfo{author}{Song, H.}, \bibinfo{author}{Wen, W.}, \bibinfo{author}{Wu,
  S.}, \bibinfo{author}{Guo, X.}, \bibinfo{year}{2025}.
\newblock \bibinfo{title}{Comprehensive review on 3d point cloud segmentation
  in plants}.
\newblock \bibinfo{journal}{Artificial Intelligence in Agriculture}
%Type = Article
\bibitem[{Song et~al.(2023)Song, Li, Yang, Shao, Pu, Yang and
  Zhai}]{10.3389/fpls.2023.1097725}
\bibinfo{author}{Song, P.}, \bibinfo{author}{Li, Z.}, \bibinfo{author}{Yang,
  M.}, \bibinfo{author}{Shao, Y.}, \bibinfo{author}{Pu, Z.},
  \bibinfo{author}{Yang, W.}, \bibinfo{author}{Zhai, R.}, \bibinfo{year}{2023}.
\newblock \bibinfo{title}{Dynamic detection of three-dimensional crop
  phenotypes based on a consumer-grade rgb-d camera}.
\newblock \bibinfo{journal}{Frontiers in Plant Science}
%Type = Article
\bibitem[{Song et~al.(2019)Song, Srinivasan, Long and
  Zhu}]{SongQingfengArticle}
\bibinfo{author}{Song, Q.}, \bibinfo{author}{Srinivasan, V.},
  \bibinfo{author}{Long, S.}, \bibinfo{author}{Zhu, X.}, \bibinfo{year}{2019}.
\newblock \bibinfo{title}{Decomposition analysis on soybean productivity
  increase under elevated co2 using 3d canopy model reveals synergestic effects
  of co2 and light in photosynthesis}.
\newblock \bibinfo{journal}{Annals of botany}
%Type = Article
\bibitem[{Staedler et~al.(2013)Staedler, Masson and
  Schönenberger}]{10.1371/journal.pone.0075295}
\bibinfo{author}{Staedler, Y.M.}, \bibinfo{author}{Masson, D.},
  \bibinfo{author}{Schönenberger, J.}, \bibinfo{year}{2013}.
\newblock \bibinfo{title}{Plant tissues in 3d via x-ray tomography: Simple
  contrasting methods allow high resolution imaging}.
\newblock \bibinfo{journal}{PLOS ONE}
%Type = Article
\bibitem[{Stausberg et~al.(2024)Stausberg, Jost, Klingbeil and
  Kuhlmann}]{rs16244720}
\bibinfo{author}{Stausberg, L.}, \bibinfo{author}{Jost, B.},
  \bibinfo{author}{Klingbeil, L.}, \bibinfo{author}{Kuhlmann, H.},
  \bibinfo{year}{2024}.
\newblock \bibinfo{title}{A 3d surface reconstruction pipeline for plant
  phenotyping}.
\newblock \bibinfo{journal}{Remote Sensing}
%Type = Inproceedings
\bibitem[{Stava et~al.(2014)Stava, Pirk, Kratt, Chen, M{\v{e}}ch, Deussen and
  Benes}]{stava2014inverse}
\bibinfo{author}{Stava, O.}, \bibinfo{author}{Pirk, S.},
  \bibinfo{author}{Kratt, J.}, \bibinfo{author}{Chen, B.},
  \bibinfo{author}{M{\v{e}}ch, R.}, \bibinfo{author}{Deussen, O.},
  \bibinfo{author}{Benes, B.}, \bibinfo{year}{2014}.
\newblock \bibinfo{title}{Inverse procedural modelling of trees}, in:
  \bibinfo{booktitle}{Computer Graphics Forum}, \bibinfo{organization}{Wiley
  Online Library}.
%Type = Misc
\bibitem[{Sun et~al.(2022)Sun, Qing, Tan and Xu}]{2211.15766}
\bibinfo{author}{Sun, J.}, \bibinfo{author}{Qing, C.}, \bibinfo{author}{Tan,
  J.}, \bibinfo{author}{Xu, X.}, \bibinfo{year}{2022}.
\newblock
%Type = Inproceedings
\bibitem[{Sun et~al.(2020a)Sun, Kretzschmar, Dotiwalla, Chouard, Patnaik, Tsui,
  Guo, Zhou, Chai, Caine et~al.}]{sun2020scalability}
\bibinfo{author}{Sun, P.}, \bibinfo{author}{Kretzschmar, H.},
  \bibinfo{author}{Dotiwalla, X.}, \bibinfo{author}{Chouard, A.},
  \bibinfo{author}{Patnaik, V.}, \bibinfo{author}{Tsui, P.},
  \bibinfo{author}{Guo, J.}, \bibinfo{author}{Zhou, Y.}, \bibinfo{author}{Chai,
  Y.}, \bibinfo{author}{Caine, B.}, et~al., \bibinfo{year}{2020}a.
\newblock \bibinfo{title}{Scalability in perception for autonomous driving:
  Waymo open dataset}, in: \bibinfo{booktitle}{Proceedings of the IEEE/CVF
  conference on computer vision and pattern recognition},
%Type = Article
\bibitem[{Sun et~al.(2020b)Sun, Li, Chee, Paterson, Jiang, Xu, Robertson,
  Adhikari and Shehzad}]{SUN2020195}
\bibinfo{author}{Sun, S.}, \bibinfo{author}{Li, C.}, \bibinfo{author}{Chee,
  P.W.}, \bibinfo{author}{Paterson, A.H.}, \bibinfo{author}{Jiang, Y.},
  \bibinfo{author}{Xu, R.}, \bibinfo{author}{Robertson, J.S.},
  \bibinfo{author}{Adhikari, J.}, \bibinfo{author}{Shehzad, T.},
  \bibinfo{year}{2020}b.
\newblock \bibinfo{title}{Three-dimensional photogrammetric mapping of cotton
  bolls in situ based on point cloud segmentation and clustering}.
\newblock \bibinfo{journal}{ISPRS Journal of Photogrammetry and Remote Sensing}
%Type = Article
\bibitem[{Sun et~al.(2017)Sun, Li and Paterson}]{rs9040377}
\bibinfo{author}{Sun, S.}, \bibinfo{author}{Li, C.}, \bibinfo{author}{Paterson,
  A.H.}, \bibinfo{year}{2017}.
\newblock \bibinfo{title}{In-field high-throughput phenotyping of cotton plant
  height using lidar}.
\newblock \bibinfo{journal}{Remote Sensing}
%Type = Article
\bibitem[{Sun et~al.(2018)Sun, Li, Paterson, Jiang, Xu, Robertson, Snider and
  Chee}]{10.3389/fpls.2018.00016}
\bibinfo{author}{Sun, S.}, \bibinfo{author}{Li, C.}, \bibinfo{author}{Paterson,
  A.H.}, \bibinfo{author}{Jiang, Y.}, \bibinfo{author}{Xu, R.},
  \bibinfo{author}{Robertson, J.S.}, \bibinfo{author}{Snider, J.L.},
  \bibinfo{author}{Chee, P.W.}, \bibinfo{year}{2018}.
\newblock \bibinfo{title}{In-field high throughput phenotyping and cotton plant
  growth analysis using lidar}.
\newblock \bibinfo{journal}{Frontiers in Plant Science}
%Type = Inproceedings
\bibitem[{Sun et~al.(2023a)Sun, Rebain, Liao, Tankovich, Yazdani, Yi and
  Tagliasacchi}]{NeuralBFinproceedings}
\bibinfo{author}{Sun, W.}, \bibinfo{author}{Rebain, D.}, \bibinfo{author}{Liao,
  R.}, \bibinfo{author}{Tankovich, V.}, \bibinfo{author}{Yazdani, S.},
  \bibinfo{author}{Yi, K.}, \bibinfo{author}{Tagliasacchi, A.},
  \bibinfo{year}{2023}a.
\newblock \bibinfo{title}{Neuralbf: Neural bilateral filtering for top-down
  instance segmentation on point clouds},
%Type = Article
\bibitem[{Sun et~al.(2023b)Sun, Zhang, Sun, Li, Yu, Miao, Zhang, Li, Zhao, Hu,
  Xin, Chen and Zhu}]{Soybean}
\bibinfo{author}{Sun, Y.}, \bibinfo{author}{Zhang, Z.}, \bibinfo{author}{Sun,
  K.}, \bibinfo{author}{Li, S.}, \bibinfo{author}{Yu, J.},
  \bibinfo{author}{Miao, L.}, \bibinfo{author}{Zhang, Z.}, \bibinfo{author}{Li,
  Y.}, \bibinfo{author}{Zhao, H.}, \bibinfo{author}{Hu, Z.},
  \bibinfo{author}{Xin, D.}, \bibinfo{author}{Chen, Q.}, \bibinfo{author}{Zhu,
  R.}, \bibinfo{year}{2023}b.
\newblock \bibinfo{title}{Soybean-mvs: Annotated three-dimensional model
  dataset of whole growth period soybeans for 3d plant organ segmentation}.
\newblock \bibinfo{journal}{Agriculture}
%Type = Inproceedings
\bibitem[{Tang et~al.(2022)Tang, Liu, Li, Lin and Han}]{tang2022torchsparse}
\bibinfo{author}{Tang, H.}, \bibinfo{author}{Liu, Z.}, \bibinfo{author}{Li,
  X.}, \bibinfo{author}{Lin, Y.}, \bibinfo{author}{Han, S.},
  \bibinfo{year}{2022}.
\newblock \bibinfo{title}{{TorchSparse: Efficient Point Cloud Inference
  Engine}},
%Type = Inproceedings
\bibitem[{Tang et~al.(2020)Tang, Liu, Zhao, Lin, Lin, Wang and
  Han}]{tang2020searching}
\bibinfo{author}{Tang, H.}, \bibinfo{author}{Liu, Z.}, \bibinfo{author}{Zhao,
  S.}, \bibinfo{author}{Lin, Y.}, \bibinfo{author}{Lin, J.},
  \bibinfo{author}{Wang, H.}, \bibinfo{author}{Han, S.}, \bibinfo{year}{2020}.
\newblock \bibinfo{title}{Searching efficient 3d architectures with sparse
  point-voxel convolution},
%Type = Inproceedings
\bibitem[{Tang et~al.(2023a)Tang, Yang, Liu, Hong, Yu, Li, Dai, Wang and
  Han}]{tangandyang2023torchsparse++}
\bibinfo{author}{Tang, H.}, \bibinfo{author}{Yang, S.}, \bibinfo{author}{Liu,
  Z.}, \bibinfo{author}{Hong, K.}, \bibinfo{author}{Yu, Z.},
  \bibinfo{author}{Li, X.}, \bibinfo{author}{Dai, G.}, \bibinfo{author}{Wang,
  Y.}, \bibinfo{author}{Han, S.}, \bibinfo{year}{2023}a.
\newblock \bibinfo{title}{{TorchSparse++: Efficient Point Cloud Engine}},
%Type = Inproceedings
\bibitem[{Tang et~al.(2023b)Tang, Yang, Liu, Hong, Yu, Li, Dai, Wang and
  Han}]{tangandyang2023torchsparse}
\bibinfo{author}{Tang, H.}, \bibinfo{author}{Yang, S.}, \bibinfo{author}{Liu,
  Z.}, \bibinfo{author}{Hong, K.}, \bibinfo{author}{Yu, Z.},
  \bibinfo{author}{Li, X.}, \bibinfo{author}{Dai, G.}, \bibinfo{author}{Wang,
  Y.}, \bibinfo{author}{Han, S.}, \bibinfo{year}{2023}b.
\newblock \bibinfo{title}{Torchsparse++: Efficient training and inference
  framework for sparse convolution on gpus},
%Type = Article
\bibitem[{Tang et~al.(2024)Tang, Ao, Li, Huang, Xie, Wang, Wang and
  Guo}]{TANG2024103903}
\bibinfo{author}{Tang, S.}, \bibinfo{author}{Ao, Z.}, \bibinfo{author}{Li, Y.},
  \bibinfo{author}{Huang, H.}, \bibinfo{author}{Xie, L.},
  \bibinfo{author}{Wang, R.}, \bibinfo{author}{Wang, W.}, \bibinfo{author}{Guo,
  R.}, \bibinfo{year}{2024}.
\newblock \bibinfo{title}{Treenet3d : A large scale tree benchmark for 3d tree
  modeling, carbon storage estimation and tree segmentation}.
\newblock \bibinfo{journal}{International Journal of Applied Earth Observation
  and Geoinformation}
%Type = Inproceedings
\bibitem[{Tchapmi et~al.(2017)Tchapmi, Choy, Armeni, Gwak and
  Savarese}]{SEGCloudinproceedings}
\bibinfo{author}{Tchapmi, L.}, \bibinfo{author}{Choy, C.},
  \bibinfo{author}{Armeni, I.}, \bibinfo{author}{Gwak, J.},
  \bibinfo{author}{Savarese, S.}, \bibinfo{year}{2017}.
\newblock \bibinfo{title}{Segcloud: Semantic segmentation of 3d point clouds},
%Type = Article
\bibitem[{Terryn et~al.(2020)Terryn, Calders, Disney, Origo, Malhi, Newnham,
  Raumonen, {Å kerblom} and Verbeeck}]{TERRYN2020170}
\bibinfo{author}{Terryn, L.}, \bibinfo{author}{Calders, K.},
  \bibinfo{author}{Disney, M.}, \bibinfo{author}{Origo, N.},
  \bibinfo{author}{Malhi, Y.}, \bibinfo{author}{Newnham, G.},
  \bibinfo{author}{Raumonen, P.}, \bibinfo{author}{{Å kerblom}, M.},
  \bibinfo{author}{Verbeeck, H.}, \bibinfo{year}{2020}.
\newblock \bibinfo{title}{Tree species classification using structural features
  derived from terrestrial laser scanning}.
\newblock \bibinfo{journal}{ISPRS Journal of Photogrammetry and Remote Sensing}
%Type = Article
\bibitem[{Thomas et~al.(2019)Thomas, Qi, Deschaud, Marcotegui, Goulette and
  Guibas}]{thomas2019KPConv}
\bibinfo{author}{Thomas, H.}, \bibinfo{author}{Qi, C.R.},
  \bibinfo{author}{Deschaud, J.E.}, \bibinfo{author}{Marcotegui, B.},
  \bibinfo{author}{Goulette, F.}, \bibinfo{author}{Guibas, L.J.},
  \bibinfo{year}{2019}.
\newblock \bibinfo{title}{Kpconv: Flexible and deformable convolution for point
  clouds}.
\newblock
%Type = Article
\bibitem[{{Tiozzo Fasiolo} et~al.(2023){Tiozzo Fasiolo}, Scalera, Maset and
  Gasparetto}]{TIOZZOFASIOLO2023104514}
\bibinfo{author}{{Tiozzo Fasiolo}, D.}, \bibinfo{author}{Scalera, L.},
  \bibinfo{author}{Maset, E.}, \bibinfo{author}{Gasparetto, A.},
  \bibinfo{year}{2023}.
\newblock \bibinfo{title}{Towards autonomous mapping in agriculture: A review
  of supportive technologies for ground robotics}.
\newblock \bibinfo{journal}{Robotics and Autonomous Systems}
%Type = Inproceedings
\bibitem[{Tobin et~al.(2017)Tobin, Fong, Ray, Schneider, Zaremba and
  Abbeel}]{10.1109/IROS.2017.8202133}
\bibinfo{author}{Tobin, J.}, \bibinfo{author}{Fong, R.}, \bibinfo{author}{Ray,
  A.}, \bibinfo{author}{Schneider, J.}, \bibinfo{author}{Zaremba, W.},
  \bibinfo{author}{Abbeel, P.}, \bibinfo{year}{2017}.
\newblock \bibinfo{title}{Domain randomization for transferring deep neural
  networks from simulation to the real world}, in: \bibinfo{booktitle}{2017
  IEEE/RSJ International Conference on Intelligent Robots and Systems (IROS)},
  \bibinfo{publisher}{IEEE Press}.
%Type = Inbook
\bibitem[{Tong et~al.(2020)Tong, Gong, Tian and
  Shen}]{LearningandMemorizinginbook}
\bibinfo{author}{Tong, H.}, \bibinfo{author}{Gong, D.}, \bibinfo{author}{Tian,
  Z.}, \bibinfo{author}{Shen, C.}, \bibinfo{year}{2020}.
\newblock \bibinfo{title}{Learning and Memorizing Representative Prototypes for
  3D Point Cloud Semantic and Instance Segmentation}.
\newblock
%Type = Article
\bibitem[{Tracy et~al.(2020)Tracy, Nagel, Postma, Fassbender, Wasson and
  Watt}]{tracy2020crop}
\bibinfo{author}{Tracy, S.R.}, \bibinfo{author}{Nagel, K.A.},
  \bibinfo{author}{Postma, J.A.}, \bibinfo{author}{Fassbender, H.},
  \bibinfo{author}{Wasson, A.}, \bibinfo{author}{Watt, M.},
  \bibinfo{year}{2020}.
\newblock \bibinfo{title}{Crop improvement from phenotyping roots: highlights
  reveal expanding opportunities}.
\newblock \bibinfo{journal}{Trends in plant science}
%Type = Article
\bibitem[{Turgut et~al.(2022)Turgut, Dutagaci, Galopin and
  Rousseau}]{turgut:hal-03144153}
\bibinfo{author}{Turgut, K.}, \bibinfo{author}{Dutagaci, H.},
  \bibinfo{author}{Galopin, G.}, \bibinfo{author}{Rousseau, D.},
  \bibinfo{year}{2022}.
\newblock \bibinfo{title}{{Segmentation of structural parts of rosebush plants
  with 3d point-based deep learning methods}}.
\newblock \bibinfo{journal}{{Plant Methods}}
%Type = Inproceedings
\bibitem[{Varney et~al.(2020)Varney, Asari and Graehling}]{9150622}
\bibinfo{author}{Varney, N.}, \bibinfo{author}{Asari, V.K.},
  \bibinfo{author}{Graehling, Q.}, \bibinfo{year}{2020}.
\newblock \bibinfo{title}{Dales: A large-scale aerial lidar data set for
  semantic segmentation}, in: \bibinfo{booktitle}{2020 IEEE/CVF Conference on
  Computer Vision and Pattern Recognition Workshops (CVPRW)},
%Type = Article
\bibitem[{Vos et~al.(2010)Vos, Evers, Buck-Sorlin, Andrieu, Chelle and
  De~Visser}]{vos2010functional}
\bibinfo{author}{Vos, J.}, \bibinfo{author}{Evers, J.B.},
  \bibinfo{author}{Buck-Sorlin, G.H.}, \bibinfo{author}{Andrieu, B.},
  \bibinfo{author}{Chelle, M.}, \bibinfo{author}{De~Visser, P.H.},
  \bibinfo{year}{2010}.
\newblock \bibinfo{title}{Functional--structural plant modelling: a new
  versatile tool in crop science}.
\newblock \bibinfo{journal}{Journal of experimental Botany}
%Type = Inproceedings
\bibitem[{Vu et~al.(2022)Vu, Kim, Luu, Nguyen and Yoo}]{vu2022softgroup}
\bibinfo{author}{Vu, T.}, \bibinfo{author}{Kim, K.}, \bibinfo{author}{Luu,
  T.M.}, \bibinfo{author}{Nguyen, X.T.}, \bibinfo{author}{Yoo, C.D.},
  \bibinfo{year}{2022}.
\newblock \bibinfo{title}{Softgroup for 3d instance segmentation on 3d point
  clouds},
%Type = Article
\bibitem[{Vu et~al.(2024)Vu, Kim, Nguyen, Luu, Kim and
  Yoo}]{10.1109/TPAMI.2023.3326189}
\bibinfo{author}{Vu, T.}, \bibinfo{author}{Kim, K.}, \bibinfo{author}{Nguyen,
  T.}, \bibinfo{author}{Luu, T.M.}, \bibinfo{author}{Kim, J.},
  \bibinfo{author}{Yoo, C.D.}, \bibinfo{year}{2024}.
\newblock \bibinfo{title}{Scalable softgroup for 3d instance segmentation on
  point clouds}.
\newblock \bibinfo{journal}{IEEE Trans. Pattern Anal. Mach. Intell.}
%Type = Article
\bibitem[{Wang(2020)}]{WANG202086}
\bibinfo{author}{Wang, D.}, \bibinfo{year}{2020}.
\newblock \bibinfo{title}{Unsupervised semantic and instance segmentation of
  forest point clouds}.
\newblock \bibinfo{journal}{ISPRS Journal of Photogrammetry and Remote Sensing}
%Type = Article
\bibitem[{Wang et~al.(2018a)Wang, Laga, Jia, Xie and
  Tabia}]{https://doi.org/10.1111/cgf.13501}
\bibinfo{author}{Wang, G.}, \bibinfo{author}{Laga, H.}, \bibinfo{author}{Jia,
  J.}, \bibinfo{author}{Xie, N.}, \bibinfo{author}{Tabia, H.},
  \bibinfo{year}{2018}a.
\newblock \bibinfo{title}{Statistical modeling of the 3d geometry and topology
  of botanical trees}.
\newblock \bibinfo{journal}{Computer Graphics Forum}
%Type = Inproceedings
\bibitem[{Wang et~al.(2019a)Wang, Huang, Hou, Zhang and Shan}]{8954040}
\bibinfo{author}{Wang, L.}, \bibinfo{author}{Huang, Y.}, \bibinfo{author}{Hou,
  Y.}, \bibinfo{author}{Zhang, S.}, \bibinfo{author}{Shan, J.},
  \bibinfo{year}{2019}a.
\newblock \bibinfo{title}{Graph attention convolution for point cloud semantic
  segmentation}, in: \bibinfo{booktitle}{2019 IEEE/CVF Conference on Computer
  Vision and Pattern Recognition (CVPR)},
%Type = Article
\bibitem[{Wang et~al.(2023)Wang, Miao, Han, Li, Zhang and
  Peng}]{WangLiuyangArticle}
\bibinfo{author}{Wang, L.}, \bibinfo{author}{Miao, Y.}, \bibinfo{author}{Han,
  Y.}, \bibinfo{author}{Li, H.}, \bibinfo{author}{Zhang, M.},
  \bibinfo{author}{Peng, C.}, \bibinfo{year}{2023}.
\newblock \bibinfo{title}{Extraction of 3d distribution of potato plant cwsi
  based on thermal infrared image and binocular stereovision system}.
\newblock \bibinfo{journal}{Frontiers in Plant Science}
%Type = Inproceedings
\bibitem[{Wang et~al.(2022a)Wang, Zheng and Wang}]{Wang_2022_CVPR}
\bibinfo{author}{Wang, L.}, \bibinfo{author}{Zheng, L.}, \bibinfo{author}{Wang,
  M.}, \bibinfo{year}{2022}a.
\newblock \bibinfo{title}{3d point cloud instance segmentation of lettuce based
  on partnet}, in: \bibinfo{booktitle}{Proceedings of the IEEE/CVF Conference
  on Computer Vision and Pattern Recognition (CVPR) Workshops},
%Type = Article
\bibitem[{Wang and Deng(2018)}]{WANG2018135}
\bibinfo{author}{Wang, M.}, \bibinfo{author}{Deng, W.}, \bibinfo{year}{2018}.
\newblock \bibinfo{title}{Deep visual domain adaptation: A survey}.
\newblock \bibinfo{journal}{Neurocomputing}
%Type = Inproceedings
\bibitem[{Wang et~al.(2018b)Wang, Suo, Ma, Pokrovsky and Urtasun}]{8578372}
\bibinfo{author}{Wang, S.}, \bibinfo{author}{Suo, S.}, \bibinfo{author}{Ma,
  W.C.}, \bibinfo{author}{Pokrovsky, A.}, \bibinfo{author}{Urtasun, R.},
  \bibinfo{year}{2018}b.
\newblock \bibinfo{title}{Deep parametric continuous convolutional neural
  networks}, in: \bibinfo{booktitle}{2018 IEEE/CVF Conference on Computer
  Vision and Pattern Recognition},
%Type = Inproceedings
\bibitem[{Wang et~al.(2018c)Wang, Yu, Huang and Neumann}]{wang2018sgpn}
\bibinfo{author}{Wang, W.}, \bibinfo{author}{Yu, R.}, \bibinfo{author}{Huang,
  Q.}, \bibinfo{author}{Neumann, U.}, \bibinfo{year}{2018}c.
\newblock \bibinfo{title}{Sgpn: Similarity group proposal network for 3d point
  cloud instance segmentation},
%Type = Inproceedings
\bibitem[{Wang et~al.(2020a)Wang, Kong, Shen, Jiang and
  Li}]{10.1007/978-3-030-58523-5_38}
\bibinfo{author}{Wang, X.}, \bibinfo{author}{Kong, T.}, \bibinfo{author}{Shen,
  C.}, \bibinfo{author}{Jiang, Y.}, \bibinfo{author}{Li, L.},
  \bibinfo{year}{2020}a.
\newblock \bibinfo{title}{Solo: Segmenting objects by locations}, in:
  \bibinfo{booktitle}{Computer Vision – ECCV 2020: 16th European Conference,
  Glasgow, UK, August 23–28, 2020, Proceedings, Part XVIII},
  \bibinfo{publisher}{Springer-Verlag}, \bibinfo{address}{Berlin, Heidelberg}.
%Type = Inproceedings
\bibitem[{Wang et~al.(2020b)Wang, Zhang, Kong, Li and
  Shen}]{NEURIPS2020_cd3afef9}
\bibinfo{author}{Wang, X.}, \bibinfo{author}{Zhang, R.}, \bibinfo{author}{Kong,
  T.}, \bibinfo{author}{Li, L.}, \bibinfo{author}{Shen, C.},
  \bibinfo{year}{2020}b.
\newblock \bibinfo{title}{Solov2: Dynamic and fast instance segmentation}, in:
  \bibinfo{editor}{Larochelle, H.}, \bibinfo{editor}{Ranzato, M.},
  \bibinfo{editor}{Hadsell, R.}, \bibinfo{editor}{Balcan, M.},
  \bibinfo{editor}{Lin, H.} (Eds.), \bibinfo{booktitle}{Advances in Neural
  Information Processing Systems}, \bibinfo{publisher}{Curran Associates,
  Inc.}.
%Type = Article
\bibitem[{Wang and Fang(2020)}]{rs12203457}
\bibinfo{author}{Wang, Y.}, \bibinfo{author}{Fang, H.}, \bibinfo{year}{2020}.
\newblock \bibinfo{title}{Estimation of lai with the lidar technology: A
  review}.
\newblock \bibinfo{journal}{Remote Sensing}
%Type = Article
\bibitem[{Wang et~al.(2022b)Wang, Hu, Ren, Yang and Zhai}]{agronomy12081865}
\bibinfo{author}{Wang, Y.}, \bibinfo{author}{Hu, S.}, \bibinfo{author}{Ren,
  H.}, \bibinfo{author}{Yang, W.}, \bibinfo{author}{Zhai, R.},
  \bibinfo{year}{2022}b.
\newblock \bibinfo{title}{3dphenomvs: A low-cost 3d tomato phenotyping pipeline
  using 3d reconstruction point cloud based on multiview images}.
\newblock \bibinfo{journal}{Agronomy}
%Type = Article
\bibitem[{Wang et~al.(2019b)Wang, Sun, Liu, Sarma, Bronstein and
  Solomon}]{dgcnn}
\bibinfo{author}{Wang, Y.}, \bibinfo{author}{Sun, Y.}, \bibinfo{author}{Liu,
  Z.}, \bibinfo{author}{Sarma, S.E.}, \bibinfo{author}{Bronstein, M.M.},
  \bibinfo{author}{Solomon, J.M.}, \bibinfo{year}{2019}b.
\newblock \bibinfo{title}{Dynamic graph cnn for learning on point clouds}.
\newblock
%Type = Article
\bibitem[{Wang et~al.(2017)Wang, Verboven and Nicolai}]{wang2017contrast}
\bibinfo{author}{Wang, Z.}, \bibinfo{author}{Verboven, P.},
  \bibinfo{author}{Nicolai, B.}, \bibinfo{year}{2017}.
\newblock \bibinfo{title}{Contrast-enhanced 3d micro-ct of plant tissues using
  different impregnation techniques}.
\newblock \bibinfo{journal}{Plant methods}
%Type = Article
\bibitem[{Wang et~al.(2018d)Wang, Wang, Yang, Pan and Han}]{WANG20181}
\bibinfo{author}{Wang, Z.}, \bibinfo{author}{Wang, K.}, \bibinfo{author}{Yang,
  F.}, \bibinfo{author}{Pan, S.}, \bibinfo{author}{Han, Y.},
  \bibinfo{year}{2018}d.
\newblock \bibinfo{title}{Image segmentation of overlapping leaves based on
  chan–vese model and sobel operator}.
\newblock \bibinfo{journal}{Information Processing in Agriculture}
%Type = Article
\bibitem[{Wang et~al.(2015)Wang, Zhang, Fang, Mathiopoulos, Tong, Qu, Xiao, Li
  and Chen}]{6922535}
\bibinfo{author}{Wang, Z.}, \bibinfo{author}{Zhang, L.}, \bibinfo{author}{Fang,
  T.}, \bibinfo{author}{Mathiopoulos, P.T.}, \bibinfo{author}{Tong, X.},
  \bibinfo{author}{Qu, H.}, \bibinfo{author}{Xiao, Z.}, \bibinfo{author}{Li,
  F.}, \bibinfo{author}{Chen, D.}, \bibinfo{year}{2015}.
\newblock \bibinfo{title}{A multiscale and hierarchical feature extraction
  method for terrestrial laser scanning point cloud classification}.
\newblock \bibinfo{journal}{IEEE Transactions on Geoscience and Remote Sensing}
%Type = Article
\bibitem[{Weinmann et~al.(2015)Weinmann, Schmidt, Mallet, Hinz, Rottensteiner
  and Jutzi}]{Weinmannarticle}
\bibinfo{author}{Weinmann, M.}, \bibinfo{author}{Schmidt, A.},
  \bibinfo{author}{Mallet, C.}, \bibinfo{author}{Hinz, S.},
  \bibinfo{author}{Rottensteiner, F.}, \bibinfo{author}{Jutzi, B.},
  \bibinfo{year}{2015}.
\newblock \bibinfo{title}{Contextual classification of point cloud data by
  exploiting individual 3d neighborhoods}.
\newblock \bibinfo{journal}{ISPRS Annals of the Photogrammetry, Remote Sensing
  and Spatial Information Sciences}
%Type = Article
\bibitem[{Winiwarter et~al.(2022)Winiwarter, {Esmorís Pena}, Weiser, Anders,
  {Martínez Sánchez}, Searle and Höfle}]{heliosPlusPlus}
\bibinfo{author}{Winiwarter, L.}, \bibinfo{author}{{Esmorís Pena}, A.M.},
  \bibinfo{author}{Weiser, H.}, \bibinfo{author}{Anders, K.},
  \bibinfo{author}{{Martínez Sánchez}, J.}, \bibinfo{author}{Searle, M.},
  \bibinfo{author}{Höfle, B.}, \bibinfo{year}{2022}.
\newblock \bibinfo{title}{Virtual laser scanning with helios++: A novel take on
  ray tracing-based simulation of topographic full-waveform 3d laser scanning}.
\newblock \bibinfo{journal}{Remote Sensing of Environment}
%Type = Article
\bibitem[{Wu et~al.(2022)Wu, Wen, Gou, Lu, Zhang, Zheng, Xiang, Chen and
  Guo}]{10.3389/fpls.2022.897746}
\bibinfo{author}{Wu, S.}, \bibinfo{author}{Wen, W.}, \bibinfo{author}{Gou, W.},
  \bibinfo{author}{Lu, X.}, \bibinfo{author}{Zhang, W.},
  \bibinfo{author}{Zheng, C.}, \bibinfo{author}{Xiang, Z.},
  \bibinfo{author}{Chen, L.}, \bibinfo{author}{Guo, X.}, \bibinfo{year}{2022}.
\newblock \bibinfo{title}{A miniaturized phenotyping platform for individual
  plants using multi-view stereo 3d reconstruction}.
\newblock \bibinfo{journal}{Frontiers in Plant Science}
%Type = Article
\bibitem[{Wu et~al.(2020)Wu, Wen, Wang, Fan, Wang, Gou and Guo}]{WU20201848437}
\bibinfo{author}{Wu, S.}, \bibinfo{author}{Wen, W.}, \bibinfo{author}{Wang,
  Y.}, \bibinfo{author}{Fan, J.}, \bibinfo{author}{Wang, C.},
  \bibinfo{author}{Gou, W.}, \bibinfo{author}{Guo, X.}, \bibinfo{year}{2020}.
\newblock \bibinfo{title}{Mvs-pheno: A portable and low-cost phenotyping
  platform for maize shoots using multiview stereo 3d reconstruction}.
\newblock \bibinfo{journal}{Plant Phenomics}
%Type = Inproceedings
\bibitem[{Wu et~al.(2019)Wu, Qi and Li}]{PointConvinproceedings}
\bibinfo{author}{Wu, W.}, \bibinfo{author}{Qi, Z.}, \bibinfo{author}{Li, F.},
  \bibinfo{year}{2019}.
\newblock \bibinfo{title}{Pointconv: Deep convolutional networks on 3d point
  clouds},
%Type = Incollection
\bibitem[{Wyatt(2016)}]{WYATT201651}
\bibinfo{author}{Wyatt, J.}, \bibinfo{year}{2016}.
\newblock \bibinfo{title}{Grain and plant morphology of cereals and how
  characters can be used to identify varieties}, in: \bibinfo{editor}{Wrigley,
  C.}, \bibinfo{editor}{Corke, H.}, \bibinfo{editor}{Seetharaman, K.},
  \bibinfo{editor}{Faubion, J.} (Eds.), \bibinfo{booktitle}{Encyclopedia of
  Food Grains (Second Edition)}. \bibinfo{edition}{second edition} ed..
  \bibinfo{publisher}{Academic Press}, \bibinfo{address}{Oxford},
%Type = Article
\bibitem[{Xi et~al.(2020)Xi, Hopkinson, Rood and Peddle}]{XI20201}
\bibinfo{author}{Xi, Z.}, \bibinfo{author}{Hopkinson, C.},
  \bibinfo{author}{Rood, S.B.}, \bibinfo{author}{Peddle, D.R.},
  \bibinfo{year}{2020}.
\newblock \bibinfo{title}{See the forest and the trees: Effective machine and
  deep learning algorithms for wood filtering and tree species classification
  from terrestrial laser scanning}.
\newblock \bibinfo{journal}{ISPRS Journal of Photogrammetry and Remote Sensing}
%Type = Article
\bibitem[{Xiang et~al.(2024)Xiang, Wielgosz, Kontogianni, Peters, Puliti,
  Astrup and Schindler}]{XIANG2024114078}
\bibinfo{author}{Xiang, B.}, \bibinfo{author}{Wielgosz, M.},
  \bibinfo{author}{Kontogianni, T.}, \bibinfo{author}{Peters, T.},
  \bibinfo{author}{Puliti, S.}, \bibinfo{author}{Astrup, R.},
  \bibinfo{author}{Schindler, K.}, \bibinfo{year}{2024}.
\newblock \bibinfo{title}{Automated forest inventory: Analysis of high-density
  airborne lidar point clouds with 3d deep learning}.
\newblock \bibinfo{journal}{Remote Sensing of Environment}
%Type = Article
\bibitem[{Xiao et~al.(2020)Xiao, Chai, Shao, Shen, Wang, Wang, Sui and
  Ma}]{rs12020269}
\bibinfo{author}{Xiao, S.}, \bibinfo{author}{Chai, H.}, \bibinfo{author}{Shao,
  K.}, \bibinfo{author}{Shen, M.}, \bibinfo{author}{Wang, Q.},
  \bibinfo{author}{Wang, R.}, \bibinfo{author}{Sui, Y.}, \bibinfo{author}{Ma,
  Y.}, \bibinfo{year}{2020}.
\newblock \bibinfo{title}{Image-based dynamic quantification of aboveground
  structure of sugar beet in field}.
\newblock \bibinfo{journal}{Remote Sensing}
%Type = Article
\bibitem[{Xie et~al.(2023)Xie, Du, Ma and
  Cen}]{doi:10.34133/plantphenomics.0040}
\bibinfo{author}{Xie, P.}, \bibinfo{author}{Du, R.}, \bibinfo{author}{Ma, Z.},
  \bibinfo{author}{Cen, H.}, \bibinfo{year}{2023}.
\newblock \bibinfo{title}{Generating 3d multispectral point clouds of plants
  with fusion of snapshot spectral and rgb-d images}.
\newblock \bibinfo{journal}{Plant Phenomics}
%Type = Article
\bibitem[{Xie et~al.(2024)Xie, Ma, Du, Yang, Jiang and Cen}]{XIE20241624}
\bibinfo{author}{Xie, P.}, \bibinfo{author}{Ma, Z.}, \bibinfo{author}{Du, R.},
  \bibinfo{author}{Yang, X.}, \bibinfo{author}{Jiang, Y.},
  \bibinfo{author}{Cen, H.}, \bibinfo{year}{2024}.
\newblock \bibinfo{title}{An unmanned ground vehicle phenotyping-based method
  to generate three-dimensional multispectral point clouds for deciphering
  spatial heterogeneity in plant traits}.
\newblock \bibinfo{journal}{Molecular Plant}
%Type = Article
\bibitem[{Xie et~al.(2020)Xie, Tian and Zhu}]{9028090}
\bibinfo{author}{Xie, Y.}, \bibinfo{author}{Tian, J.}, \bibinfo{author}{Zhu,
  X.X.}, \bibinfo{year}{2020}.
\newblock \bibinfo{title}{Linking points with labels in 3d: A review of point
  cloud semantic segmentation}.
\newblock \bibinfo{journal}{IEEE Geoscience and Remote Sensing Magazine}
%Type = Article
\bibitem[{Xin et~al.(2023)Xin, Sun, Bartholomeus and
  Kootstra}]{10.3389/fpls.2023.1045545}
\bibinfo{author}{Xin, B.}, \bibinfo{author}{Sun, J.},
  \bibinfo{author}{Bartholomeus, H.}, \bibinfo{author}{Kootstra, G.},
  \bibinfo{year}{2023}.
\newblock \bibinfo{title}{3d data-augmentation methods for semantic
  segmentation of tomato plant parts}.
\newblock \bibinfo{journal}{Frontiers in Plant Science}
%Type = Inproceedings
\bibitem[{Xu et~al.(2021)Xu, Ding, Zhao and Qi}]{xu2021paconv}
\bibinfo{author}{Xu, M.}, \bibinfo{author}{Ding, R.}, \bibinfo{author}{Zhao,
  H.}, \bibinfo{author}{Qi, X.}, \bibinfo{year}{2021}.
\newblock \bibinfo{title}{Paconv: Position adaptive convolution with dynamic
  kernel assembling on point clouds},
%Type = Article
\bibitem[{Xu and Li(2022)}]{doi:10.34133/2022/9760269}
\bibinfo{author}{Xu, R.}, \bibinfo{author}{Li, C.}, \bibinfo{year}{2022}.
\newblock \bibinfo{title}{A review of high-throughput field phenotyping
  systems: Focusing on ground robots}.
\newblock \bibinfo{journal}{Plant Phenomics}
%Type = Inbook
\bibitem[{Yang et~al.(2019)Yang, Wang, Clark, Hu, Wang, Markham and
  Trigoni}]{10.5555/3454287.3454892}
\bibinfo{author}{Yang, B.}, \bibinfo{author}{Wang, J.}, \bibinfo{author}{Clark,
  R.}, \bibinfo{author}{Hu, Q.}, \bibinfo{author}{Wang, S.},
  \bibinfo{author}{Markham, A.}, \bibinfo{author}{Trigoni, N.},
  \bibinfo{year}{2019}.
\newblock \bibinfo{title}{Learning object bounding boxes for 3D instance
  segmentation on point clouds}.
%Type = Article
\bibitem[{Yang et~al.(2020)Yang, Feng, Zhang, Zhang, Doonan, Batchelor, Xiong
  and Yan}]{YANG2020187}
\bibinfo{author}{Yang, W.}, \bibinfo{author}{Feng, H.}, \bibinfo{author}{Zhang,
  X.}, \bibinfo{author}{Zhang, J.}, \bibinfo{author}{Doonan, J.H.},
  \bibinfo{author}{Batchelor, W.D.}, \bibinfo{author}{Xiong, L.},
  \bibinfo{author}{Yan, J.}, \bibinfo{year}{2020}.
\newblock \bibinfo{title}{Crop phenomics and high-throughput phenotyping: Past
  decades, current challenges, and future perspectives}.
\newblock \bibinfo{journal}{Molecular Plant}
%Type = Article
\bibitem[{Yang et~al.(2024)Yang, Miao, Tian, Wang, Zhao, Lin, Zhu, Yang and
  Xu}]{SYAUMaize}
\bibinfo{author}{Yang, X.}, \bibinfo{author}{Miao, T.}, \bibinfo{author}{Tian,
  X.}, \bibinfo{author}{Wang, D.}, \bibinfo{author}{Zhao, J.},
  \bibinfo{author}{Lin, L.}, \bibinfo{author}{Zhu, C.}, \bibinfo{author}{Yang,
  T.}, \bibinfo{author}{Xu, T.}, \bibinfo{year}{2024}.
\newblock \bibinfo{title}{Maize stem–leaf segmentation framework based on
  deformable point clouds}.
\newblock \bibinfo{journal}{ISPRS Journal of Photogrammetry and Remote Sensing}
%Type = Article
\bibitem[{Yi et~al.(2016)Yi, Kim, Ceylan, Shen, Yan, Su, Lu, Huang, Sheffer and
  Guibas}]{10.1145/2980179.2980238}
\bibinfo{author}{Yi, L.}, \bibinfo{author}{Kim, V.G.}, \bibinfo{author}{Ceylan,
  D.}, \bibinfo{author}{Shen, I.C.}, \bibinfo{author}{Yan, M.},
  \bibinfo{author}{Su, H.}, \bibinfo{author}{Lu, C.}, \bibinfo{author}{Huang,
  Q.}, \bibinfo{author}{Sheffer, A.}, \bibinfo{author}{Guibas, L.},
  \bibinfo{year}{2016}.
\newblock \bibinfo{title}{A scalable active framework for region annotation in
  3d shape collections}.
\newblock \bibinfo{journal}{ACM Trans. Graph.}
%Type = Inproceedings
\bibitem[{Yi et~al.(2019)Yi, Zhao, Wang, Sung and Guibas}]{8953913}
\bibinfo{author}{Yi, L.}, \bibinfo{author}{Zhao, W.}, \bibinfo{author}{Wang,
  H.}, \bibinfo{author}{Sung, M.}, \bibinfo{author}{Guibas, L.J.},
  \bibinfo{year}{2019}.
\newblock \bibinfo{title}{Gspn: Generative shape proposal network for 3d
  instance segmentation in point cloud}, in: \bibinfo{booktitle}{2019 IEEE/CVF
  Conference on Computer Vision and Pattern Recognition (CVPR)},
%Type = Inproceedings
\bibitem[{You et~al.(2022)You, Grimm and Davidson}]{OpticalFlowinproceedings}
\bibinfo{author}{You, A.}, \bibinfo{author}{Grimm, C.},
  \bibinfo{author}{Davidson, J.}, \bibinfo{year}{2022}.
\newblock \bibinfo{title}{Optical flow-based branch segmentation for complex
  orchard environments},
%Type = Article
\bibitem[{Yun et~al.(2025)Yun, Eichhorn, Jin, Yuan, Fang, Lu, Wang and
  Zhang}]{YUN2025110319}
\bibinfo{author}{Yun, T.}, \bibinfo{author}{Eichhorn, M.P.},
  \bibinfo{author}{Jin, S.}, \bibinfo{author}{Yuan, X.}, \bibinfo{author}{Fang,
  W.}, \bibinfo{author}{Lu, X.}, \bibinfo{author}{Wang, X.},
  \bibinfo{author}{Zhang, H.}, \bibinfo{year}{2025}.
\newblock \bibinfo{title}{A framework for phenotyping rubber trees under
  intense wind stress using laser scanning and digital twin technology}.
\newblock \bibinfo{journal}{Agricultural and Forest Meteorology}
%Type = Article
\bibitem[{Zarei et~al.(2024)Zarei, Li, Schnable, Lyons, Pauli, Barnard and
  Benes}]{ZAREI2024108922}
\bibinfo{author}{Zarei, A.}, \bibinfo{author}{Li, B.},
  \bibinfo{author}{Schnable, J.C.}, \bibinfo{author}{Lyons, E.},
  \bibinfo{author}{Pauli, D.}, \bibinfo{author}{Barnard, K.},
  \bibinfo{author}{Benes, B.}, \bibinfo{year}{2024}.
\newblock \bibinfo{title}{Plantsegnet: 3d point cloud instance segmentation of
  nearby plant organs with identical semantics}.
\newblock \bibinfo{journal}{Computers and Electronics in Agriculture}
%Type = Misc
\bibitem[{Zhai et~al.(2024)Zhai, Wang, Li, Jiang, Zhou, Wang, Jin, Guan and
  Wang}]{zhai2024cropcraftinverseproceduralmodeling}
\bibinfo{author}{Zhai, A.J.}, \bibinfo{author}{Wang, X.}, \bibinfo{author}{Li,
  K.}, \bibinfo{author}{Jiang, Z.}, \bibinfo{author}{Zhou, J.},
  \bibinfo{author}{Wang, S.}, \bibinfo{author}{Jin, Z.}, \bibinfo{author}{Guan,
  K.}, \bibinfo{author}{Wang, S.}, \bibinfo{year}{2024}.
\newblock
%Type = Article
\bibitem[{Zhang et~al.(2024)Zhang, Song, Ou, Liu, Li, Lu, Xu, Su, Jiang, Ding,
  Xia, Guo, Wu, Zhang, Wang and Jin}]{doi:10.34133/plantphenomics.0190}
\bibinfo{author}{Zhang, S.}, \bibinfo{author}{Song, Y.}, \bibinfo{author}{Ou,
  R.}, \bibinfo{author}{Liu, Y.}, \bibinfo{author}{Li, S.},
  \bibinfo{author}{Lu, X.}, \bibinfo{author}{Xu, S.}, \bibinfo{author}{Su, Y.},
  \bibinfo{author}{Jiang, D.}, \bibinfo{author}{Ding, Y.},
  \bibinfo{author}{Xia, H.}, \bibinfo{author}{Guo, Q.}, \bibinfo{author}{Wu,
  J.}, \bibinfo{author}{Zhang, J.}, \bibinfo{author}{Wang, J.},
  \bibinfo{author}{Jin, S.}, \bibinfo{year}{2024}.
\newblock \bibinfo{title}{Scag: A stratified, clustered, and growing-based
  algorithm for soybean branch angle extraction and ideal plant architecture
  evaluation}.
\newblock \bibinfo{journal}{Plant Phenomics}
%Type = Inproceedings
\bibitem[{Zhao et~al.(2019)Zhao, Jiang, Fu and Jia}]{8954075}
\bibinfo{author}{Zhao, H.}, \bibinfo{author}{Jiang, L.}, \bibinfo{author}{Fu,
  C.W.}, \bibinfo{author}{Jia, J.}, \bibinfo{year}{2019}.
\newblock \bibinfo{title}{Pointweb: Enhancing local neighborhood features for
  point cloud processing}, in: \bibinfo{booktitle}{2019 IEEE/CVF Conference on
  Computer Vision and Pattern Recognition (CVPR)},
%Type = Article
\bibitem[{Zheng et~al.(2013)Zheng, Moskal and Kim}]{6249740}
\bibinfo{author}{Zheng, G.}, \bibinfo{author}{Moskal, L.M.},
  \bibinfo{author}{Kim, S.H.}, \bibinfo{year}{2013}.
\newblock \bibinfo{title}{Retrieval of effective leaf area index in
  heterogeneous forests with terrestrial laser scanning}.
\newblock \bibinfo{journal}{IEEE Transactions on Geoscience and Remote Sensing}
%Type = Article
\bibitem[{Zhu et~al.(2020)Zhu, Liu, Xie, Guo, Li and
  Ma}]{zhu2020quantification}
\bibinfo{author}{Zhu, B.}, \bibinfo{author}{Liu, F.}, \bibinfo{author}{Xie,
  Z.}, \bibinfo{author}{Guo, Y.}, \bibinfo{author}{Li, B.},
  \bibinfo{author}{Ma, Y.}, \bibinfo{year}{2020}.
\newblock \bibinfo{title}{Quantification of light interception within
  image-based 3-d reconstruction of sole and intercropped canopies over the
  entire growth season}.
\newblock \bibinfo{journal}{Annals of botany}
%Type = Article
\bibitem[{Zhu et~al.(2024)Zhu, Zhai, Ren, Xie, Du, He, Cui, Wang, Ye, Wang
  et~al.}]{zhu2024crops3d}
\bibinfo{author}{Zhu, J.}, \bibinfo{author}{Zhai, R.}, \bibinfo{author}{Ren,
  H.}, \bibinfo{author}{Xie, K.}, \bibinfo{author}{Du, A.},
  \bibinfo{author}{He, X.}, \bibinfo{author}{Cui, C.}, \bibinfo{author}{Wang,
  Y.}, \bibinfo{author}{Ye, J.}, \bibinfo{author}{Wang, J.}, et~al.,
  \bibinfo{year}{2024}.
\newblock \bibinfo{title}{Crops3d: a diverse 3d crop dataset for realistic
  perception and segmentation toward agricultural applications}.
\newblock \bibinfo{journal}{Scientific Data}
%Type = Inproceedings
\bibitem[{Zolanvari et~al.(2019)Zolanvari, Ruano, Rana, Cummins, Smolic,
  Da~Silva and Rahbar}]{DublinCityinproceedings}
\bibinfo{author}{Zolanvari, I.}, \bibinfo{author}{Ruano, S.},
  \bibinfo{author}{Rana, A.}, \bibinfo{author}{Cummins, A.},
  \bibinfo{author}{Smolic, A.}, \bibinfo{author}{Da~Silva, R.},
  \bibinfo{author}{Rahbar, M.}, \bibinfo{year}{2019}.
\newblock

\end{thebibliography}

% Biography
%\bio{}
% Here goes the biography details.
%\endbio

%\bio{pic1}
% Here goes the biography details.
%\endbio

\end{document}